\documentclass[preprint,12pt]{elsarticle}

\usepackage{amssymb}
\usepackage{amsmath}

\usepackage{todonotes}
\usepackage{xcolor}
\usepackage{graphicx}

\newcommand{\xmark}{\ding{55}}%
\usepackage{rotating}
\usepackage{tabularx}
\usepackage{amssymb}
\usepackage{pifont}
\usepackage{adjustbox}
\usepackage{caption}
\usepackage{booktabs}
\usepackage{subcaption}
\usepackage{url}

\newcommand{\greencheck}{{\color{green}\checkmark}}
\newcommand{\redxmark}{{\color{red}\xmark}}

\usepackage[acronym]{glossaries}
\newacronym{ALS}{ALS}{airborne laser scanning}
\newacronym{MLS}{MLS}{mobile laser scanning}
\newacronym{LoD}{LoD}{level of detail}
\newacronym{OGC}{OGC}{Open Geospatial Consortium}
\newacronym{GML}{GML}{Geography Markup Language}
\newacronym{ASAM}{ASAM}{Association for Standardization of Automation and Measuring Systems}
\newacronym{TLS}{TLS}{terrestrial laser scanning}
\newacronym{UAV}{UAV}{unmanned aerial vehicle}
\newacronym{HD}{HD}{high definition}
\newacronym{RANSAC}{RANSAC}{RANdom SAmple Consensus}
\newacronym{ROI}{ROI}{region of interest}
\newacronym{DEM}{DEM}{digital elevation model}
\newacronym{ICP}{ICP}{iterative closest point}
\newacronym{NLOS}{NLOS}{non-line-of-sight}
\newacronym{SfM}{SfM}{structure from motion}
\newacronym{FME}{FME}{Feature Manipulation Engine}
\newacronym{OSM}{OSM}{OpenStreetMap} 
\newacronym{RMSE}{RMSE}{root mean square error}
\newacronym{CPT}{CPT}{conditional probability table}
\newacronym{DST}{DST}{Dempster–Shafer theory}
\newacronym{BN}{BN}{Bayesian network}
\newacronym{GIS}{GIS}{Geographic Information System}
\newacronym{PPD}{PPD}{posterior probability distribution}
\newacronym{CI}{CI}{confidence interval}
\newacronym{IFC}{IFC}{Industry Foundation Classes}
\newacronym{CRS}{CRS}{coordinate reference system}
\newacronym{LoFG}{LoFG}{Level of Facade Generalization}
\newacronym{BNN}{BNN}{Bayesian neural network}
\newacronym{BNNs}{BNNs}{Bayesian neural networks}
\newacronym{BIM}{BIM}{building information modeling}
\newacronym{BIMs}{BIMs}{building information models}
\newacronym{AEC}{AEC}{Architecture, Engineering, and Construction}
\newacronym{LoA}{LOA}{Level of Accuracy}
\newacronym{DoF}{DoF}{degrees of freedom}
\newacronym{CNN}{CNN}{convolutional neural network}
\newacronym{CAD}{CAD}{computer-aided design}
\newacronym{DT}{DT}{Digital Twin}
\newacronym{UDT}{UDT}{Urban Digital Twin}
\newacronym{SUMO}{SUMO}{Simulation of Urban MObility}
\newacronym{NeRF}{NeRF}{Neural Radiance Fields}
\newacronym{3DGS}{3DGS}{3D Gaussian Splatting}
\newacronym{CM}{CM}{2D Conflict Maps}
\newacronym{DM}{DM}{Diffusion Probabilistic Models}
\newacronym{TIR}{TIR}{Thermal Infrared}
\newacronym{UAS}{UAS}{uncrewed aerial system}
\newacronym{RTK}{RTK}{Real Time Kinematic}

\journal{ISPRS Journal of Photogrammetry and Remote Sensing}

\begin{document}

\begin{frontmatter}



\title{TUM2TWIN:\\ Introducing the Large-Scale Multimodal Urban Digital Twin Benchmark Dataset} 


\author[photogrammetry]{Olaf Wysocki}
\author[geoinformatics]{Benedikt Schwab}
\author[photogrammetry]{Manoj Kumar Biswanath}
\author[photogrammetry]{Michael Greza}
\author[photogrammetry]{Qilin Zhang}
\author[photogrammetry]{Jingwei Zhu}
\author[geoinformatics]{Thomas Froech}
\author[geoinformatics]{Medhini Heeramaglore}
\author[geoinformatics]{Ihab Hijazi}
\author[geoinformatics]{Khaoula Kanna}
\author[traffic]{Mathias Pechinger}
\author[sipeo]{Zhaiyu Chen}
\author[sipeo,dlr]{Yao Sun}
\author[architecture]{Alejandro Rueda Segura}
\author[geodesy]{Ziyang Xu}
\author[geodesy]{Omar AbdelGafar}
\author[CCBE]{Mansour Mehranfar}
\author[3dAIlab]{Chandan Yeshwanth}
\author[3dAIlab]{Yueh-Cheng Liu}
\author[greenTechno]{Hadi Yazdi}
\author[rsApplications]{Jiapan Wang}
\author[dlr]{Stefan Auer}
\author[rsApplications]{Katharina Anders}
\author[traffic]{Klaus Bogenberger}
\author[CCBE]{André Borrmann}
\author[3dAIlab]{Angela Dai}
\author[hm]{Ludwig Hoegner}
\author[geodesy]{Christoph Holst}
\author[geoinformatics]{Thomas H. Kolbe}
\author[greenTechno]{Ferdinand Ludwig}
\author[visualComptAI]{Matthias Nießner}
\author[architecture]{Frank Petzold}
\author[sipeo]{Xiao Xiang Zhu}
\author[photogrammetry]{Boris Jutzi}

\affiliation[photogrammetry]{organization={Photogrammetry and Remote Sensing},
}
            
\affiliation[geoinformatics]{organization={Geoinformatics},
} 
        
\affiliation[traffic]{organization={Traffic Engineering and Control},
}

\affiliation[sipeo]{organization={Data Science in Earth Observation},
}

\affiliation[geodesy]{organization={Engineering Geodesy},
}

\affiliation[3dAIlab]{organization={3D AI Laboratory},
}

\affiliation[visualComptAI]{organization={Visual Computing \& Artificial Intelligence Lab},
}

\affiliation[CCBE]{organization={Computing in Civil and Building
Engineering},}

\affiliation[greenTechno]{organization={Green Technologies in Landscape Architecture},}

\affiliation[rsApplications]{organization={Remote Sensing Applications},}
            
\affiliation[architecture]{organization={Architectural Informatics, Technical University of Munich (TUM)},
}
            
\affiliation[hm]{organization={Geoinformatics, University of Applied Sciences Munich (HM)}
}

\affiliation[dlr]{organization={Remote Sensing Technology Institute, Photogrammetry and Image Analysis Department, German Aerospace Center (DLR)}
}

\begin{abstract}
Urban Digital Twins (UDTs) have become essential for managing cities and integrating complex, heterogeneous data from diverse sources. 
Creating UDTs involves challenges at multiple process stages, including acquiring accurate 3D source data, reconstructing high-fidelity 3D models, maintaining models' updates, and ensuring seamless interoperability to downstream tasks.
Current datasets are usually limited to one part of the processing chain, hampering comprehensive~\gls{UDT}s validation.
To address these challenges, we introduce the first comprehensive multimodal Urban Digital Twin benchmark dataset: TUM2TWIN. 
This dataset includes georeferenced, semantically aligned 3D models and networks along with various terrestrial, mobile, aerial, and satellite observations boasting 32 data subsets over roughly 100,000 $m^2$ and currently 767 GB of data. 
By ensuring georeferenced indoor-outdoor acquisition, high accuracy, and multimodal data integration, the benchmark supports robust analysis of sensors and the development of advanced reconstruction methods. 
Additionally, we explore downstream tasks demonstrating the potential of TUM2TWIN, including novel view synthesis of NeRF and Gaussian Splatting, solar potential analysis, point cloud semantic segmentation, and LoD3 building reconstruction. 
We are convinced this contribution lays a foundation for overcoming current limitations in UDT creation, fostering new research directions and practical solutions for smarter, data-driven urban environments. 
The project is available under: \url{https://tum2t.win}
\end{abstract}




\end{frontmatter}


\section{Introduction}
\label{intro}
%
%
\begin{figure*}
    \centering
    \includegraphics[width=0.9\linewidth]{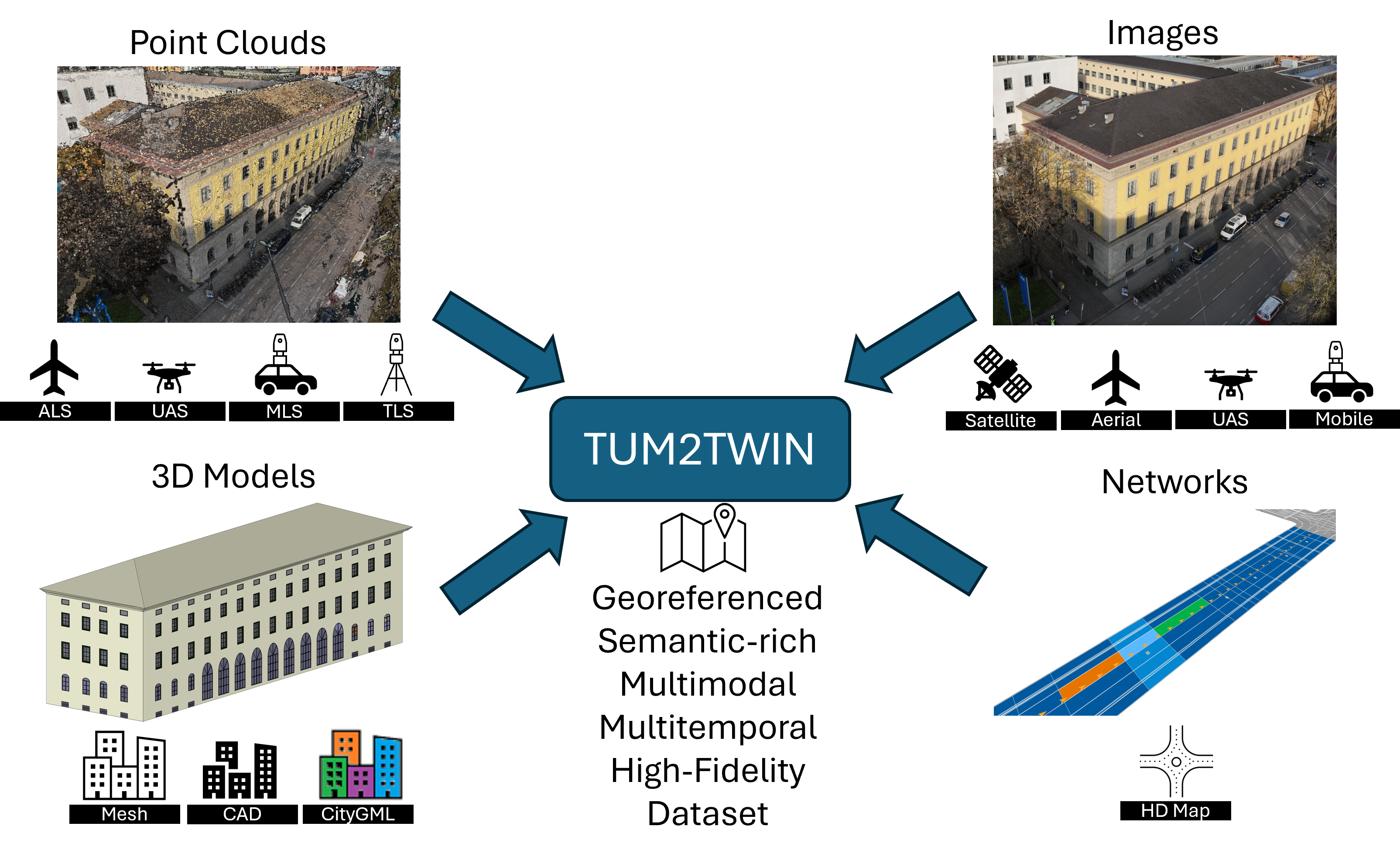}
    \caption{TUM2TWIN: Georeferenced, semantic-rich, multimodal, multitemporal, and high-fidelity benchmark dataset for the development of urban digital twins (UDT).  }
    \label{fig:MainOverview}
\end{figure*}
\begin{figure*}
    \centering
    \includegraphics[width=0.9\linewidth]{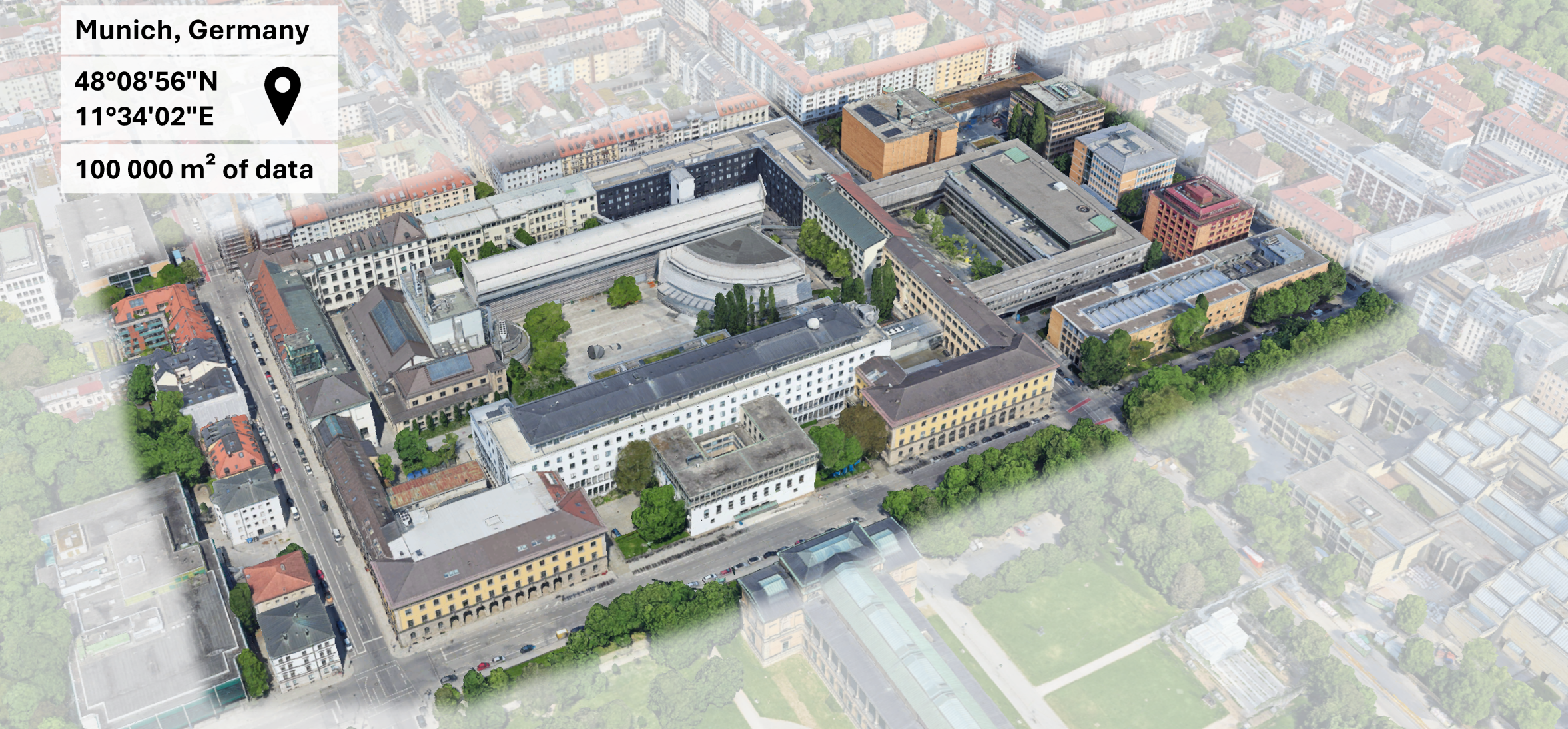}
    \caption{TUM2TWIN covers approximately 100 000 $m^2$ of the center of Munich, Germany (highlighted), boasting 32 data subsets and currently totaling 767 GB.  }
    \label{fig:overviewVisual}
\end{figure*}
The idea of a~\gls{DT} has been commonly attributed to the NASA team in the 1960s, when they created a digital representation of the Apollo mission \cite{allen2021digital}.
The next decades have witnessed proliferation of~\gls{DT}s in the product life cycle management and industrial machines with the main objective to record object's state, history, and performance data \cite{DENG2021125,Kolbe2021}.

Unlike the standard~\gls{DT},~\gls{UDT} centers around city objects which are large-scale and posses heterogeneous information of various stakeholders, rendering them hardly tractable.   
As such, the common strategy is to follow the reverse engineering approach and twin 3D objects by i) acquiring 3D data with various sensors, such as cameras or laser scanners; ii) followed by 3D automatic or manual surface modeling; and iii) frequent updates on the object state \cite{HAALA2010570,Kolbe2021,grieves2014digital}.

Although a great body of research has been devoted to researching methods for robust 3D data acquisition and 3D reconstruction there still remain unresolved challenges.
These pertain to shortcomings of particular sensor types, e.g., inability to acquire laser-based point clouds owing to laser penetration of translucent objects \cite{wysocki2023scan2lod3};
and also to 3D reconstruction methods limitations, e.g., hierarchical reasoning about 3D object semantics insufficiently satisfying international modeling standards such as~\gls{IFC}- and CityGML requirements.

Above all, we identify the lack of high-accuracy and multimodal data as the crucial impediment in both understanding sensor limitations and developing robust methods, especially in the era of data-driven machine learning methods.

To address the identified research gap, in TUM2TWIN, we propose:
\begin{itemize}
    \item Introducing to date the first Urban Digital Twin benchmark dataset comprising multitemporal, multimodal, and high-fidelity currently boasting 32 data instances of a real city.
    \item Presenting to date the first georeferenced semantic 3D building models at~\gls{LoD}1, 2, and 3, as well as terrestrial and aerial images and laser scanning point clouds with up to the cm-level global accuracy.
    \item Identifying potential and already realized use cases for multimodal data setup enabling opening new research directions. 
\end{itemize}

\section{Related Benchmark Datasets}
\label{rw}
%
%
The research community has been introducing various benchmark datasets for decades \cite{scharstein2002taxonomy}.
Their primary purpose is to enable i) evaluating one's novel method on the comprehensive set of examples, ii) allowing for homogeneous comparison to other methods, and iii) presenting tangible blueprints for theoretical data concepts. 
Therefore, in any benchmark dataset, the so-called ground-truth \footnote{also known as reference data} goal is essential, which usually translates to the ideal, desired outcome when applying a method on a specific set.
For example, the ground truth data can represent referenced 3D geometrical representation, given raw 2D sensor data. 

Including some notion of semantics in benchmark datasets is nowadays common (Tab.~\ref{tab:comparisonTable}).
An excellent example of this trend is image-based benchmarks, which now provide billions of training and validation samples, rendering image semantic segmentation an already production-ready solution \cite{kirillov2023segment}. 
This trend has also been translated to other domains, as we observe in the point cloud segmentation or object reconstruction domain.
Still, such challenging topics as facade semantic segmentation on point clouds or large-scale semantic 3D reconstruction have not been seen in multiple datasets, mainly owing to the much more cumbersome and tedious annotation process than image annotation.

Owing to the recent trend of immediate accessibility of various data types, benchmark datasets have also witnessed their increasing availability. 
Yet, most of the datasets concentrate on a selected issue with a limited number of modalities. 
One of the well-known examples is the semantic3D.net \cite{hackel2017semantic3d}, where the manually annotated semantic point cloud classes serve as an evaluation for point cloud semantic segmentation approaches.
This semantic point cloud dataset, along with several notable ones \cite{SydneyDatasetde2013unsupervised,archDatasetPaper,deschaud2021pariscarla3d, tan2020toronto,serna2014parisMadame} significantly contribute the research on the large-scale semantic segmentation of point clouds, enabling testing of multiple methods. 
Yet, these datasets do not comprise any other data representations enabling further validation of multimodal approaches.
For example, semantic point clouds are frequently used for 3D semantic object reconstruction, which necessitates 3D ground-truth models ideally derived from the same input point cloud.

Another limitation of the typical benchmark dataset is its lack of a multitemporal ground truth. 
Usually, they comprise single-timestamp data acquired and implicitly assume scene time-wise coherence.
The rationale behind this correlates with the objective of concentrating on the single sensor or single modality.
Worth noting are exceptions for the change-detection-oriented dataset, which are, however, usually also limited to the single modality of aerial or satellite observations \cite{yushengChangeDetectionReview}.  

Coherence can also be fostered by leveraging global coordinate reference systems, enabling georeferencing and tractability of changes in relation to any object in any modality located on the globe.
As such, it is crucial to maintain the georeferencing aspect when dealing with multitemporal and multimodal data, especially in the context of~\gls{UDT}.
However, this aspect is often neglected, and most benchmarks concentrate on the locally defined coordinate systems, which limits their re-utilization in other approaches \cite{tan2020toronto,armeni20163d}. 
Also, it limits their validation set, as there are limited reference datasets co-aligned, even though there might be many of other resourceful datasets in place.
For example, drone-based point cloud acquisition, when georeferenced, can be integrated with national georeferenced point clouds stemming from airplane-based acquisition \cite{nex2024usegeo}.
\begin{table*}[!htb]
    \captionsetup{size=footnotesize}
    \caption{Related Digital Twin Benchmark Datasets (PCD - point clouds, IMG - Images, NET - Networks, 3DM - 3D Models).} 
    \label{tab:comparisonTable}
    \centering
    \begin{adjustbox}{width=1\textwidth, center}
        \begin{tabular}{lccccccccc}
            \toprule
            Name  & Year & Data Type   &  World   &  \# Data Instances   & Multitemporal &  Multimodal & Georeferenced & Semantics \\
            \midrule
            Sydney Urban Objects Dataset \cite{SydneyDatasetde2013unsupervised}  &   2013   &   PCD   &   real   &   1   & \redxmark &  \redxmark  & \redxmark & \greencheck \\
            Paris-rue-Madame database \cite{serna2014parisMadame}   &   2014   &   PCD   &   real   &   1   & \redxmark   &   \redxmark  & \redxmark & \greencheck   \\
            semantic3D.net \cite{hackel2017semantic3d}  &   2017   &   PCD   &   real   &   2  & \redxmark &   \redxmark  & \redxmark & \greencheck  \\
            ArCH \cite{archDatasetPaper}  &   2020   &   PCD   &   real   &   1   & \redxmark &   \greencheck  & \redxmark & \greencheck  \\
            Toronto-3D \cite{tan2020toronto}  &   2020   &   PCD   &   real   &   1   & \redxmark &   \redxmark &  \redxmark & \greencheck  \\
            Whu-TLS \cite{dong2020registration}  &   2020   &   PCD   &   real   &   1   & \redxmark &   \redxmark  &  \redxmark  & \redxmark \\
            Paris-CARLA-3D \cite{deschaud2021pariscarla3d} &   2021   &   PCD   &   real/synthetic   &   2   & \redxmark &   \redxmark  & \redxmark & \greencheck  \\
            LOD3 Road Space Models \cite{ingolstadtLoD3} &   2021   &   NET/3DM   &   real   &   4   & \redxmark &   \greencheck  &  \greencheck & \greencheck  \\
            Hessigheim 3D \cite{hessigheim} & 2021   &   PCD/3DM   &   real   &   2   & \redxmark &   \greencheck  & \redxmark & \greencheck  \\
            SUM \cite{gaoSUM} & 2021 &   PCD/3DM   &   real   &   2   & \redxmark &   \greencheck  & \redxmark & \greencheck  \\
            KITTI-360 \cite{liao2021kitti}  &   2021   &   PCD   &   real   &   3   & \redxmark &   \redxmark  &  \redxmark  & \greencheck  \\
            UrbanScene3D \cite{lin2022capturingUrban3Dscene}  &   2022  &   PCD/IMG/3DM   &   synthetic   &   3   & \redxmark &   \greencheck  & \redxmark & \greencheck  \\
            Building3D \cite{wang2023building3d} &   2023  &   PCD/3DM   &   real   &   4   & \redxmark &   \greencheck  & \redxmark & \greencheck  \\
            SUD \cite{SUDdata_SiliviaGonzalez}  &   2023   &   PCD   &   real   &   1  & \redxmark &   \redxmark  &  \redxmark & \greencheck \\
            Building-PCC \cite{buildingpcc} &   2024  &   PCD/3DM   &   real   &  4   & \redxmark &   \greencheck  & \redxmark & \greencheck  \\      
            
            \hline
            \textbf{TUM2TWIN (ours)} &   2025   &   PCD/IMG/NET/3DM   &   real/synthetic   &   32   & \greencheck &   \greencheck  & \greencheck & \greencheck \\
            \bottomrule
        \end{tabular}
    \end{adjustbox}
\end{table*}


\section{The TUM2TWIN Benchmark Dataset}
\label{method}
\begin{figure*}
    \centering
    \includegraphics[width=\linewidth]{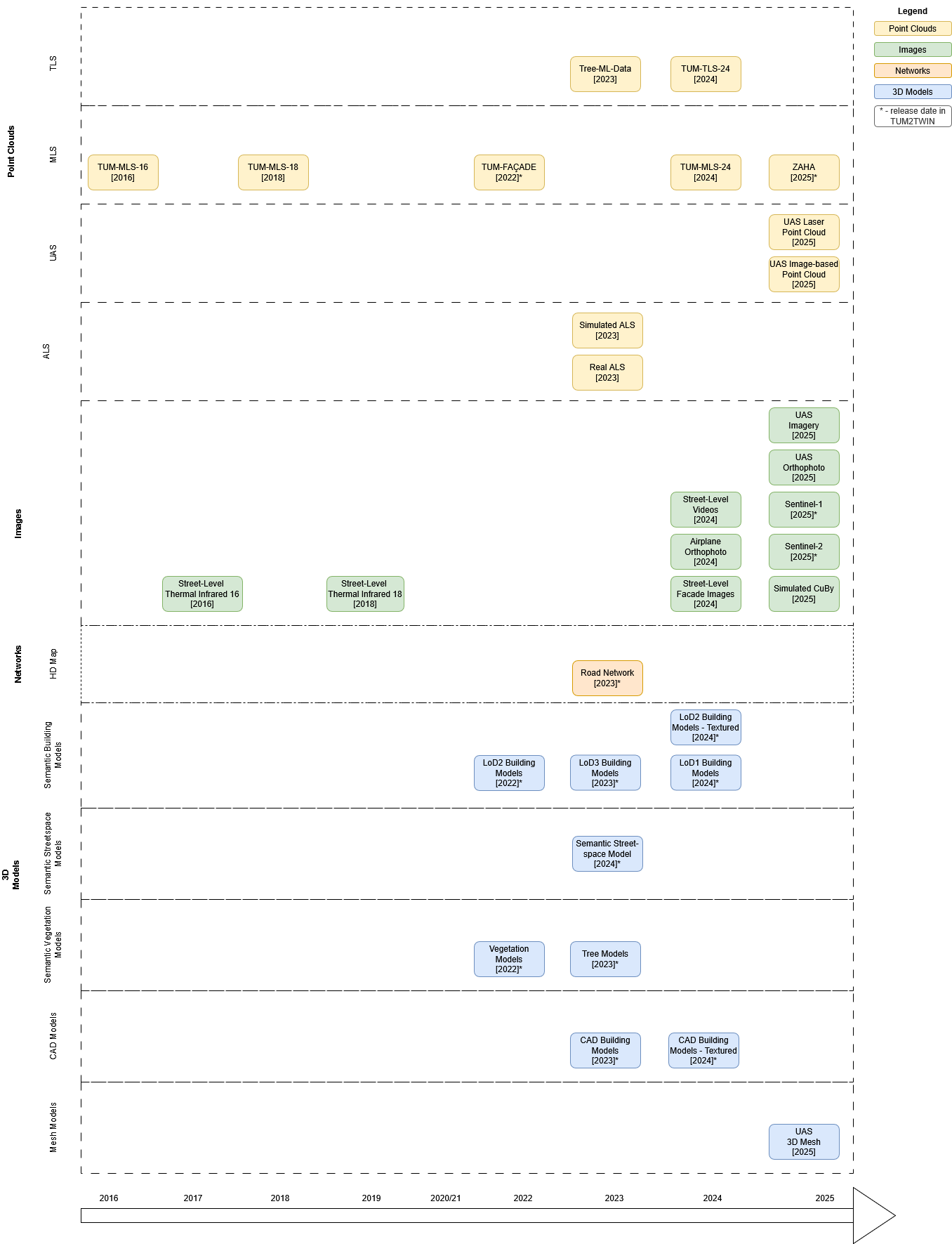}
    \caption{Timeline of the TUM2TWIN benchmark dataset. }
    \label{fig:dataTimeline}
\end{figure*}


%
\begin{figure*}
    \centering
    \includegraphics[width=0.9\linewidth]{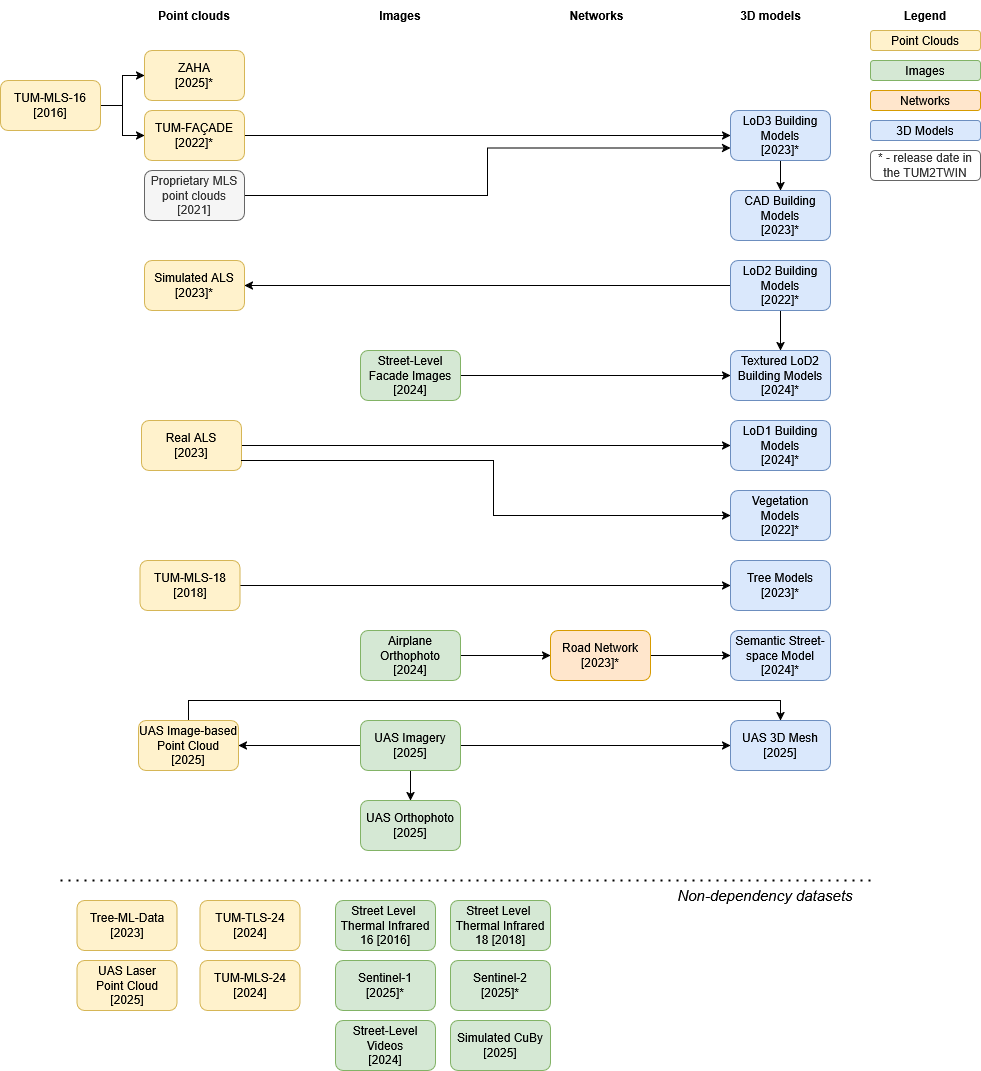}
    \caption{Data dependency graph presenting the relation of the TUM2TWIN sub-datasets. Each data is bound to the spatial and temporal accuracy of the source data. The release data can differ to the acquisition date as marked by asterisk ($*$). }
    \label{fig:dataDependency}
\end{figure*}

\begin{figure*}
    \centering
    \includegraphics[width=\linewidth]{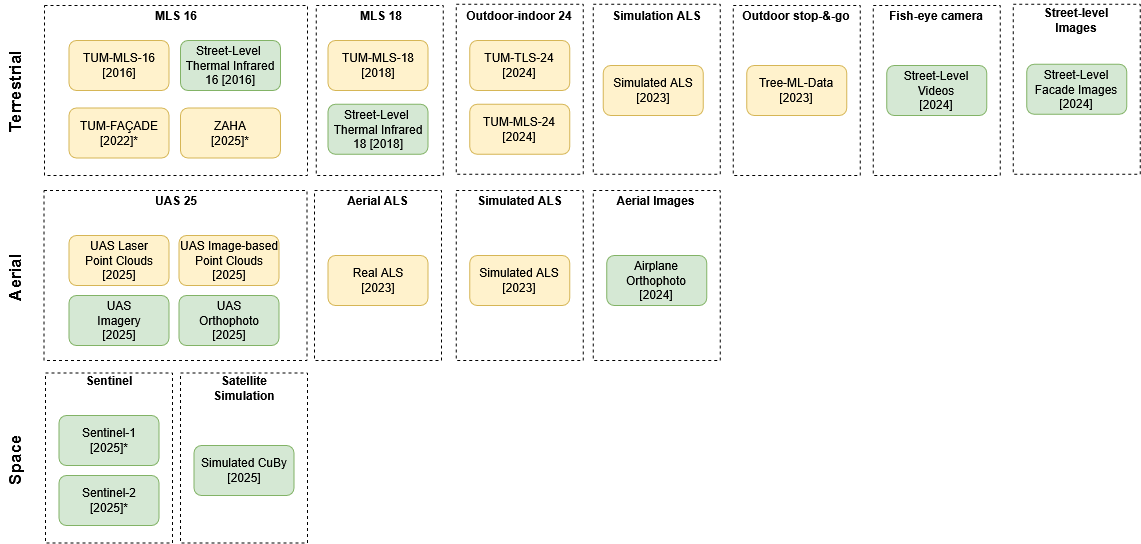}
    \caption{TUM2TWIN data acquisition campaigns (w/o derived high-level representations of Networks and 3D Models).  }
    \label{fig:campaigns}
\end{figure*}

In this contribution, we present TUM2TWIN, the benchmark dataset for Urban Digital Twins (UDTs) development (Figure~\ref{fig:MainOverview}).
We define four major pillars of such benchmark dataset: point clouds (Sec.~\ref{sec:pcl}), images (Sec.~\ref{sec:img}), networks (Sec.~\ref{sec:networks}), and 3D models (Sec.~\ref{sec:3Dmodels}), as visualized in Figure~\ref{fig:dataDependency}.
These are further divided into subsets, e.g., point cloud is a parent of Terrestrial Laser Scanning (TLS), which in turn is a parent of TUM-TLS-24 (Figure~\ref{fig:dataDependency}).
The crucial feature of each of the dataset is their georeferencing and assessed accuracy aspects, enabling their joint geometrical analysis (Table~\ref{tab:comparisonTable} and Table~\ref{tab:dataQuality}).
In essence, the georeferencing serve as unique and earth-global positional identifier enabling association of any object in the dataset via globally defined xyz coordinates in relation to the global coordinate reference system center. 
While timestamp supports multitemporal analysis (Figure~\ref{fig:dataTimeline}). 
Additional geometric, radiometric, and semantic features are listed in Table 2.
Also, we provide an overview of the source data acquisition campaigns in Figure~\ref{fig:campaigns} that comprises terrestrial, aerial, and space acquisition.
%
\begin{table*}[!ht]
    \captionsetup{size=footnotesize}
    \caption{TUM2TWIN Data Quality (PCD - Point clouds, IMG - Images, NET - Networks, 3DM - 3D Models, \text{*} - source data).} 
    \label{tab:dataQuality}
    \centering
    \begin{adjustbox}{width=\textwidth, center}
        \begin{tabular}{|l|l|l|l|l|l|l|l|l|}
        \hline
            Data & Data Type & Sensor & Platform & Acquisition & Release & Abs. Acc. [m] & Rel. Acc. [m] & Quantity [GB] \\ \hline
            TUM-TLS-24 & PCD & Leica P50/RTC360  & Tripod & 2024 & 2025 & 0.007 & 0.001 & 95 \\ \hline
            TreeML-Data  & PCD & Riegl VZ-400i & Car & 2023 & 2024 & 0.02 & 0.02 & 8 \\ \hline
            TUM-MLS-16 & PCD & Velodyne HDL-64E & Car & 2016 & 2016 & 0.5 & 0.2 &  60 \\ \hline
            TUM-MLS-18 & PCD & Velodyne HDL-64E & Car & 2018 & 2018 & 0.5  & 0.2 &  90  \\ \hline
            TUM-MLS-24 & PCD & FlexScan 22/BLK ARC & Backpack & 2024 & 2025 & 0.2  & 0.1 &  100  \\ \hline
            TUM-FAÇADE & PCD & Velodyne HDL-64E & Car & 2016 & 2021 & 0.5 & 0.2 &  15  \\ \hline
            ZAHA & PCD & Velodyne HDL-64E & Car & 2016 & 2025 & 0.5 & 0.2 &  22  \\ \hline
            UAS Laser Scan & PCD & Zenmuse L2 & UAS & 2024 & 2025 & 0.08 &  0.05 & 3 \\ \hline
            UAS Image-based Scan & PCD & Zenmuse L2 & UAS & 2024 & 2025 & 0.04 &  0.02 & 4 \\ \hline
            Simulated ALS & PCD & Leica HYPERION2+ (sim.) & Aeroplane (sim.) & 2022 & 2023 & 0.05  & - & 329  \\ \hline
            Real ALS & PCD & - & Aeroplane & 2022 & 2022 & 0.21 & 0.21 & 0.5 \\ \hline
            Street-Level Thermal Infrared 16 & IMG & Jenoptik IR-TCM 640 & Car & 2016 & 2017 & 0.5 & - & 8  \\ \hline
            Street-Level Thermal Infrared 18 & IMG & Jenoptik IR-TCM 640 & Car & 2018 & 2019 & 0.5 & - & 15  \\ \hline 
            Street-Level Video & IMG & GoPro Hero 11 & Handheld/Car & 2024 & 2024 & 1.0 & - & 1 \\ \hline
            Street-Level Facade Images & IMG & Sony $\alpha$7 & Handheld & 2024 & 2024 & 1.0 & - & 0.5 \\ \hline
            Airplane Orthophoto & IMG & - & Aeroplane & 2021 & 2021 & 0.2 & 0.2 & 0.4 \\ \hline
            UAS Orthophoto & IMG & Zenmuse L2 & UAS & 2024 & 2025 & 0.04 & 0.02 & 1.7 \\ \hline
            UAS Imagery & IMG & Zenmuse L2 & UAS & 2024 & 2025 & 0.02 & - & 6 \\ \hline
            Sentinel-1 & IMG &  C-band synthetic-aperture radar & Satellite & 2022 \& 2023 & 2025 & 5-40 &  5-40 & 1.6 \\ \hline
            Sentinel-2 & IMG & Multispectral (13 band) imager & Satellite & 2022 \& 2023 & 2025 & 10-60 &  10-60 & 2.4 \\ \hline
            CuBy Simulated Satellite Image & IMG & Artificial 580 mm Linescanner & Satellite (sim.) & - & 2025 & 4.0 &  - & - \\ \hline 
            Road Network & NET & -/- & Aeroplane\text{*} & 2021 & 2023 & 0.2 &  0.2 &  0.03 \\ \hline
            LoD1 Building Models & 3DM & -/- & Footprint/Aeroplane\text{*} & 2024 & 2024 & 0.02 &  0.83 & 0.03 \\ \hline
            LoD2 Building Models & 3DM & -/- & Footprint/Aeroplane\text{*} & - & 2022 & 0.02 &  0.2 & 0.3 \\ \hline
            LoD2 Textured Building Models & 3DM & -/Sony $\alpha$7\text{*} & LoD2BM/Handheld\text{*} & - & 2024 & 0.02 &  0.2 & 0.1 \\ \hline
            LoD3 Building Models & 3DM & -/Velodyne HDL-64E \& MoSES\text{*} & LoD2BM/Car\text{*} & - & 2024 & 0.02 & 0.05 & 0.2 \\ \hline
            Semantic Streetspace Model & 3DM & -/- & Road Network\text{*} & 2023 & 2024 & 0.2 & 0.2 & 0.5 \\ \hline
            Vegetation Models & 3DM & -/- & Aeroplane\text{*} & - & 2022 & 0.21 &  0.09 & 0.6 \\ \hline
            Tree Models & 3DM & Velodyne HDL-64E\text{*} & Car\text{*} & 2018 & 2023 & 0.5 &  0.2 & 0.03 \\ \hline
            CAD Building Models & 3DM & -/Velodyne HDL-64E \& MoSES\text{*} & Footprint/Aeroplane\text{*} & - & 2024 & 0.02 & 0.05 & 0.1 \\ \hline
            Textured CAD Building Models & 3DM & -/Sony $\alpha$7\text{*} & LoD2BM/Handheld\text{*} & - & 2024 & 0.02 &  0.2 & 0.1 \\ \hline
            UAS 3D Mesh & 3DM & Zenmuse L2\text{*} & UAS\text{*} & 2024 & 2025 & 0.04 &  0.02 & 0.1 \\ \hline
        \end{tabular}
    \end{adjustbox}
\end{table*}
%
%
\begin{table*}[!ht]
    \label{tab:dq}
    \setlength\tabcolsep{3.5pt} 
    \scriptsize
    \centering
        \caption{Data Features (I - Intensity, RGB - Optical spectrum, SAR - Synthetic Aperture Radar, MULT - Multispectral)}
    \begin{tabular}{|l|l|l|l|}
    \hline
        Data & Geometry & Radiometry & Semantics \\ \hline
        TUM-TLS-24 & 3D & I, RGB & - \\ \hline
        TreeML-Data & 3D & I, RGB & \greencheck \\ \hline
        TUM-MLS-16 & 3D & I & \greencheck \\ \hline
        TUM-MLS-18 & 3D & I & \greencheck \\ \hline
        TUM-MLS-24 & 3D & I, RGB & - \\ \hline
        TUM-FA\c{C}ADE & 3D & - & \greencheck \\ \hline
        ZAHA & 3D & - & \greencheck \\ \hline
        UAS Laser Scan & 3D & I, RGB & - \\ \hline
        UAS Image-based Scan & 3D & I, RGB & - \\ \hline
        Simulated ALS & 3D & - & - \\ \hline
        Real ALS & 3D & I, RGB & \greencheck \\ \hline
        Street-Level Thermal Infrared 16 & 2D & INF & - \\ \hline
        Street-Level Thermal Infrared 18 & 2D & INF & - \\ \hline
        Street-Level Video & 2D & RGB & - \\ \hline
        Street-Level Facade Images & 2D & RGB & - \\ \hline
        Airplane Orthophoto & 2D & RGB & - \\ \hline
        UAS Orthophoto & 2D & RGB & - \\ \hline
        UAS Imagery & 2D & RGB & - \\ \hline
        Sentinel-1 & 2D & SAR & - \\ \hline
        Sentinel-2 & 2D & MULT & - \\ \hline
        Simulated CuBy & 2D & RGB & - \\ \hline
        Road Network & 3D & - & \greencheck \\ \hline
        LoD1 Building Models & 3D & - & \greencheck \\ \hline
        LoD2 Building Models & 3D & - & \greencheck \\ \hline
        LoD2 Textured Building Models & 3D & RGB & \greencheck \\ \hline
        LoD3 Building Models & 3D & - & \greencheck \\ \hline
        Semantic Streetspace Model & 3D & - & \greencheck \\ \hline
        Vegetation Models & 3D & - & - \\ \hline
        Tree Models & 3D & - & - \\ \hline
        CAD Building Models & 3D & - & - \\ \hline
        Textured CAD Building Models & 3D & RGB & - \\ \hline
        UAS 3D Mesh  & 3D & RGB & - \\ \hline        
    \end{tabular}
\end{table*}

\subsection{\textbf{Point Clouds}}
\label{sec:pcl}

We define a point cloud as a collection of data points in 3D space, where each point represents a location on the surface of an object or environment and can be extended by additional scalar values representing, e.g., color or intensity. 
They serve as the raw data for creating detailed digital representations of real-world objects, which can be processed into 3D models or meshes for further use.
We differentiate between \gls{TLS}, \gls{MLS}, \gls{UAS}, and \gls{ALS} point clouds. 
\gls{TLS} pertains to static terrestrial laser scanning offering the highest accuracy in our dataset; \gls{MLS} refers to scanners mounted on a backpack or a vehicle; \gls{UAS} to laser scanners mounted on a flying \gls{UAS} platfrom; while \gls{ALS} describes airborne acquisition. 
Point cloud is defined:
\[
P = \{p_i = (x_i, y_i, z_i, a_1, a_2, ..., a_m) \in \mathbb{R}^3, \forall i \in \{1, 2, ..., N\} \}
\]

where:
\begin{itemize}
    \item \( P \) is the point cloud, a set of \( N \) 3D points.
    \item \( p_i \) represents an individual point.
    \item \( (x_i, y_i, z_i) \) are the coordinates of the point in 3D space.
    \item  \( a_1, a_2, ..., a_m \) represent additional properties such as RGB color values, reflectance, or surface normals.
\end{itemize}

\subsubsection{\textbf{TLS}}
\label{sec:tls}
\begin{figure*}
\centering \includegraphics[width=\linewidth]{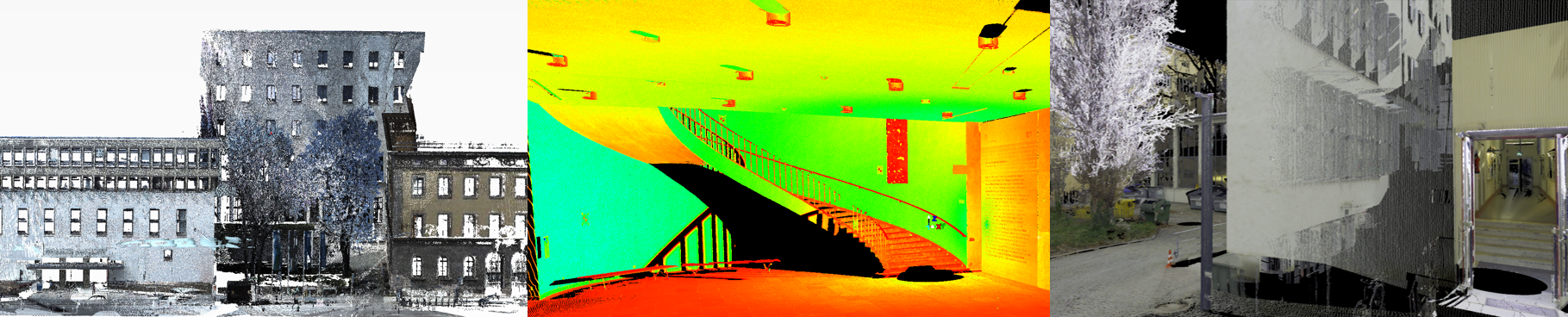}
\caption{Partial outdoor and indoor example scenarios included in TUM-TLS-24. The left is an outdoor scene colored with RGB, and the middle is an indoor scene colored with intensity, the right part shows an example of the transition of outdoor-to-indoor.}
\label{fig:TUM-TLS-24}
\end{figure*}

%
\noindent\textbf{TUM-TLS-24}  
Maintaining a consistent frame of reference for all data is essential for fully exploiting the potential of the TUM2TWIN dataset.
Within the TUM2TWIN dataset, TUM-TLS-24 represents the most accurate and highest-density georeferenced point cloud data available, thereby providing a strong foundation for integrating and fusing diverse data types (Fig.~\ref{fig:TUM-TLS-24}).

TUM-TLS-24 is a high-precision terrestrial laser scanning (TLS) point cloud dataset encompassing indoor and outdoor environments. Data are acquired in April 2024 using a Leica ScanStation P50 and a Leica RTC360. The data collection is conducted on a station-by-station basis and comprises a total of 99 acquisition points (stations).
The dataset comprehensively covers most of the main building's indoor and outdoor public spaces at TUM main campus. Moreover, over 100 black and white (B\&W) targets are deployed during acquisition to facilitate precise registration and georeferencing.
Ultimately, the mean absolute error (MAE) of all scans after registration is 1.2 mm (relative accuracy), while the MAE following georeferencing based on ground control points is 7.1 mm (absolute accuracy).

\noindent\textbf{TreeML-Data}
Recent advancements in remote sensing techniques expand opportunities for studying urban trees, which play a crucial role in promoting human health and well-being in cities. To support these research efforts, the TreeML-Data dataset \cite{yazdi2024multilayered} is compiled. This dataset is collected using~\gls{TLS} in a stop-and-go manner in January 2023. It comprises labeled point clouds from 40 scanning projects conducted along streets in Munich, including Arcisstrasse, in front of the TUM main campus. The point clouds are classified into three categories: “Tree,” “Building,” and “Other.”

\begin{figure*}
    \centering
    \includegraphics[width=0.7\linewidth]{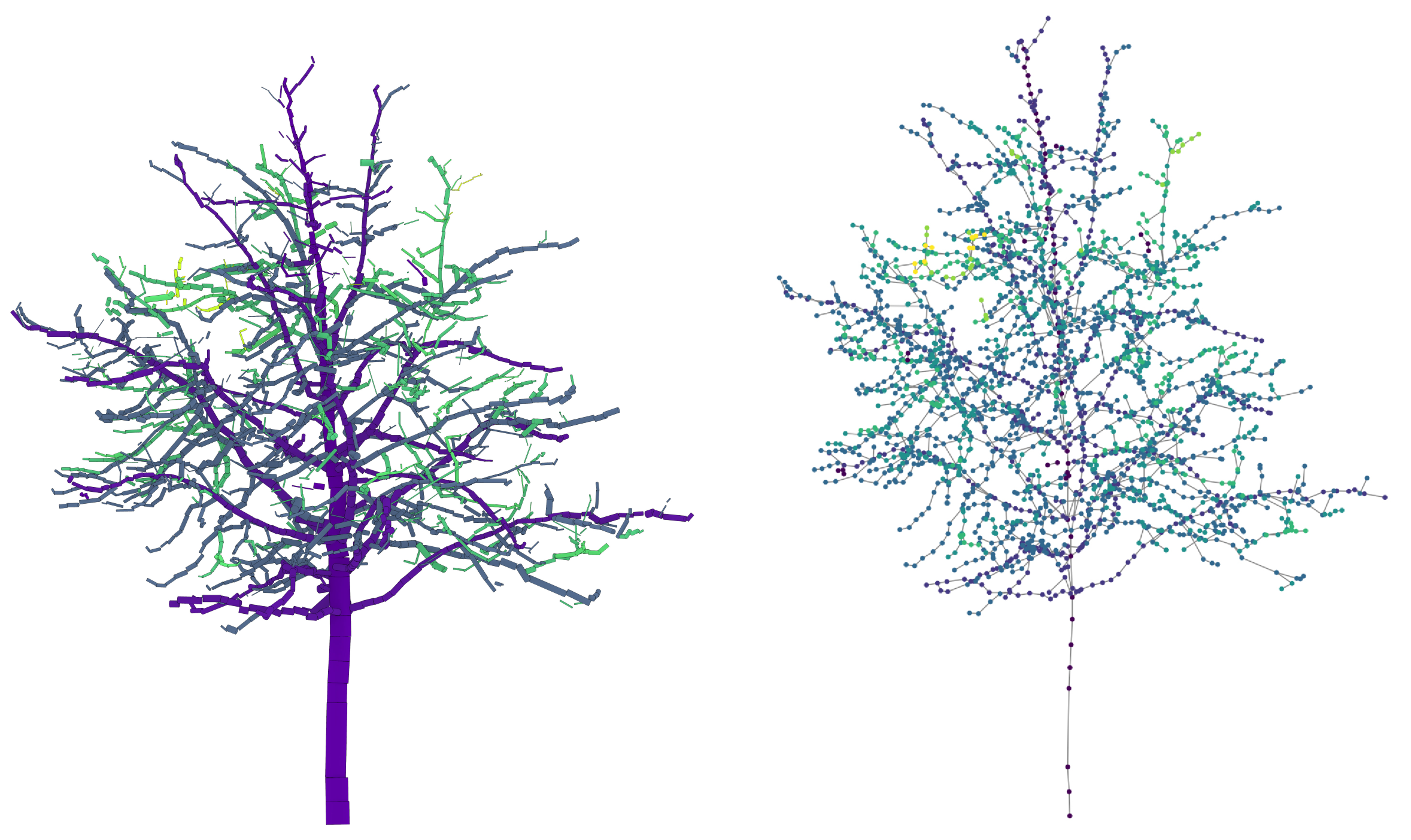}
    \caption{A tree example from the TReeML-Data, presented in two different representations: The quantitative structure model (QSM) (left) and the graph structure model (right).}
    \label{fig:TreeML-Data_2}
\end{figure*}

Additionally, the trees in the dataset are manually isolated. The dataset primarily focuses on urban trees, featuring 3,755 leaf-off point clouds (captured during winter), quantitative structure models (QSM) \cite{raumonen2013fast}, tree structure measurements, and tree graph structure models (Figure~\ref{fig:TreeML-Data_2}).

The data are collected using the Riegl VZ-400i TLS laser scanner. To ensure precise georeferencing, a Leica Zeno FLX100 high-precision GPS antenna is mounted on the laser scanner, capturing the global location of each scan. The scanner is initially configured to the “Panorama40” resolution (40 mdeg). GNSS records indicate that the positional accuracy is approximately 2 cm in most cases \cite{yazdi2023treeml}.

\subsubsection{ \textbf{MLS} }
\label{sec:mls}
%

\noindent\textbf{TUM-MLS-16} 
TUM-MLS 2016 \cite{zhu_tum-mls-2016_2020} is a large-scale annotated dataset based on mobile laser scanning point clouds acquired at the city campus of the Technical University of Munich. The MLS data are acquired in April 2016 by Fraunhofer IOSB with the MODISSA (Mobile Distributed Situation Awareness) mobile sensor platform, which is used for hardware evaluation and software development in the contexts of automotive safety and security applications. The system is equipped with two Velodyne HDL-64E LiDAR sensors above the windshield. Both laser scanners are positioned on wedges at a $25^\circ$ angle to the horizontal, rotated outwards at a $45^\circ$ angle, which prevents measurements of the vehicle’s roof and still guarantees good coverage of the roadway in front and to the sides of the vehicle. Therefore, the facades of buildings are captured in their entire height.
The scanned point clouds are directly georeferenced while driving along the roads around the TUM city campus and the inner yard. This covers an urban scenario consisting of building facades, trees, bushes, parked vehicles, wedges, roads, grass, and so on. Each point has 3D x-, y-, and z-coordinates and intensities of the laser reflectance.
All the measured points in the scene are manually labeled with eight semantic classes following the ETH standard \cite{hackel2017semantic3d}.

\noindent\textbf{TUM-MLS-18} 
TUM-MLS 2018 is acquired on December 19th 2018 by Fraunhofer ISOB with MODISSA with similar sensor compositions.
Compared to TUM-MLS 2016, the offset for each scans are optimized globaly and provided as the reference.
The annotations are the same as the TUM-MLS2016 dataset, while the primary purpose is the epoch-wise change detection owing to two-years acquisition difference \cite{zhu_tum-mls-2016_2020}.

\begin{figure}
    \centering
    \includegraphics[width=0.6\linewidth]{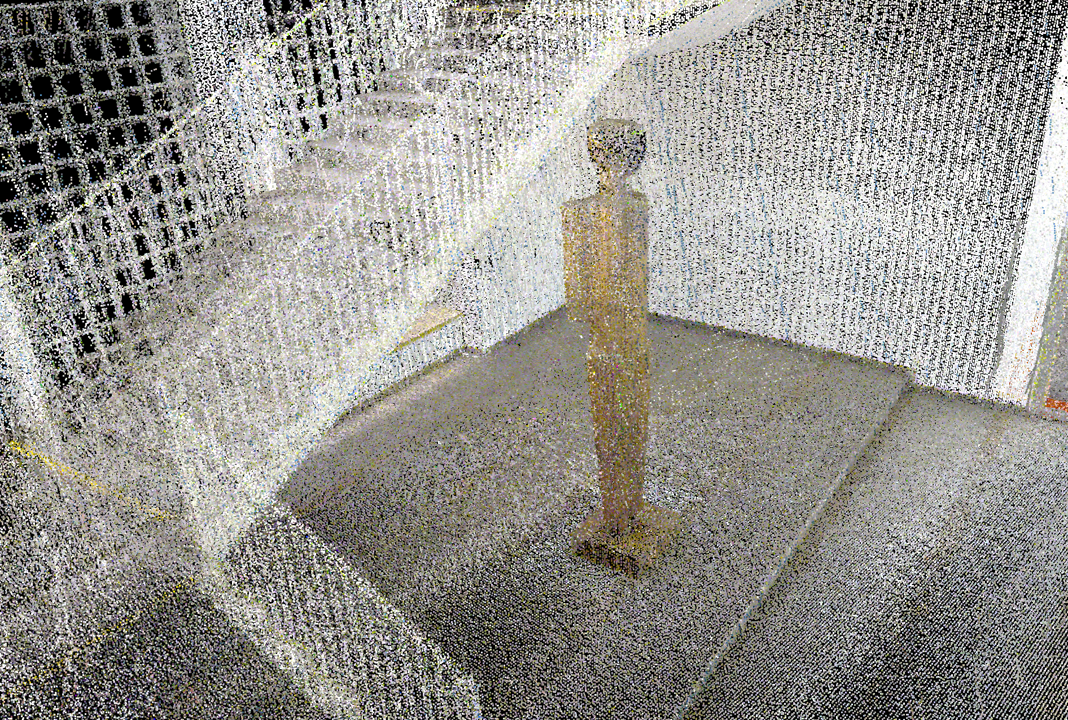}
    \caption{Colored point cloud of an indoor scene from TUM-MLS-24.}
    \label{fig:TUM_MLS_24}
\end{figure}

\noindent\textbf{TUM-MLS-24} 
TUM-MLS-2024 is an indoor~\gls{MLS} dataset acquired at TUM main campus in December 2024 (Fig.~\ref{fig:TUM_MLS_24}).
We use both the Leica BLK ARC and Z+F FlexScan 22 to capture point clouds in parallel to the TUM-TLS-24 campaign.
The objective is to assess the quality of each mobile system using static TUM-TLS-24 as the ground truth. The Leica BLK ARC provides a 360° horizontal and 270° vertical field of view, with a range from 0.5 m to 25 m and a measurement rate of 420k pts/sec, achieving ±10 mm accuracy indoors.
It features a four-camera system: a 13 MP high-resolution camera and three panoramic cameras. The Z+F FlexScan 22 incorporates a two-camera system with 20 MP sensors and utilizes the Z+F Imager 5016A laser scanner.
This sensor operates in profile scanning mode, capturing 550k pts/sec with a resolution of 10k pixels per profile and a minimum range of 0.6 m. Both systems are used in backpack mode. The dataset from the Leica BLK ARC is processed using Leica Cyclone Registration 360 Plus, while Z+F Laser Control is used for the FlexScan 22.

\noindent\textbf{ZAHA}
In ZAHA \cite{wysocki2024zahaintroducinglevelfacade}, we introduce the Level of Facade Generalization (LoFG): A novel hierarchical classification system based on international urban modeling standards. This approach ensures compatibility with real-world architectural challenges while enabling a standardized comparison of segmentation methods.  

\begin{figure*}[ht]
    \centering
    \includegraphics[width=0.9\linewidth]{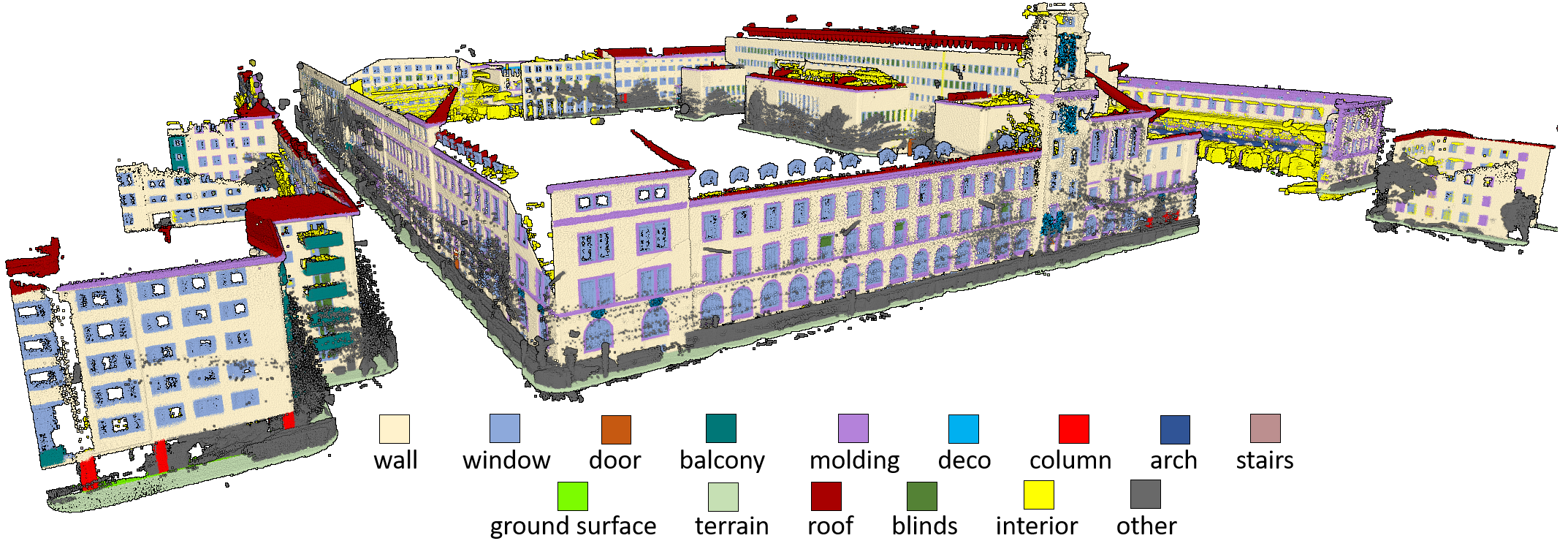}
    \caption{To date, the largest facade semantic segmentation dataset: ZAHA.}
    \label{fig:zaha}
\end{figure*}

As part of this effort, we present the largest 3D facade semantic segmentation dataset to date, comprising 601 million annotated points across five classes in LoFG2 and 15 classes in LoFG3 (Figure~\ref{fig:zaha}). 
Additionally, we evaluate baseline semantic segmentation methods on these LoFG classes and datasets, offering insights into the remaining challenges in facade segmentation.
The point cloud coordinates stem from the TUM-MLS-2016, showing re-purposing capabilities of the collected data.
We are convinced ZAHA will contribute to the advancement of 3D facade segmentation methods, playing a crucial role in the development of robust urban digital twins.

\noindent\textbf{TUM-FAÇADE}
The predecessor of ZAHA is the TUM-FAÇADE dataset, which is an extension of the TUM-MLS-2016 point clouds, enhancing the original data with 17 facade-level semantic classes. 
It consists of 17 annotated and 12 non-annotated facades, totaling 118 million georeferenced and labeled facade-level points.  
The classes definitions has been updated and hierarchically represented in the ZAHA dataset \cite{wysocki2024zahaintroducinglevelfacade,tumfacadePaper}.

\subsubsection{ \textbf{UAS}}\label{sec:uas}
\begin{figure*}[ht]
    \centering
    \includegraphics[width=1.0\linewidth]{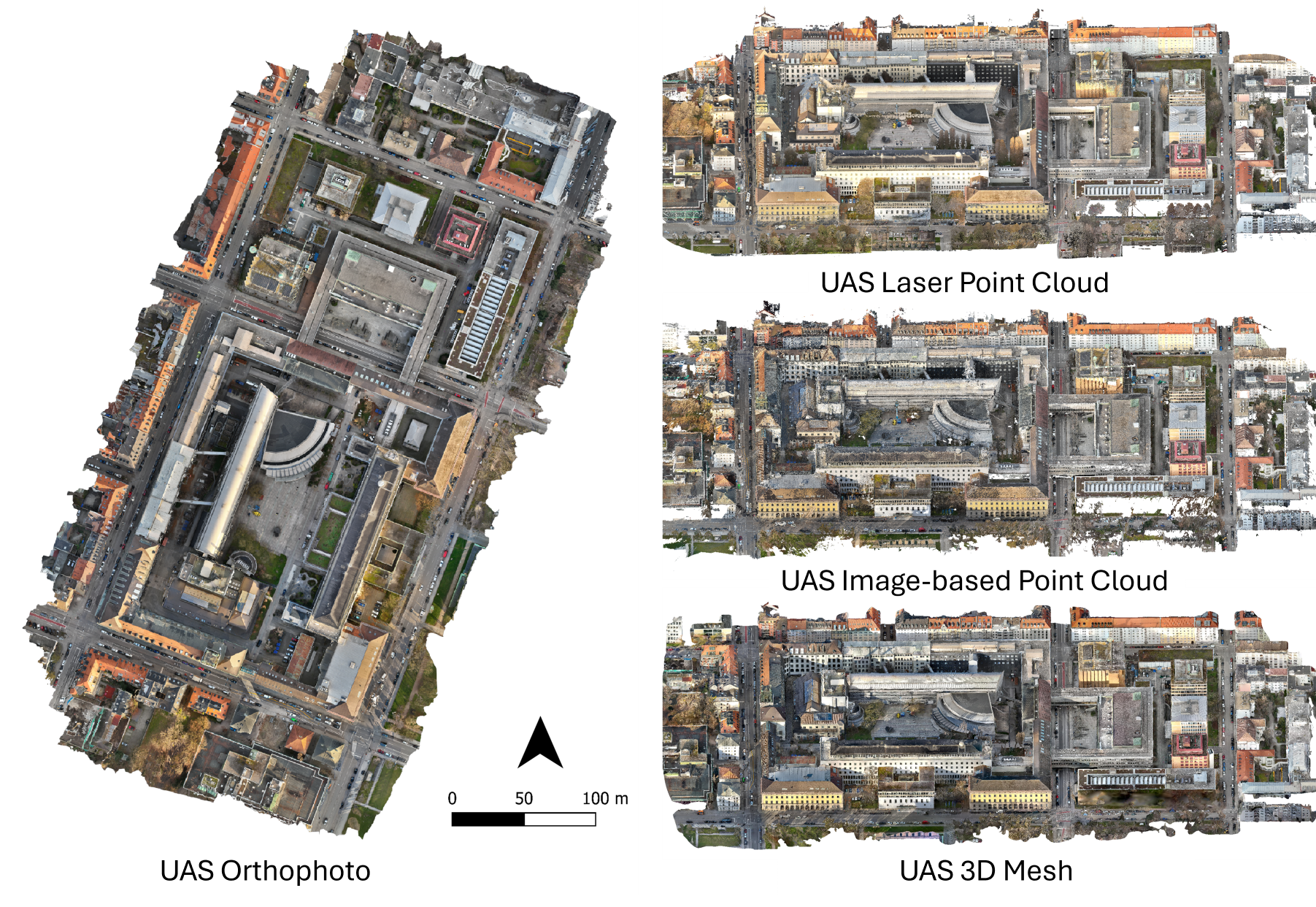}
    \caption{Overview of UAS data products of TUM main campus, including UAS Orthophoto, UAS Laser Point Cloud, UAS Image-based Point Cloud, and UAS Mesh. Additionally, we provide RTK-based trajectories and raw UAS Imagery.}
    \label{fig:uas}
\end{figure*}

\noindent\textbf{UAS Laser Point Cloud}
\gls{UAS} acquisition in December 2024 is used to capture the TUM city campus with laser scanning.
A DJI Matrice 350~\gls{RTK} is used with a Zenmuse L2 laser scanner and integrated RGB camera for point cloud coloring. 
Laser scanning performs direct high-accuracy measurements (with a mean cloud-to-cloud distance of 0.08 m compared to manually measured GNSS points from the scene) at high spatial resolution (with an average point density of 1,338 pts/m$^2$) (Fig.~\ref{fig:uas}) \cite{anders_uas_tum_downtown}.

\noindent\textbf{UAS Image-based Point Cloud}
During the same acquisition campaign as UAS Laser Point Cloud, the images have been captured.
The UAS acquisition allows capturing the building topography and its surroundings from an aerial perspective with close-range high-resolution observations compared to airplane or satellite platforms. 
Structure from Motion (SfM) is applied to reconstruct sparse 3D scenes and estimate camera poses from 2D images by detecting and matching features across overlapped images \cite{pix4dmatic}. This is followed by Multi-View Stereo (MVS), which considers multiple images of the same scene and searches for a complete and dense 3D structure as a point cloud \cite{schoenberger2016mvs}. The UAS Image-based Scan provides high 3D reconstruction accuracy (with a 0.04 m georeferencing accuracy) when images are well-textured \cite{anders_uas_tum_downtown}.
%
\subsubsection{ \textbf{ALS}}
\label{sec:als}
We distinguish two types of aerial laser scanning (ALS). \textit{Real} ALS pertains to actual surveying campaign in the real-world scenarios, while \textit{Simulated} to synthetically generated point clouds in a virtual testbed representing the actual TUM2TWIN location.

\noindent\textbf{Real ALS} The aerial laser scanning data stems from the State Office for Digitalization, Broadband and Surveying (ger. LDBV) and covers not only the TUM2TWIN area, but also the whole state of Bavaria, Germany.
For the TUM2TWIN area, its absolute accuracy is approximately 0.3 m (horizontal) and 0.12 m (vertical) with density of minimum 4 pts / $m^2$.
The data comprises also eight semantic clasess: Unclassified, Ground, Building, Water body, Object (e.g., vegetation), Bridge and basement exit, Synthetic/derived ground (e.g., terrestrial image), Synthetic/derived ground (from from a digital terrain model).
The data is regularly updated with the last update in 2022 for the TUM2TWIN area \cite{LDBVals}.

\noindent\textbf{Simulated ALS}
This synthetic~\gls{ALS} dataset is created by using the Helios++ \cite{winiwarter2022virtual} simulation toolkit to emulate aerial laser scanning, with the goal of bridging the gap between idealized models and the inherent imperfections of real-world measurements. 
The high-quality LoD2 building models from Bavaria, Germany (Section \ref{sec:semanticBldMod}), are chosen for their detailed roof representation and high global accuracy (Table~\ref{tab:dataQuality}).
In the simulation process, realistic artifacts such as sensor noise and inter-building occlusions are intentionally introduced to mirror the challenges found in actual data. 
The virtual sensor is designed to closely mimic the Leica HYPERION2+, featuring an oscillating optics system operating at a pulse frequency of 1.5 MHz and a scan frequency of 150 Hz. The simulated survey replicates an aerial mission executed by a Cirrus SR22 aircraft flying at an altitude of 400 meters with a 160-meter strip interval, ensuring a realistic scanning pattern. The training dataset comprises 281,571 buildings from Munich, totaling over 6.5 billion points, and averaging about 22,406 points per building, with an additional 10,000 buildings reserved for evaluation. 
This meticulously designed dataset serves as a robust platform for training and testing the model under conditions that closely approximate real-world laser scanning scenarios.

\subsection{\textbf{Images}}
\label{sec:img}
%
An image is a 2D representation of visual information, composed of pixels on a grid, where each pixel has color or intensity values. 
Images are usually captured using cameras of various kinds, e.g., monochrome, RGB, multispectral, or hyperspectral cameras. 
We differentiate between street-level images and videos of different spectrum, and above-ground acquisition from \gls{UAS}, airplanes, and satellites. 
Mathematically, an image is a function that maps pixel coordinates to intensity or color values:
\[
I : \mathbb{Z}^2 \to \mathbb{R}^c
\]

where:
\begin{itemize}
    \item \( I(x, y) \) represents the intensity or color value at pixel coordinates \( (x, y) \).
    \item \( \mathbb{Z}^2 \) denotes the discrete 2D grid of pixel positions.
    \item \( \mathbb{R}^c \) represents the color space, i.e.,:
    \begin{itemize}
        \item \( c = 1 \) for monochromatic grayscale images, where \( I(x, y) \in \mathbb{R} \).
        \item \( c = 3 \) for RGB images, where \( I(x, y) = (R, G, B) \in \mathbb{R}^3 \).
        \item \( c =  3 < n \leq 12 \) for multispectral images, where \( I(x, y) = (n) \in \mathbb{R}^{n} \).
        \item \( c =  12 < n \leq 300 \) for hyperspectral images, where \( I(x, y) = (n) \in \mathbb{R}^{n} \).
    \end{itemize}
\end{itemize}

\subsubsection{ \textbf{Street-Level Images}}
\label{sec:streetImages}

\noindent\textbf{Street-Level Videos}
This dataset is collected using a GoPro Hero11 monocular camera and is intended for street-view reconstruction.
The recordings take place in two distinct locations, comprising five video sequences capturing street-view traffic conditions from roads around TUM Downtown Campus (Gabelsbergerstr., Arcisstr., Luisenstr., and Theresienstr.).
Four of these sequences are recorded with the cameras mounted on a vehicle, while one is handheld, featuring dense traffic, diverse buildings, and pedestrian activity.

Unlike other officially released autonomous driving datasets, all recordings in this study are captured exclusively with GoPro Hero11 cameras, known for their simplicity and accessibility.
Notably, this dataset does not provide precisely calibrated internal camera parameters.
The inclusion of various capture modes and different scene conditions enhances our understanding of the challenges encountered by onboard cameras in moving vehicles.
Additionally, one sequence is captured on highway roads at Latitude: 35.08539, Longitude: -106.73099 (WGS84), chosen for its complex traffic dynamics, minimal building features, and high speeds to simulate real-world driving conditions.

\noindent\textbf{Street-Level Facade Images}
The perspective terrestrial RGB images, manually projected and captured using the Sony $\alpha$7 camera, are specifically acquired to validate automatic texturing processes.
The image acquisition campaign is planned to cover the facades of the building models with the fewest possible photos per triangle, ensuring consistent texture quality without the need for additional image stitching.
Additionally, panoramic images from Google Street View \cite{googleStreetView} are collected for testing automatic image projections on the building models \cite{tang2025texture2lod3}.

\subsubsection{ \textbf{Thermal Infrared}}
\label{sec:thermalInf}

\begin{figure}
    \centering
    \includegraphics[width=0.5\linewidth]{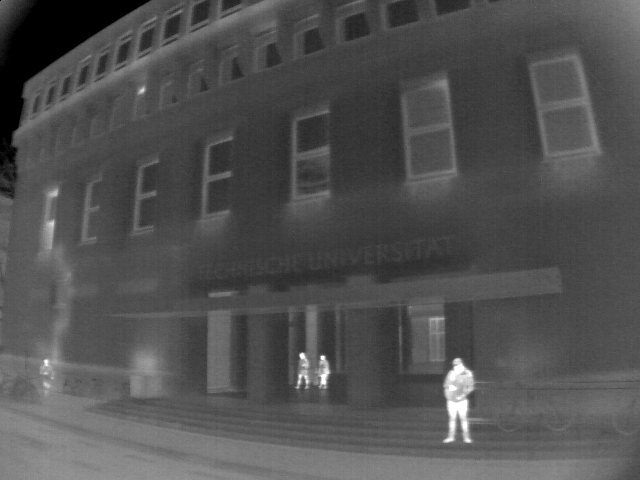}
    \caption{Example of TIR image of Street-Level Thermal Infrared 16.}
    \label{fig:fig_tir}
\end{figure}

\noindent\textbf{Street-Level Thermal Infrared 16}
Thermal infrared image sequences are recorded as part of the TUM-MLS 2016 acquisition campaign. The uncooled thermal camera used is a Jenoptik IR-TCM 640 microbolometer, with a field of view of $65.2^\circ \times 51.3^\circ$, mounted perpendicularly to the driving direction. This setup captures the facades of buildings (Figure~\ref{fig:fig_tir}). TIR images are provided as 16-bit TIFF files with lossless LZW compression, at a resolution of $640 \times 480$ pixels. The image positions are estimated using GPS-based vehicle positions, interpolated to match image timestamps, and are geometrically calibrated as described in \cite{zhu2023}.

\noindent\textbf{Street-Level Thermal Infrared 18}
The same approach as for \textit{Street-Level Thermal Infrared 16} is applied for the next epoch of thermal image acquisition.
This time the images are acquired simulatenously with the TUM-MLS 2018 campaign.

\subsubsection{ \textbf{UAS Imagery}} \label{sec:uas_photo}

\noindent\textbf{UAS Imagery}
The UAS images are simultaneously acquired with the UAS laser scanning (Sec.~\ref{sec:uas}). During the survey, 1104 images with a resolution of 5,280 $\times$ 3,956 pixels are taken by a Zenmuse L2 integrated RGM mapping camera (4/3 CMOS) mounted on a DJI Matrice 350~\gls{RTK} drone. 
The data is collected in nadir mode using automatic flight missions at an altitude of 75 m agl. Additional acquisitions are collected in manually operated flights in oblique view to capture the facades of the inner and outer campus area. The average ground sampling distance (GSD) of images is 1.6 cm, and these images are georeferenced through the~\gls{RTK} GNSS measurements and IMU of the UAS system. 
Images including nadir and oblique acquisition geometry are filtered by removing recognizable persons and license plates, finally totaling a repository of 962 images.
These images can be used for coloring laser scans, orthophoto generation, and photogrammetric reconstruction \cite{anders_uas_tum_downtown}.

\subsubsection{ \textbf{Aerial Orthophoto}} \label{sec:aerial_orthophoto}

\noindent\textbf{UAS Orthophoto}
A UAS Orthophoto (Figure~\ref{fig:uas}) covering the entire campus is generated based on the UAS images (Sec.~\ref{sec:uas_photo}) and the digital surface model (DSM) derived from the photogrammetric reconstruction using the automatic workflow in Pix4Dmatic v1.68 \cite{pix4dmatic}. The ground pixel resolution of the orthophoto is 1.6 cm according to the average GSD of UAS images, and the full dimensions are 26,507 $\times$ 35,407 $\times$ 4 bands (RGB and alpha). The coordinate reference system (CRS) is WGS 84 / UTM Zone 32N.

\noindent\textbf{Airplane Orthophoto} High-resolution nadir images are captured from altitudes exceeding 5,000 meters and are ortho-rectified achieving the ground sampling distance of 20 cm. 
Since 2017, aerial surveys have been conducted annually over approximately, which serve as the foundation for all aerial imagery products provided by the State Office for Digitalization, Broadband, and Surveying (LDBV).
The designated areas are covered in overlapping flight strips, which reduce the presence of shadowed or hidden zones.
Survey flights alternate biennially between Northern and Southern Bavaria, based on predefined planning regions \cite{LDBVorthophoto}.

\subsubsection{ \textbf{Satellite Observations}} 

Access to open data greatly expands the pool of remote sensing data users. Adding this aspect to TUM2TWIN, Sentinel-1 and Sentinel-2 data are included into the selection. The main advantage of the missions is related to continuous imaging and the opportunity to generate data stacks over time. At this point, the question arises what information Sentinel data can provide for the TUM campus, as the area is only covered by a small number of image pixels. However, reasoning gets interesting when thinking the other way around. The content in TUM2TWIN provides a comprehensive geometric and semantic understanding of scene compositions covered by individual pixels of the satellite image. Following this idea of data fusion, the combination of remote sensing data and highly detailed scene information opens interesting directions to follow.


\noindent\textbf{Sentinel-1}
The Sentinel-1 mission \cite{Torres2017} comprises three satellites, Sentinel-1A, Sentinel-1B (not active any more), and Sentinel-1C, launched in 2014, 2016, and 2024, respectively. Their primary objective is to acquire information about the physical properties of ground objects, e.g. salient signatures related to man-made structures of signatures, moderate diffuse signal responses at surfaces with noticable roughness, or dark areas in the image related to flat, smooth surfaces, e.g. lakes. The SAR principle also allows for interferometry in order to geometrically describe the shape of the earth surface and identify ground movement. Signals are emitted and detected in C-band with dual polarization. The revisit time for each satellite is 12 days. Imaging is conducted continuously on the global scale. The Sentinel-1 satellites offer four image modes, each with different spatial resolutions and coverage capacities. Depending on the selected mode, the spatial resolutions can reach up to 5x5 meters, the swatch width can be up to 410 kilometers. 

The TUM2TWIN collection contains quarterly georeferenced Sentinel-1 data sets in standard interferometric wide swath mode over the years 2022 and 2023 with a spatial resolution of 5 x 20 m and a swath width of 250 km. The selection can be expanded with available data from the archive. 
The approximate size of data acquisitions is 1.6 GB.


\begin{figure}[tbh]
	\centering
	\subfloat[]{\label{fig:S1_Q1_2022}\includegraphics[width=0.45\linewidth]{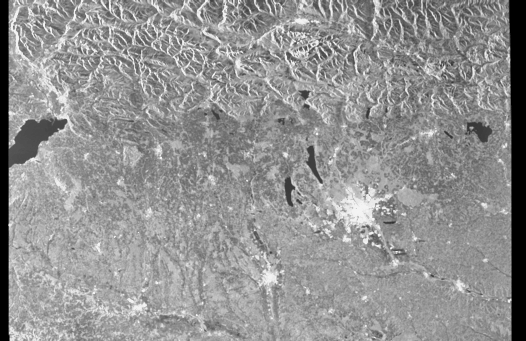}} 
	\hspace{1mm}
	\subfloat[]{\label{fig:S1_Q2_2022}\includegraphics[width=0.45\linewidth]{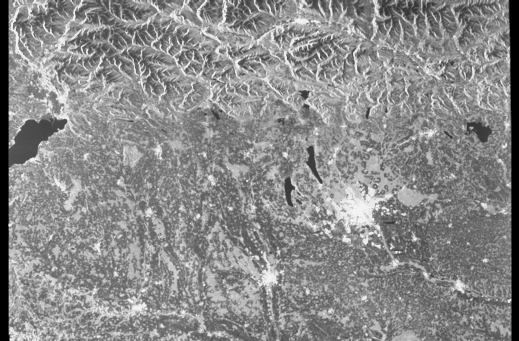}} \\
    \vspace{1mm}
	\subfloat[]{\label{fig:S1_Q3_2022}\includegraphics[width=0.45\linewidth]{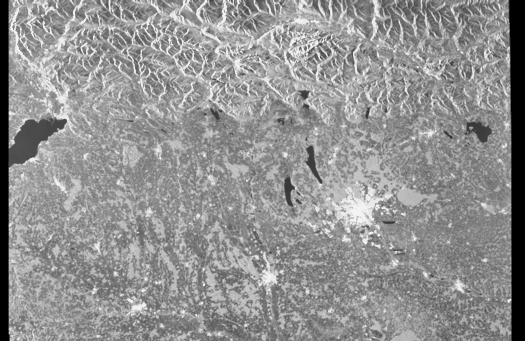}} 
    \hspace{1mm}
	\subfloat[]{\label{fig:S1_Q4_2022}\includegraphics[width=0.45\linewidth]{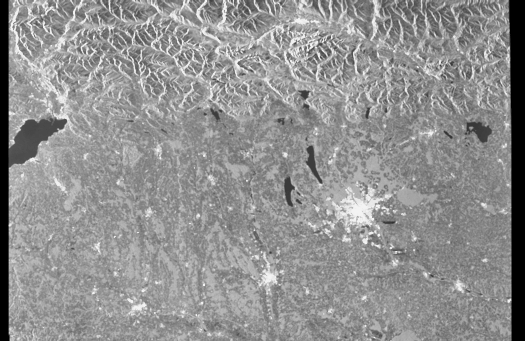}}
	\caption{Georeferenced Sentinel-1 images, quick looks for four quarters: a) Feb 09 2022, b) Apr 22 2022, c) Jul 27 2022, d) Oct 10 2022 (also 2023 is covered in TUM2TWIN). Munich represented by extended composition of salient scatterers on the center right. }
	\label{fig:Sentinel_1}
\end{figure}

\noindent\textbf{Sentinel-2}
The Sentinel-2 mission \cite{Spoto2010} \cite{Bouzinac2018} comprises three satellites with multi-spectral sensors, aimed at providing continuous imaging with medium spatial resolution on a global coverage. Launched in 2015, 2017, and 2024, the Sentinel-2 mission provides significant advantages over its multispectral predecessors. These include a revisit time of 5 days and a wider swath of 290 kilometers while maintaining remarkable spatial resolution. The multispectral instrument utilizes a push broom sensor, detecting reflected sun light with 13 spectral bands spanning from the visible spectrum to the short-wave infrared (SWIR). The spectral bands offer different spatial resolutions connected to their specific purposes and technical limits, ranging from 10 to 60 meters.

The TUM2TWIN collection contains quarterly Sentinel-2 data sets over the years 2022 and 2023, preselected for a low amount of cloud cover. The data is atmospherically corrected (bottom of atmosphere level 2A product) with the multitemporal processor MAJA \cite{hagolle2017}, developed in a cooperation between CNES, France, and DLR, Germany. The basic strategy for cloud detection, estimation of aerosol parameters, and correction of reflectance follows two main assumptions. Cloud cover is expected to change between subsequent image aquisitions, whereas ground cover is expected to remain stable. The calculation of bottom of atmosphere pixel amplitudes is based on radiative transfer models. 
The typical size of a resulting level 2A Sentinel-2 data set is about 2.4 GB. 

\noindent\textbf{Simulated CuBy}
The CuBy Satellite Network Bavaria (\cite{greza2023satellite}) consists of five 8U-CubeSats with 8-channel multispectral pushbroom cameras in its first mission stage that are due to launch in 2026. The satellites capture Bavaria on a three day repeat orbit with a swath width of 18 km and a ground sampling distance (GSD) of 4 m. Within the framework of CuBy, a satellite image simulator (\cite{lenz2023simulation}) is used to ease the development of the mission data processing chain before launch.

The camera simulator can generate artificial multispectral linescanner image strips from a high resolution basemap that is captured by a virtual sensor following a given orbit. The virtual sensor mimics a camera, with tunable specifications and camera parameters. Optical effects are applied through a modulation transfer function and filtering. 

Unlike the real data, the simulator encompass the ability to generate clouds, lens degradation and to simulate faulty pixels due to atomic oxygen and radiation.
It also allows for testing various other scenarios under different simulated parameters. 

\subsection{\textbf{Networks}}
\label{sec:networks}
%
In this context, \textit{networks} refer to the structured representation of the entire road system in a digital vectorized map.
A network is defined as a tuple:
\[
\mathcal{N} = \left( \mathcal{R}, \mathcal{J} \right)
\]
where:
\begin{itemize}
    \item \( \mathcal{R} \) is a set of roads. Each road \( r \in \mathcal{R} \) is represented by
    \[
    r = \{ \text{ID}_r, \; s_r^{\text{start}}, \; s_r^{\text{end}}, \; \gamma_r(s), \; L_r \},
    \]
    where:
    \begin{itemize}
        \item \( \text{ID}_r \) is a unique identifier,
        \item \( s_r^{\text{start}} \) and \( s_r^{\text{end}} \) denote the start and end of the road parametric domain,
        \item \( \gamma_r: [s_r^{\text{start}}, s_r^{\text{end}}] \to \mathbb{R}^2 \) is the road's reference road (centerline),
        \item \( L_r \) is the set of lane sections along the road.
    \end{itemize}
    
    \item \( \mathcal{J} \) is a set of junctions, which define the connectivity between roads. Junctions can be modeled as graphs where the nodes represent connection points and edges represent incoming or outgoing road segments.
\end{itemize}

Within each road \( r \), lane sections are defined over sub-intervals of the parameter \( s \). For a lane in a given section, its boundary can be described as an offset from the reference roads:
\[
l(s) = \gamma_r(s) + d(s) \, \mathbf{n}(s),
\]
where:
\begin{itemize}
    \item \( d(s) \) is the lateral offset function,
    \item \( \mathbf{n}(s) \) is the unit normal vector to the reference road \( \gamma_r(s) \).
\end{itemize}

\subsubsection{\textbf{HD Map}}
\label{sec:roadNetwork}
\noindent\textbf{Road Network}
%
%
To support the VR/AR simulator studies described in Section~\ref{sec:bike-simulator}, we develop a~\gls{HD} map using MathWorks RoadRunner in conjunction with orthophotos. 
This HD Map precisely represents the road network, capturing critical infrastructure details such as bike lanes, lane markings, standard road surfaces, curbs, sidewalks, and traffic lights. 
The workflow begins with integrating high-resolution orthophotos, which serve as a geospatial reference for accurately drawing road geometries. Using RoadRunner’s advanced editing tools, we model the road network, emphasizing the fidelity of bike lanes and lane markings to reflect real-world traffic conditions. 
Additionally, curbs and sidewalks are incorporated to enhance environmental realism, while strategically placed traffic lights replicate real intersections to support dynamic traffic simulations. After creating the detailed 2D layout, the~\gls{HD} map is merged with~\gls{LoD} 3 models, enriching the spatial environment with volumetric building structures. This fusion of detailed road elements, traffic control devices, and 3D models provides a comprehensive simulation environment that supports robust behavioral studies in VR/AR contexts. In addition, we can derive OpenDRIVE maps and FBX object files for seamless integration with simulation environments.
\begin{figure}
    \centering
    \includegraphics[width=0.9\linewidth]{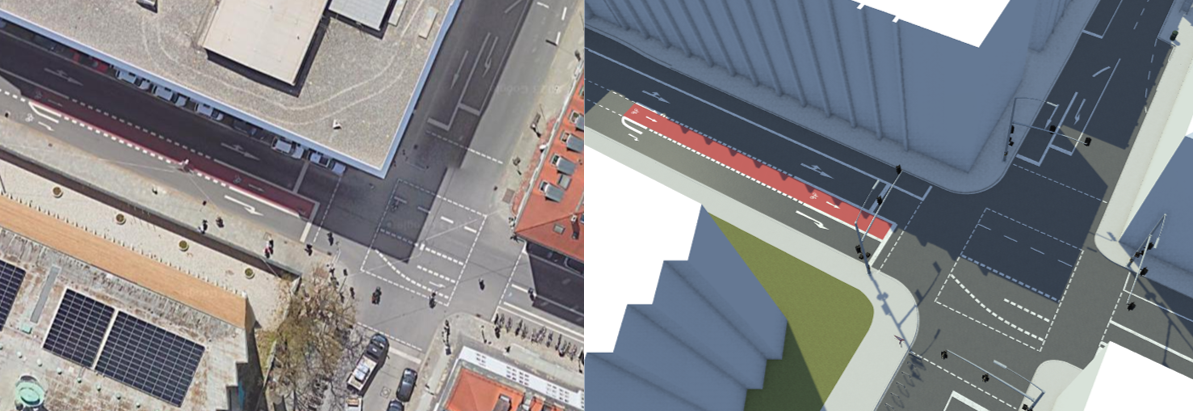}
    \caption{Side-by-Side illustration of an aerial view image (left) and the digital twin developed for VR/AR studies (right).}
    \label{fig:overview}
\end{figure}

\subsection{\textbf{3D Models}}
\label{sec:3Dmodels}
A 3D model is a digital representation of an object or scene in three dimensions. 
It is typically constructed using geometric primitives, such as vertices (points), edges (lines), and faces (surfaces), that define the shape and structure of the object. 

In our case, the triangle mesh stands as the base for modeling, whereby 3D point (vertex $v_i$) is the minimum discrete unit. 
The embedding of a triangle mesh into \(\mathbb{R}^3\) is described as a relation of 3D position \(\mathbf{p}_i\) to each vertex \(v_i \in V\):

\[
\mathbf{P} = \{\mathbf{p}_1, \dots, \mathbf{p}_V\}, \quad \mathbf{p}_i := \mathbf{p}(v_i) =
\begin{pmatrix}
x(v_i) \\
y(v_i) \\
z(v_i)
\end{pmatrix}
\in \mathbb{R}^3,
\]
Note, that each point \(\mathbf{p}\) in the interior of a triangle \([\mathbf{a}, \mathbf{b}, \mathbf{c}]\) can be described by barycentric coordinates:
\[
\mathbf{p} = \alpha \mathbf{a} + \beta \mathbf{b} + \gamma \mathbf{c}, 
\]
where
\[
\alpha + \beta + \gamma = 1, \quad \alpha, \beta, \gamma \geq 0.
\]

A triangle mesh \(\mathcal{M}\) consists of a geometric and a topological component.
The topology can be represented by a graph structure with a set of vertices
\[
V = \{v_1, \dots, v_V\}
\]
and a set of triangular faces connecting them
\[
F = \{f_1, \dots, f_F\}, \quad f_i \in V \times V \times V.
\]
Eventually, each face \(f \in F\) corresponds to a triangle in 3-space specified by its three vertex positions.

We distinguish three geometric model representations in the following subsections \cite{Kolbe2021}.
i) Semantic 3D models extend the definition by semantics attached to each geometric object at different hierarchical abstraction levels (e.g., city, building, facade), which are represented by B-Rep and volume obtained by accumulation of bounding faces.
The faces can be extended by additional texture information.
ii) Computer-aided design (CAD) models, where we opt for the B-Rep representation as well with shallow semantics only concerning leaf objects.
The faces can be extended by additional texture information.
iii) Mesh models represent data model where no semantics is included and no low-poly representation is provided, in contrast to i) and ii).
The faces can be extended by additional texture information.
%


\subsubsection{\textbf{Semantic Building Models}}
\label{sec:semanticBldMod}

\begin{figure}
    \centering
    \includegraphics[width=0.7\linewidth]{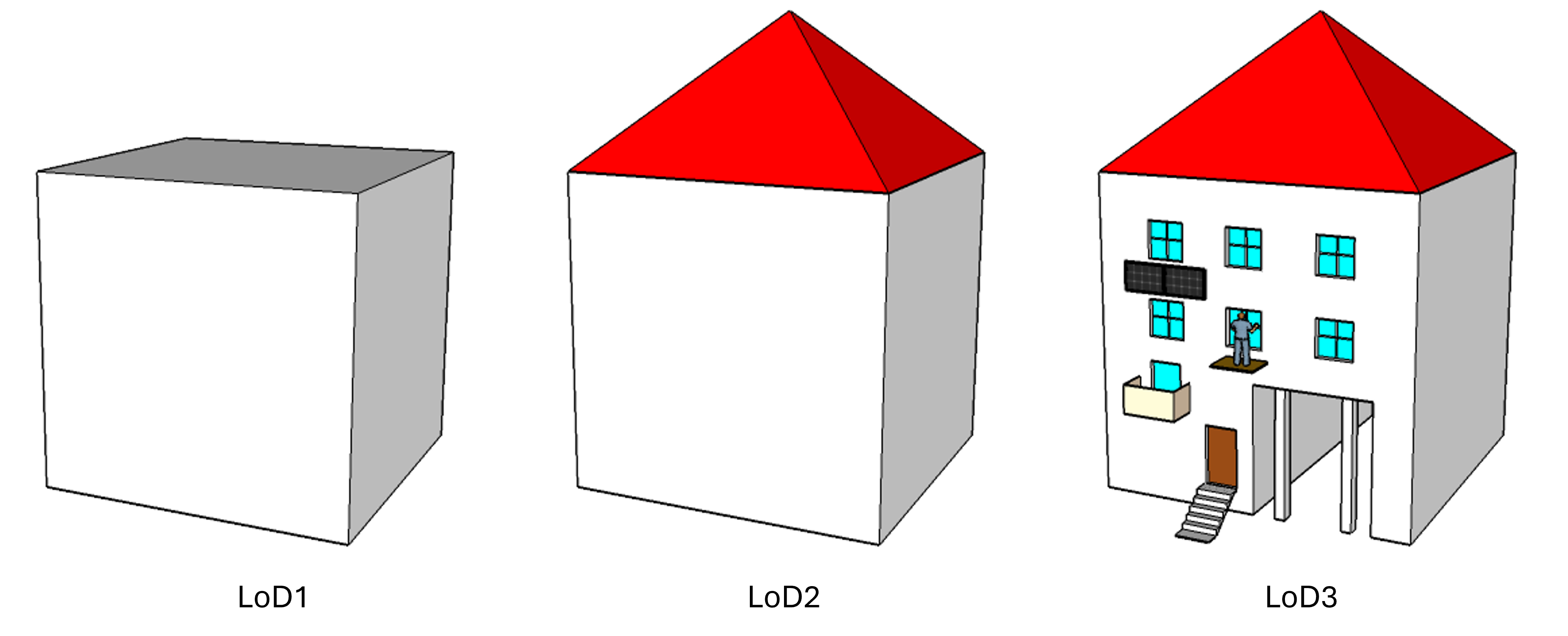}
    \caption{The primary difference between different~\gls{LoD}s: While~\gls{LoD}1 is an approximated polyhedral shape of a building (left),~\gls{LoD}2 displays additional roof types (center), and~\gls{LoD}3 complements it with facade details (right). }
    \label{fig:lods}
\end{figure}

To exchange semantic, geometric, topological and appearance information of cities and landscapes interoperably, the CityGML standard has become internationally established \cite{wysocki2024reviewing}.
It is issued by the Open Geospatial Consortium (OGC) and defines a conceptual data model as well as an encoding for the Geography Markup Language (GML) \cite{kolbeOGCCityGeography2021,kutznerOGCCityGeography2023}.
CityGML is based on the geographic information system standards from the ISO 191XX series, which include a comprehensive geometry model.
The entities of a city and landscape are decomposed into a hierarchical data structure that captures inter-object relationships \cite{Kolbe2021}.
As shown in Figure~\ref{fig:lods}, one object can be represented by multiple~\glspl{LoD}.
In order to ensure the direct application of semantic models across use cases and existing software solutions, such as the 3D City Database, we provide them in accordance with the standard \cite{biljeckiApplications3DCity2015,yao3DCityDB3DGeodatabase2018}.

\noindent \textbf{LoD1} 
The key elements in the established reconstruction approach for~\gls{LoD}1 building models are footprints and height.
To maintain the cross-consistency of each~\gls{LoD} buildings, we extracted footprints from~\gls{LoD}2 ground surface (\textit{gml:GroundSurface}), while the height information stemmed from the Real ALS (Sec.~\ref{sec:als}) data. 
We deployed the well-established \textit{3dfier} \cite{3dfier} software for the reconstruction \cite{chenhao_msc_2023}.

\noindent \textbf{LoD2}
A large number of public authorities worldwide provide and maintain~\gls{LoD}2 building models as open data, which includes the complete building stock of Germany, Switzerland, Poland, and large parts of Japan \cite{wysocki2024reviewing}.
The building models of the Bavarian State Mapping Agency cover the TUM main campus and are based on the official cadastre guaranteeing georeferencing accuracy in the centimeter range \cite{RoschlaubBatscheider}.
The stable management of object identifiers by public authorities forms the basis for integrating other representations.

\noindent \textbf{LoD2 - Textured}
Here, the textures serve as an extension of the base~\gls{LoD}2 with the geometry remaining intact.
The addition to the above-mentioned ~\gls{LoD}2 buildings, are optical textures projected onto the facades of buildings. 
The data stems from a manual terrestrial acquisition of images described in Section~\ref{sec:streetImages}, which are also co-aligned using~\gls{LoD}3 facade elements. 
Yet, still minor projection distortion exist owing to the deployed perspective camera;
The primary application of the data is for testing automatic texturing approaches \cite{tang2025texture2lod3}.
For further modeling details see our published manual~\footnote{\url{https://tum2t.win/tutorials/Facade-texturing-using-SketchUp}}.

\noindent \textbf{LoD3}
Analogously to~\gls{LoD}1 models, we used footprints and geometry of~\gls{LoD}2 as a base for~\gls{LoD}3 modeling.
The main difference pertains to correcting the geometry in presence of overhangs (roof geometry) and enriching facades in openings (i.e., windows, doors) and building installation (e.g., stairs) if exceeding the threshold of intrusion or extrusion by 10 cm.
The 3D measurements of combined proprietary point clouds \cite{mofa} and TUM-MLS-16 (Sec.~\ref{sec:mls}) are used for modeling. Additional 3D library of~\gls{CAD} models of facade elements is created to allow 3D elements modeling.
For further modeling details see our published manual~\footnote{\url{https://creating-citygml-datasets.readthedocs.io/en/latest/creation-guidelines/lod3-models-based-on-point-clouds.html}}.

\subsubsection{\textbf{Semantic Streetspace Models}}
\begin{figure}
    \centering
    \includegraphics[width=0.7\linewidth]{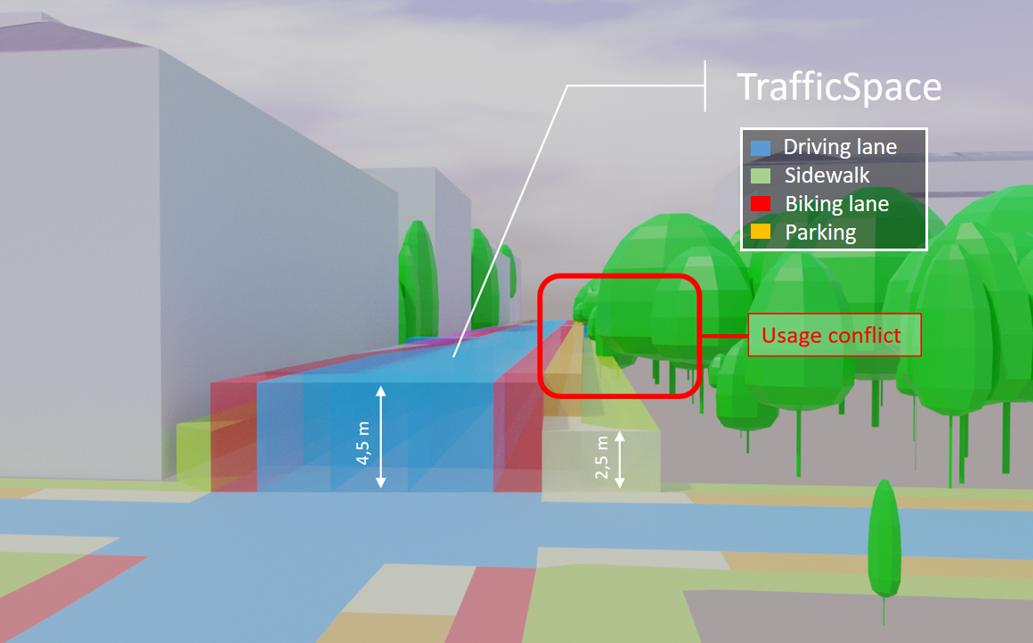}
    \caption{Volumetric geometry representations of traffic spaces with different heights according to respective functions \cite{beil2024}.}
    \label{fig:citygmlTrafficSpace}
\end{figure}

\noindent \textbf{Semantic Streetspace Model}
The Transportation Module has been significantly revised in CityGML version 3.0 enabling the representation of the street space in three granularity levels: area, way, and lane.
At granularity level~\emph{lane} each individual lane is modeled separately with predecessor and successor relations enabling routing applications.
As shown in Figure~\ref{fig:citygmlTrafficSpace}, traffic spaces describe where the actual traffic takes place and can be represented by volumetric geometries, whereas traffic areas refer to the ground surfaces of these traffic spaces.

All objects in the HD map, discussed in Section~\ref{sec:networks}, are defined with parametric geometries relative to the road reference line.
To supplement this with a representation utilizing explicit absolute coordinates, we converted the OpenDRIVE dataset to CityGML 3.0 using the tool \textit{r:trån}
\cite{schwab2020}.
\subsubsection{\textbf{Semantic Vegetation Models}}

\noindent \textbf{Vegetation Models}
The trees for the entire city of Munich, inlcuding the TUM campus, have been reconstructed
using~\gls{ALS} point clouds, multispectral imagery, and~\gls{LoD}2 building models \cite{muenzinger2022}.
The tree crown is reconstructed using geometric primitives based on a set of derived parameters, including an ellipse fitted to the convex hull and the highest laser scanning return within the crown.
The dataset encompasses 2.7 million tree objects with tree species attributes.

\noindent \textbf{Tree Models}
Trees in the vicinity of the TUM main campus are in-situ surveyed for diameter at breast height \cite{zagst_msc_2023}.
The 3D tree shape is derived from the TUM-MLS-18 scans.
The dataset comprises around 50 tree objects with height, trunk and crown diameter.

\subsubsection{ \textbf{CAD Models}}
Geometrically,~\gls{CAD} models are typically described using parametric and explicit representations of i) Boundary Representation (B-Rep): Defines objects through their surfaces composed of edges, vertices, and faces, which form a closed (watertight) shell;
ii) Constructive Solid Geometry (CSG): Builds complex shapes by combining primitive solids (e.g., cubes, cylinders) using Boolean operations (union, difference, intersection);
or iii) by Parametric Models: Allow geometry to be defined by parameters and constraints, enabling easy modification and design automation.
In our case, we focus solely on the B-Rep representation, remaining consistent to our other models.
The data is available in the .SKP-format.

\noindent \textbf{CAD Building Models}
The created CAD models are derivatives of the semantic building models at~\gls{LoD}3 discussed in Section~\ref{sec:semanticBldMod}.
The primary difference pertains to the flatten hierarchy of the data representation, while geometry remains intact.
We acknowledge the large CAD-oriented community leveraging CAD modalities, and foresee applications in outdoor object reconstruction.

\noindent \textbf{CAD Building Models - Textured}
Analogously, to the above-mentioned \textit{CAD Building Models} the created CAD models are derivatives of the textured semantic building models at~\gls{LoD}2 discussed in Section~\ref{sec:semanticBldMod}.
The primary difference pertains to the flatten hierarchy of the data representation, while geometry and texture remains intact.

\subsubsection{ \textbf{Mesh Models}} \label{sec:uas_3dmesh}

\noindent \textbf{UAS 3D Mesh}
A 3D mesh model of the TUM main campus (Fig.~\ref{fig:uas}) is reconstructed based on UAS images captured from a DJI Matrice 350~\gls{RTK} equipped with the RGB camera integrated into a Zenmuse L2 laser scanner (Section~\ref{sec:uas_photo}). 
3D mesh models are reconstructed using an automatic pipeline in Pix4Dmatic (v1.68) representing the standard mesh reconstruction approach \cite{pix4dmatic}. 
The whole model is available in OBJ format, including more than 1 million triangles and about 494k vertices. The coordinate reference system (CRS) is WGS 84 / UTM Zone 32N.


\section{TUM2TWIN Current Downstream Tasks}
\label{useCase}
\begin{figure*}
    \centering
    \includegraphics[width=0.9\linewidth]{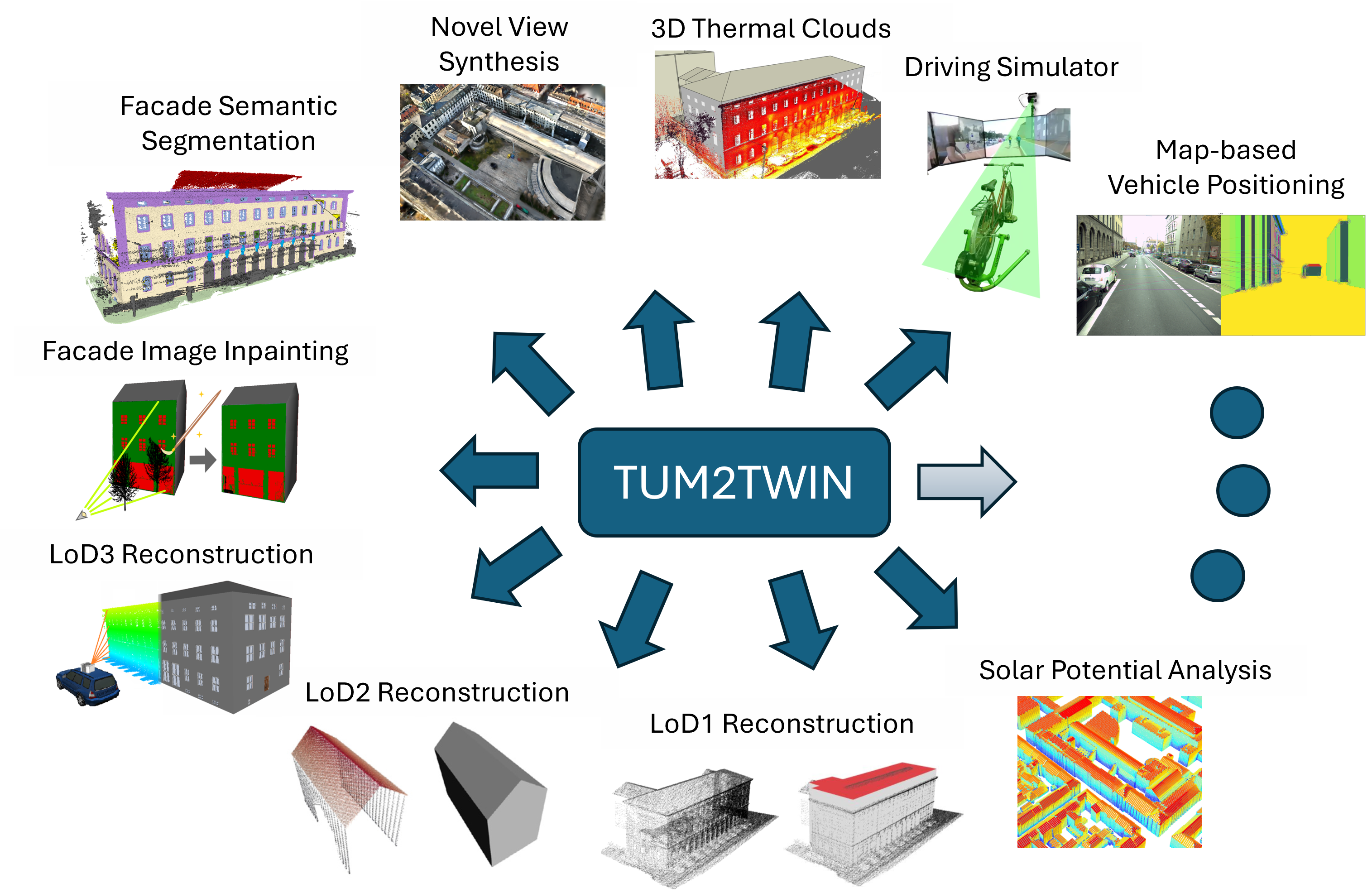}
    \caption{The TUM2TWIN-unlocked downstream tasks: Multiple research areas already unveiled with a potential to cater for many more. The research encompasses map-based vehicle positioning (Sec.~\ref{sec:map-basedVehicle}), LoD1 building reconstruction (Sec.~\ref{sec:lod1rec}), LoD2 building reconstruction (Sec.~\ref{sec:lod2rec}), LoD3 reconstruction (Sec.~\ref{sec:lod3rec}),  facade semantic segmentation (Sec.~\ref{sec:facadeSegmentation}), facade image inpainting (Sec.~\ref{sec:facadeInp}), 3D thermal point cloud projection ( Sec.~\ref{sec:thermalCloud}), novel view synthesis (Sec.~\ref{sec:nvs}), driving simulators development (Sec.~\ref{sec:bike-simulator}), and solar potential analysis (Sec.~\ref{sec:solarAnalysis}). Additionally, we elaborate on potential downstream tasks in Section~\ref{sec:PotentialUseCase}. } 
    \label{fig:usecasesFigure}
\end{figure*}
To showcase usability of the presented TUM2TWIN dataset, we present selected research already leveraging the ubiquity and comprehensiveness of the benchmark dataset for the specific use cases; exemplary shown in Figure~\ref{fig:usecasesFigure}.
%
%
\subsection{ \textbf{Map-based Vehicle Positioning}}
\label{sec:map-basedVehicle}
Positioning in dense urban environments is challenging for GNSS-based systems due to signal obstructions in urban canyons. To address this, vehicles often rely on alternative cues such as visual data to enhance localization accuracy.

Cameras are particularly useful due to their prevalence in vehicles and ability to capture environmental features. Traditional and novel feature matching techniques are commonly used, but require georeferenced 3D maps for global positioning.
A method is introduced that utilizes high-fidelity LoD3 models and imagery for vehicle localization, where images are fused with semantic 3D geometry, and evaluation of LoD3’s advantages over LoD2 in camera-based localization is presented \cite{bieringer2024analyzing}.

\noindent \textbf{TUM2TWIN provides:}
LoD3 and LoD2 ground truth models (Sec.~\ref{sec:semanticBldMod}), street-level images with vehicle trajectory (Sec.~\ref{sec:streetImages}).

\subsection{\textbf{LoD1 Building Reconstruction}}
\label{sec:lod1rec}
Within the broader field of city model reconstruction, semantic 3D building reconstruction has become a central focus owing to the significance of buildings in urban environments and their diverse range of applications.
For~\gls{LoD}1, while many approaches rely on detailed 3D observations, an alternative observation-free strategy has also gained attention, since they do not require detailed roof reconstruction \cite{wysocki2024reviewing}.

To validate the performance of various methods one needs to evaluate reconstruction results compared to ground truth. 
Usually, this are i) laser scanning observations, ii) or manually modeled~\gls{LoD}1 building models.
A comprehensive analysis of various modalities limitations (\gls{ALS} vs~\gls{MLS}) is presented, where currently available open source methods for~\gls{LoD}1 reconstruction and various building shapes are compared \cite{chenhao_msc_2023}.

\noindent \textbf{TUM2TWIN provides:}~\gls{ALS} point clouds (Sec.~\ref{sec:als}),~\gls{MLS} point clouds (Sec.~\ref{sec:mls}), footprints and baseline LoD1, 2, and 3 building models (Sec.~\ref{sec:semanticBldMod}).

\subsection{ \textbf{LoD2 Building Reconstruction}}
\label{sec:lod2rec}
Unlike~\gls{LoD}1 building models,~\gls{LoD}2 building models require complex roof type reconstruction, which is both challenging and opening new applications, such as solar potential analysis for photovoltaic panels placement \cite{biljeckiApplications3DCity2015}. 

To bridge the abstraction gap between existing city-building models and their underlying instances, PolyGNN~\cite{PolyGNNZhaiyu} has been developed using a large-scale synthetic~\gls{ALS} dataset with well-defined polyhedral ground truths. 
Furthermore, a transferability analysis has been conducted, demonstrating that PolyGNN, although trained solely on simulated ALS data for LoD2 building reconstruction, generalizes effectively to real-world ALS data. 
PolyGNN learns a piecewise planar occupancy function, underpinned by polyhedral decomposition, enabling efficient and scalable 3D building reconstruction. 

\noindent \textbf{TUM2TWIN provides:} Real~\gls{ALS} and synthetic~\gls{ALS} (Sec.~\ref{sec:als}),~\gls{LoD}1, 2, and 3 building models (Sec.~\ref{sec:semanticBldMod}).

\subsection{ \textbf{LoD3 Building Reconstruction}}
\label{sec:lod3rec}
Yet additional challenge imposes~\gls{LoD}3 reconstruction which requires geometric representation of the facade elements. 
There are multiple applications given this modality is available, such as autonomous driving car testing to flood risk simulations \cite{wysockiMLS2LoD3}.

The methods developed based on the TUM2TWIN include Scan2LoD3 \cite{wysocki2023scan2lod3} or Texture2LoD3 \cite{tang2025texture2lod3}, and are not limited to them \cite{wang2024framework}.
For example, Scan2LoD3 introduces a method where uncertainty-aware analysis of~\gls{MLS} laser rays with 3D models yields conflict maps that delineate openings. Then, it is deployed as evidence for late-fusion of segmented point clouds and  segmented street-level images. 

\noindent \textbf{TUM2TWIN provides:}~\gls{LoD}3 building models and ground truth textures (Sec.~\ref{sec:semanticBldMod}),~\gls{MLS} facade-annotated point clouds (Sec.~\ref{sec:mls}), and street-level images (Sec.~\ref{sec:streetImages}). 

\subsection{ \textbf{3D Facade Semantic Segmentation} }
\label{sec:facadeSegmentation}

Facade semantic segmentation remains a persistent challenge in photogrammetry and computer vision. 
Despite decades of research introducing various segmentation methods, there is still a lack of comprehensive facade classes and datasets that capture architectural diversity: 
Robust and standardized 3D facade semantic segmentation may unlock more comprehensive urban space interpretation and robust facade-level semantic surface reconstruction.

A method for 3D semantic facade segmentation is introduced that leverages geometric features \cite{jutziFeatures} as an early injection vector into the standard deep learning networks \cite{yuetanDeepLearning}. 
Other publications further explore this direction using the TUM2TWIN data \cite{wysockiVisibility}, or create evaluation baselines and evaluation benchmark datasets, i.e., the largest facade segmentation dataset: ZAHA \cite{wysocki2024zahaintroducinglevelfacade}.

\noindent \textbf{TUM2TWIN provides:}~\gls{MLS} facade-annotated point clouds (Sec.~\ref{sec:mls}).

\subsection{ \textbf{Facade Image Inpainting}}
\label{sec:facadeInp}

\begin{figure}
    \centering
    \includegraphics[width=0.8\linewidth]{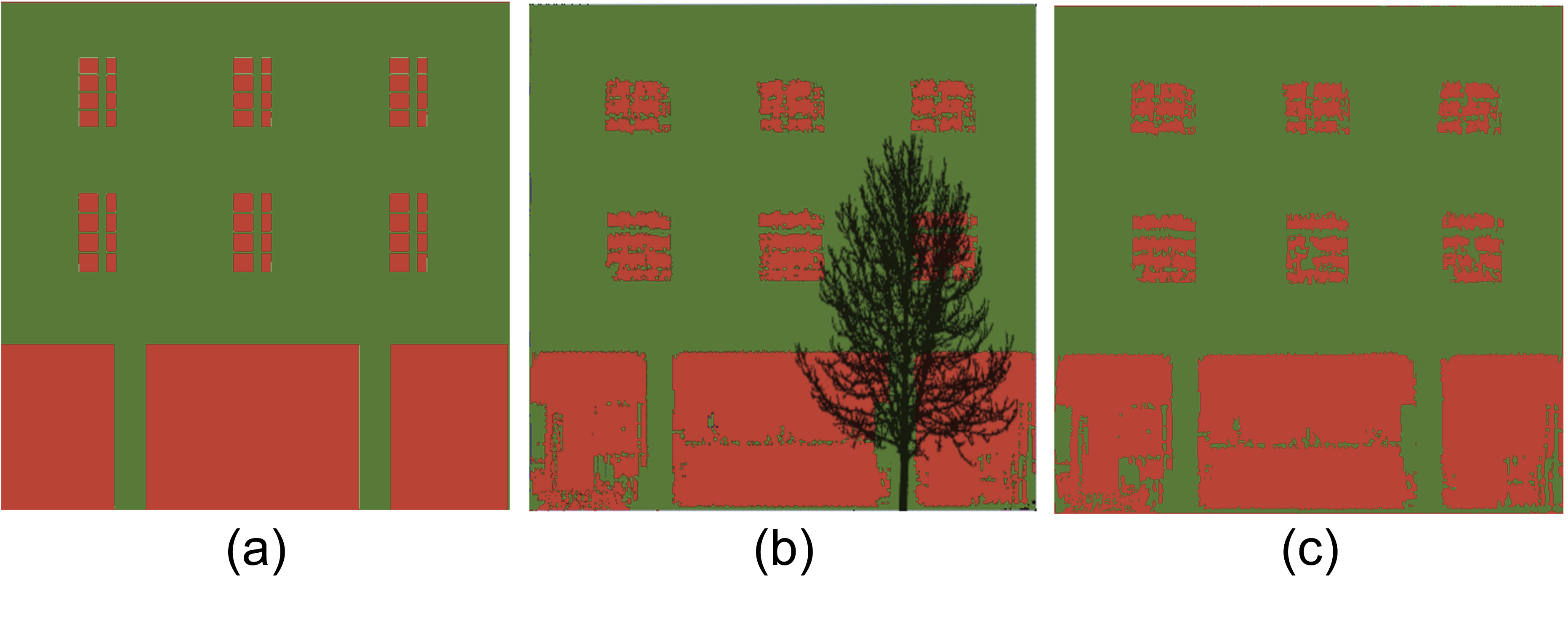}
    \caption{Exemplary facade image inpainting: a) Ground-Truth CM derived from semantic LoD3 building model b) Masked CM derived from semantic LoD2 building model and corresponding MLS point cloud c) Inpainting result using a personalized~\gls{DM}. Figure adapted from \cite{facadiffyFroech}.}
    \label{fig:facadeInpaintingExample}
\end{figure}

\gls{CM} indicating potential conflicts between existing semantic LoD2 building models and corresponding point clouds have diverse applications for example in LoD3 building reconstruction (\cite{wysocki2023scan2lod3}) or change detection (\cite{tuttas2015validation}). 
As Figure~\ref{fig:facadeInpaintingExample} b) illustrates, laser rays are frequently obstructed by objects such as vegetation in~\gls{MLS} measurements, thus yielding incomplete~\gls{CM}s. 

In their method on leveraging~\gls{DM} for completing~\gls{CM}s, \cite{facadiffyFroech} utilize the semantic LoD3 building models (Section 3.6.1), to obtain ground truth information for validating their approach.
Figure~\ref{fig:facadeInpaintingExample} provides an example, illustrating  a) a~\gls{CM} derived from a semantic LoD3 building model, b) the corresponding, partially incomplete~\gls{CM}, derived from~\gls{MLS} point clouds and the LoD2 building model, and c) the~\gls{CM} that has been completed with a~\gls{DM}.

The rich semantic information provided within the LoD3 models makes it possible to include prior knowledge about the behavior of facade components into the process of deriving ground-truth~\gls{CM}s. 
For instance, windows and doors are considered to be causing conflicts due to the voyeur effect (\cite{tuttas_reconstruction_2013}), while the largest identified wall surface is considered to be confirming.

\noindent \textbf{TUM2TWIN provides:}~\gls{MLS} point clouds (Sec.~\ref{sec:mls}),~\gls{TLS} point clouds (Sec.~\ref{sec:tls}),~\gls{LoD}2 and 3 building models (Sec.~\ref{sec:semanticBldMod}).

\subsection{ \textbf{3D Thermal Point Clouds Projection}}
\label{sec:thermalCloud}

The digital representation of energy distribution in buildings can serve a multitude of purposes in today’s world - such as energy consumption inspection, structural health monitoring, etc. In this study, the aim is to develop a workflow for thermal mapping to generate the digital thermal representations.

\begin{figure*}
    \centering
    \includegraphics[width=0.7\linewidth]{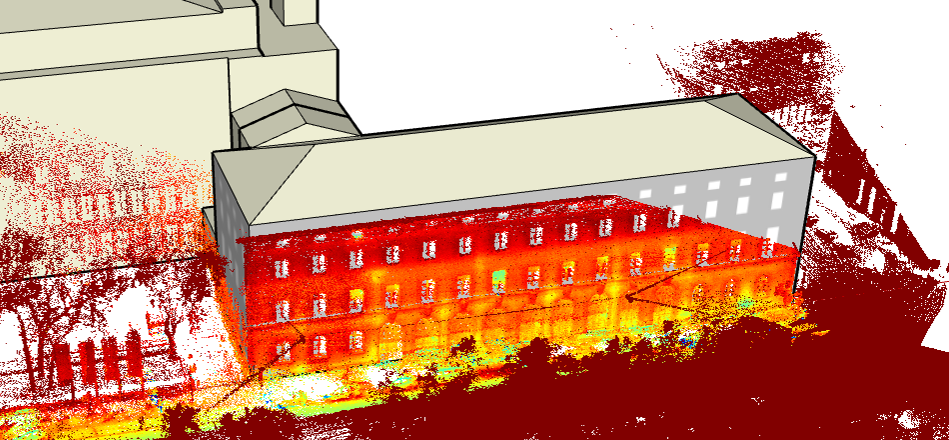}
    \caption{Building model overlayed with 3D thermal point clouds for projection.}
    \label{fig:thermal_projection}
\end{figure*}

For this purpose, the TUM2TWIN dataset serves as a benchmark dataset. The dataset provides laser scanner point clouds (TUM-MLS-16), Street-Level~\gls{TIR} image sequences,~\gls{LoD}2 and refined~\gls{LoD}3 CityGML building models. As a pre-processing step, the laser scanner point clouds are fused with the~\gls{TIR} images to generate 3D thermal point clouds \cite{zhu2021fusion}. Then the building models are overlayed with the thermal point clouds as shown in Figure~\ref{fig:thermal_projection}. A mapping algorithm projects these thermal point clouds to building facades to generate the thermal textures \cite{biswanath2023thermal}. The projection algorithm uses Nearest-Neighbor and Bilinear Interpolation method.
These generated thermal textures can be seen as an enrichment of digital twins of building, which is also part of a bigger multi-scale digital twin project – AI4TWINNING \cite{borrmann2023ai4twinning}


\noindent \textbf{TUM2TWIN provides:} Various building model types (Sec.~\ref{sec:3Dmodels}), street-level thermal infrared imagery (Sec.~\ref{sec:thermalInf}).

\subsection{ \textbf{Novel View Synthesis: NeRF and Gaussian Splatting} }
\label{sec:nvs}

\gls{NeRF} has been widely adopted for novel view synthesis by implicitly representing a scene as a continuous volumetric function trained with a set of posed images \cite{mildenhall2021nerf}. 
By learning a mapping from 3D spatial coordinates and viewing directions to color and density values,~\gls{NeRF} facilitates high-fidelity 3D scene reconstruction \cite{petrovska2024vision}.

The TUM2TWIN dataset provides a robust foundation for NeRF-based urban digital twin applications, as it offers georeferenced high-resolution multi-view images, ensuring accurate spatial consistency and high-quality reconstructions. 
In this study, the nerfacto is employed, an enhanced~\gls{NeRF} variant implemented in Nerfstudio \cite{tancik2023nerfstudio}. 
Nerfacto integrates components from mip-NeRF 360 \cite{barron2022mip} and instant-NGP \cite{muller2022instant} to achieve a balance between reconstruction quality and computational efficiency.

\begin{figure*}[h]
    \centering
    \begin{subfigure}{0.32\textwidth}
        \centering
        \includegraphics[width=\linewidth]{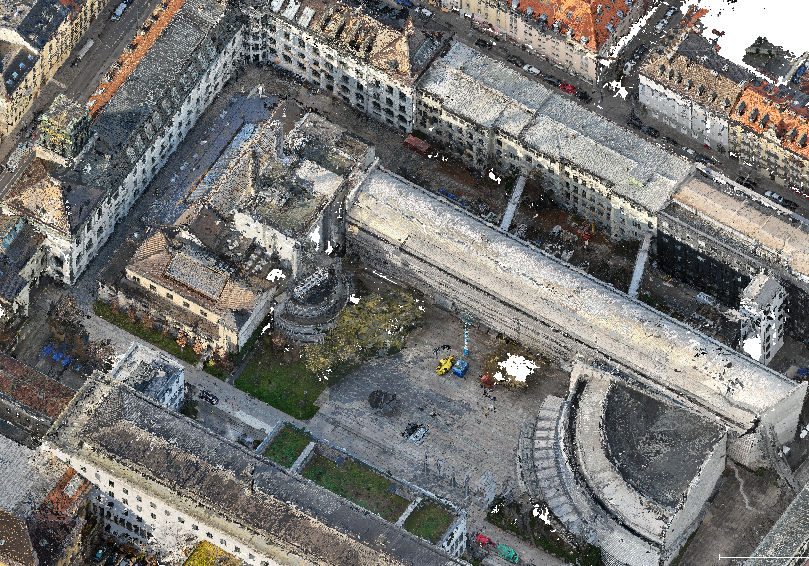}
        \caption{}
        \label{fig:gt}
    \end{subfigure}
    \hfill
    \begin{subfigure}{0.3\textwidth}
        \centering
        \includegraphics[width=\linewidth]{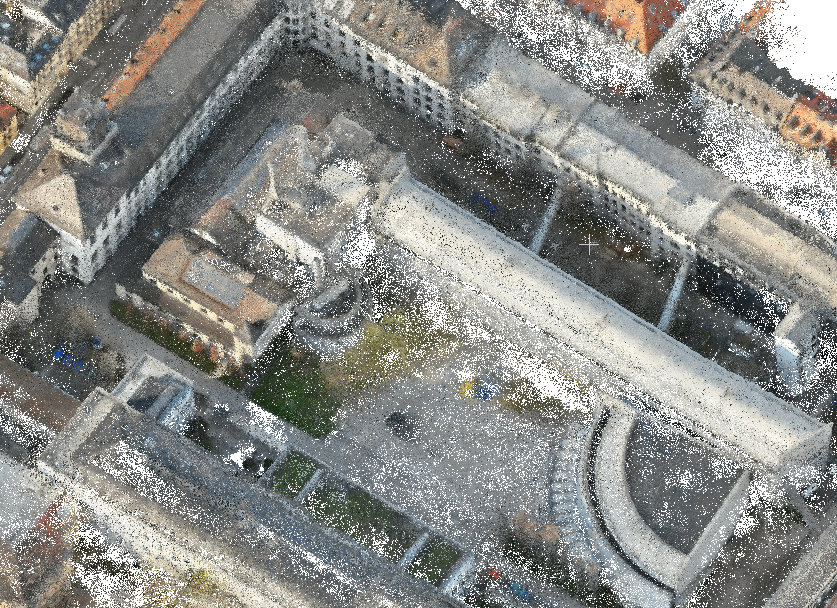}
        \caption{}
        \label{fig:nerf_reconstruction}
    \end{subfigure}
    \hfill
    \begin{subfigure}{0.32\textwidth}
        \centering
        \includegraphics[width=\linewidth]{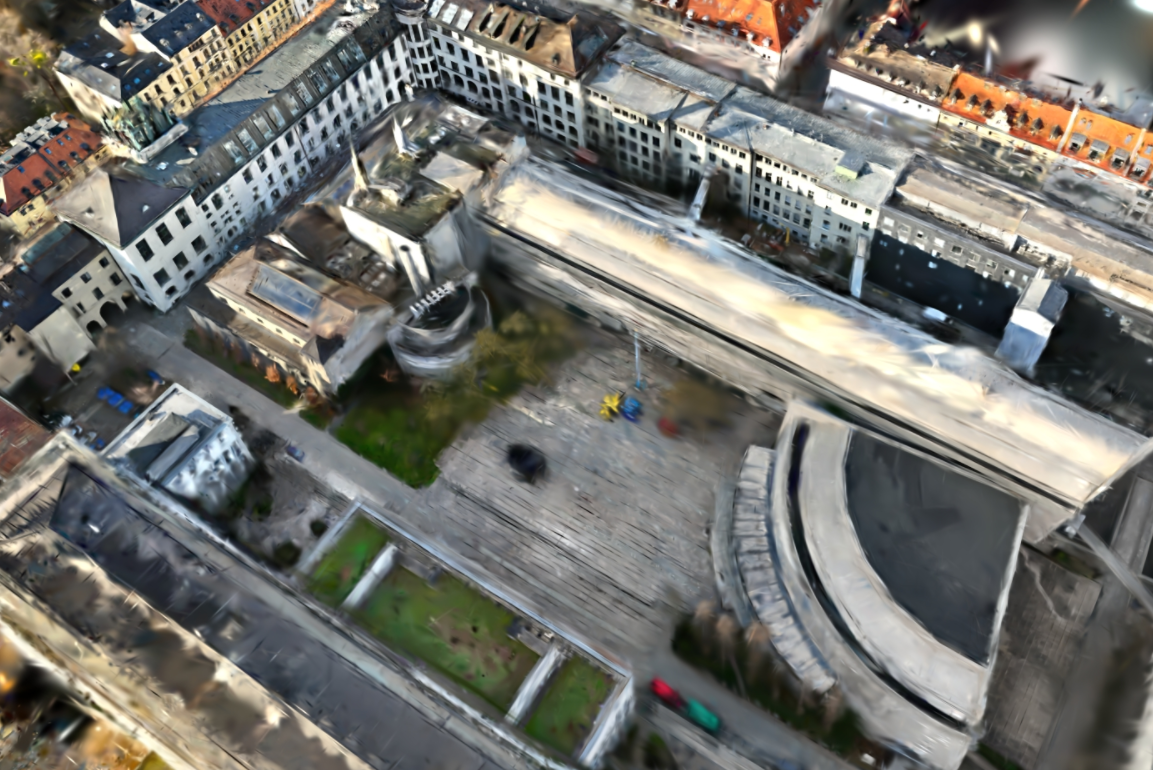}
        \caption{}
        \label{fig:gs_reconstruction}
    \end{subfigure}
    \caption{Comparison of reconstruction results using the TUM2TWIN dataset. (a) Ground Truth (described in Section~\ref{sec:uas}), (b) NeRF-based reconstruction, and (c) Gaussian Splatting-based reconstruction.}
    \label{fig:reconstruction_comparison}
\end{figure*}

Figure~\ref{fig:nerf_reconstruction} illustrates the NeRF-based reconstruction results obtained using the TUM2TWIN dataset. 
The model successfully captures fine-grained urban details, including building facades, streets, and occlusions. 
%


\gls{3DGS} has emerged as an efficient representation for novel view synthesis and real-time 3D scene reconstruction \cite{kerbl20233d}. 
Unlike volumetric rendering methods such as~\gls{NeRF}, which rely on implicit neural fields, 3DGS represents a scene as a collection of anisotropic 3D Gaussians, enabling direct rasterization and significantly reducing rendering time \cite{zhang2024cdgs}. 
This approach offers compact scene representations while maintaining high reconstruction fidelity. 

Similar to NeRF-based methods,~\gls{3DGS} benefits from high-resolution multi-view images and precise spatial information, making the TUM2TWIN dataset well-suited for~\gls{3DGS} applications. 

Figure~\ref{fig:reconstruction_comparison} provides a comparative visualization of the reconstruction results of NeRF and Gaussian Splatting approaches. 
It illustrates the structural and visual variations between the NeRF-based and 3DGS-based reconstructions using the TUM2TWIN dataset, with the Ground Truth, obtained from Section~\ref{sec:uas}, included for reference.

\noindent \textbf{TUM2TWIN provides:}~\gls{UAS} images with~\gls{UAS} pose and trajectory (Sec. \ref{sec:uas_photo}), \gls{UAS} point clouds as ground truth (Sec.~\ref{sec:uas}), 3D mesh baseline (Sec.~\ref{sec:uas_3dmesh}).

\subsection{ \textbf{Driving and Biking Simulator} }
\label{sec:bike-simulator}

The bicycle and e-scooter simulator are designed to provide immersive experiences using either CAVE-VR or head-mounted VR technology. The simulator incorporates a 3-degree-of-freedom (3DOF) motion platform to enhance realism and user engagement, replicating dynamic movements corresponding to the virtual environment. This immersive setup is supported by the~\gls{HD} map created through the TUM2TWIN project, ensuring detailed and realistic infrastructure representation. FBX files derived from the~\gls{HD} map are directly imported into the Unity simulation environment, providing a consistent 3D world. Simultaneously, the OpenDRIVE files generated from the same HD map are integrated with~\gls{SUMO}, a microscopic traffic simulation software. This dual integration ensures both the VR environment and traffic simulation are synchronized within the same coordinate frame, enabling seamless interaction between virtual environments and traffic dynamics. The entire system is connected via Sumonity, facilitating real-time data exchange between the simulator and microscopic traffic simulation agents. This synchronization allows for the coordinated movement of study participants within the simulator and virtual traffic, enhancing the ecological validity of behavioral studies.

\noindent \textbf{TUM2TWIN provides:} Road network as HD Map (Sec.~\ref{sec:roadNetwork}), Semantic 3D building models (Sec.~\ref{sec:semanticBldMod}).

\subsection{ \textbf{Solar Potential Analysis} }
\label{sec:solarAnalysis}

Among renewable energy sources, solar power is often the preferred choice for buildings aiming to enhance energy sustainability and reduce reliance on fossil fuels. Integrating photovoltaic (PV) panels in urban planning can significantly contribute to reducing greenhouse gas emissions, mitigating the adverse effects of climate change, and promoting energy independence. The TUM2TWIN dataset provides a robust basis for assessing solar potential on both building facades and rooftops. The detailed LoD3 models, which include architectural elements such as windows, balconies, and facade textures, enable precise estimations of available surface areas for PV panel installations.

\begin{figure*}
    \centering
    \includegraphics[width=0.9\linewidth]{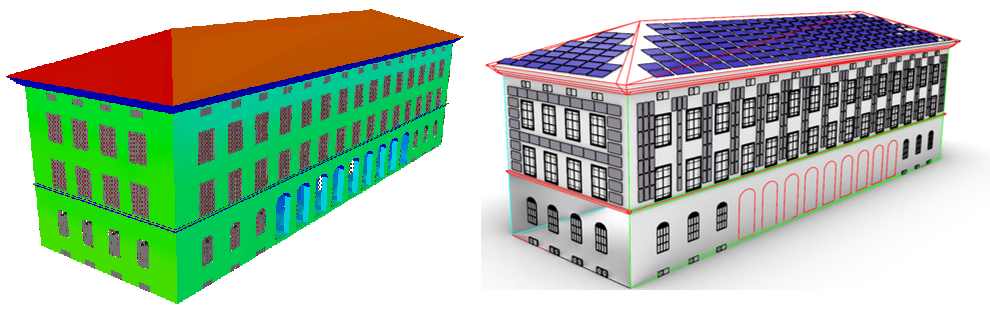}
    \caption{Solar potential analysis (left) and photovoltaic panel placement (right) on the TUM2TWIN.}
    \label{fig:solar_potential}
\end{figure*}

To enable solar potential analysis, the SunPot tool \cite{willenborg2018integration} is used on the TUM2TWIN dataset to calculate solar irradiation, taking into account the shadowing effects of surrounding buildings and vegetation. The resulting textured LoD3 model is then used to determine optimal PV placements and estimate the potential energy yield from facade-integrated solar panels, as can be seen in Figure~\ref{fig:solar_potential}.

This approach is particularly valuable for dense urban environments, where rooftop space is often limited, making facade-mounted PV systems a crucial strategy for maximizing renewable energy generation. By integrating LoD3-based solar potential analysis within Urban Digital Twins, city planners can support data-driven decision-making for sustainable energy planning and facilitate more accurate simulations of urban energy dynamics.

\noindent \textbf{TUM2TWIN provides:} Semantic 3D building models at~\gls{LoD}3 (Sec.~\ref{sec:semanticBldMod}).



\section{TUM2TWIN Potential Downstream Tasks}
\label{sec:PotentialUseCase}
In this Section, we analyze the potential of selected applications of our comprehensive TUM2TWIN dataset.
This research direction have not been yet published, yet the TUM2TWIN components are essential for such downstream tasks, further exemplifying the contribution of the TUM2TWIN dataset. 

\subsection{ \textbf{2D Facade Understanding} }
%
High-quality datasets are scarce in architectural research, posing challenges for facade understanding, especially in labor-intensive tasks like segmentation, where annotation is costly and time-consuming. As a way of approaching the data scarcity challenge, deep generative models can be used to automatically generate annotated images. This includes the generation of semantic segmentation masks, a usual requirement for any subsequent facade understanding efforts. As described in \cite{caadria2025}, adapting such pipelines to different domains requires only a minimal manual annotated dataset, containing typically five or fewer samples per desired class. These models not only generate annotated images but also provide valuable features containing useful semantics enabling various downstream tasks.

The multiple domains covered by the TUM2TWIN dataset provide a valuable benchmark for validating such efforts. Although semantic segmentation masks are not available in the image-based portion of the dataset, the annotated point clouds and~\gls{LoD}3 models offer a potential set of segmentation classes. By using these classes alongside the available street-view imagery inside the TUM2TWIN dataset, the adaptability of such methodologies can be evaluated. Additionally, TUM2TWIN provides a valuable benchmark for image-based efforts dedicated to the parsing of facades and the induction of architectural grammars, potentially contributing to  2D facade understanding.

\noindent \textbf{TUM2TWIN provides:} Semantic 3D building models at~\gls{LoD}3 (Sec.~\ref{sec:semanticBldMod}), Textured building models at~\gls{LoD}2 (Sec.~\ref{sec:semanticBldMod}), Binary openings' ground truth masks (Sec.~\ref{sec:semanticBldMod}).

\subsection{ \textbf{Georeferenced BIM Reconstruction} }

As part of the endeavor to digitalize the built environment, the TUM2TWIN dataset facilitates the creation of multi-scale, multi-level digital city models by integrating detailed Building Information Models (BIM) with broader city models.

In addition to the as-designed BIM models, which are often created in a local coordinate system, common Scan-to-BIM practices mostly face challenges in generating georeferenced digital building models and integrating them with city models. Indoor laser scanning primarily captures a building’s interior in a local coordinate system, often lacking information about facade wall thicknesses and key architectural elements such as openings and facade structures.

High-quality geometric-semantic information at various levels can support the development of automated methods for creating complete and accurate georeferenced BIM models within city-scale semantic models.
By leveraging different levels of semantic detail in city models and applying techniques such as point cloud semantic segmentation and instance segmentation, it is possible to identify common architectural elements -- such as doors and windows -- which can serve as key reference points for georeferencing BIM models.

\noindent \textbf{TUM2TWIN provides:} Georeferenced~\gls{TLS} point clouds (Sec.~\ref{sec:tls}), Georeferenced semantic 3D models at~\gls{LoD}1, 2, and 3 (Sec.~\ref{sec:semanticBldMod}).

\subsection{ \textbf{Consistency Measures Across Representations} }
The concept of the~\gls{UDT} is becoming increasingly important as greater amounts of data are collected on a variety of aspects of the real world.~\gls{UDT}s are not monolithic entities where a single model can satisfactorily describe and fulfill the requirements of all urban applications.~\gls{UDT}s can be better understood as a collection of data from different representations that describe the different facets of the same real-world object in different structures and levels of detail. For these diverse representations to be coherent, the models must be coherent with each other in all modelling aspects. In order to verify coherence, the correspondence between two concrete data sets (3D models) of different forms of representation must be determined on the basis of an objective measure. The models from the TUM2TWIN dataset have been collected as a first step to an interconnected digital twin. 
%

Further, RichVoxels (Semantically enriched voxels) are used as a common representation to match the various 3D model types amongst each other to objectively describe their spatio-semantic coherence \cite{heeramaglore2022} .

\noindent \textbf{TUM2TWIN provides:} Various coregistered and georeferenced representations of the same objects: Point clouds (Sec.~\ref{sec:pcl}), images  (Sec.~\ref{sec:img}), networks  (Sec.~\ref{sec:networks}), and 3D models (Sec.~\ref{sec:3Dmodels}).

\subsection{ \textbf{Multimodal Coregistration} }

\begin{figure*}
    \centering
    \includegraphics[width=0.9\linewidth]{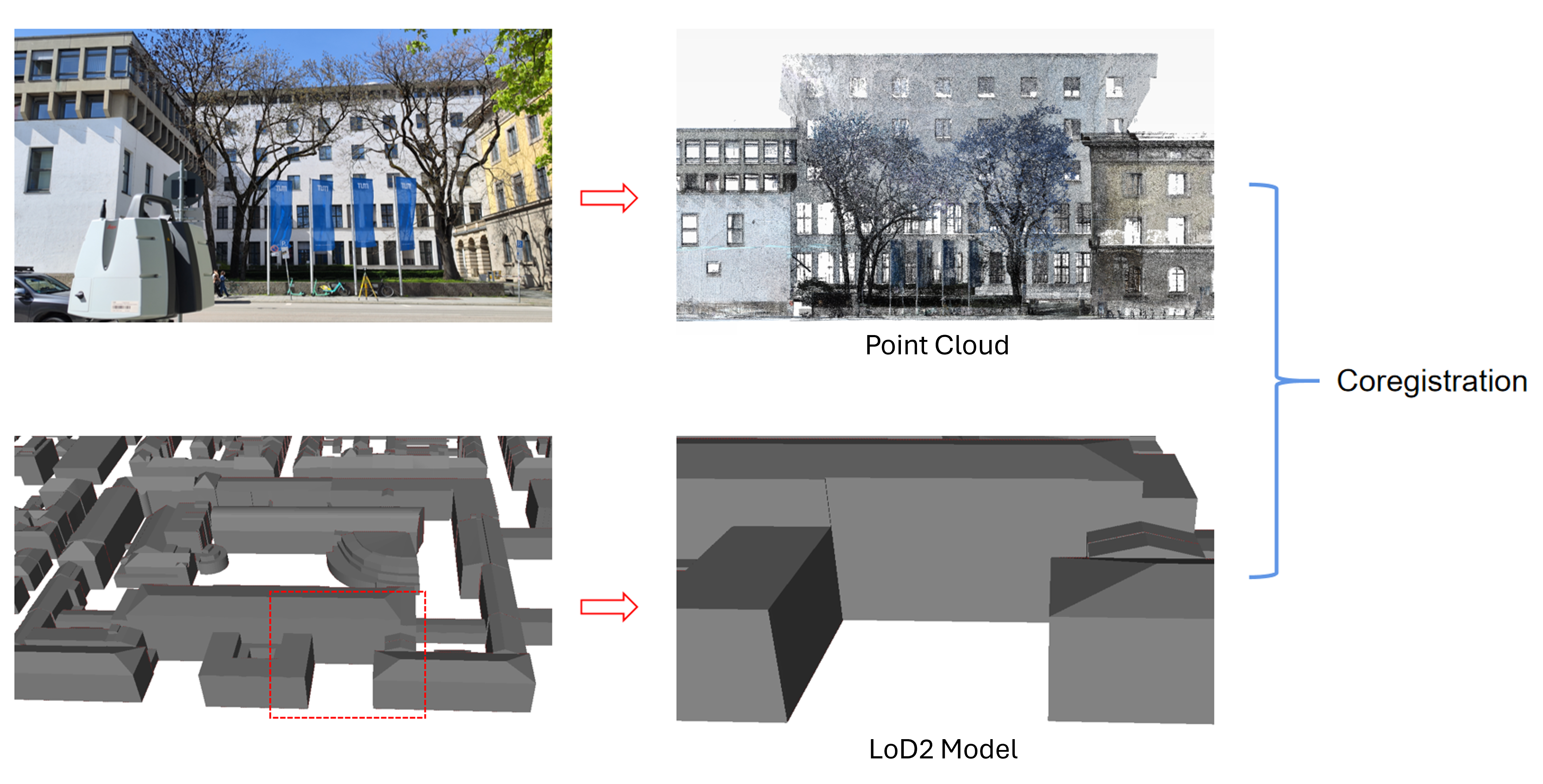}
    \caption{A potential downstream tasks of multimodal registration of~\gls{LoD}2 models and point clouds. }
    \label{fig:UseCsae-Coreg}
\end{figure*}

TUM2TWIN comprises various georeferenced data types from different sources of the same scene. 
For instance, the point clouds from TUM-TLS-24 and TUM-MLS-24 cover the same area. This characteristic makes the dataset highly suitable for investigating the registration and fusion of multimodal point clouds, offering an excellent testbed and robust data foundation, as we show in Figure~\ref{fig:UseCsae-Coreg}. 
Furthermore, TUM-MLS-24 is acquired using two distinct mobile laser scanning systems (Z+F FlexScan 22 and Leica BLK ARC), resulting in point clouds that vary significantly in quality and density. This variability provides valuable support for research on registration methods that account for point cloud uncertainty.

Additionally, the TUM2TWIN dataset includes an extensive collection of semantic 3D models that spatially overlap with TUM-TLS-24 and TUM-MLS-24, thereby enabling registration studies between point clouds and 3D models. Typically, 3D models as representations of as-designed conditions are not as current or detailed as as-built point clouds. Consequently, achieving precise registration between these data types is a prerequisite for subsequent 3D model enhancement. The inherent diversity of the TUM2TWIN dataset thus provides a basis for conducting multi-source data registration research.

\noindent \textbf{TUM2TWIN provides:} Various coregistered and georeferenced representations of the same objects: Point clouds (Sec.~\ref{sec:pcl}), images  (Sec.~\ref{sec:img}), networks  (Sec.~\ref{sec:networks}), and 3D models (Sec.~\ref{sec:3Dmodels}).

\subsection{ \textbf{Complementing Indoor Digital Twins} }

A concurrently collected dataset of indoor 3D scans is the ScanNet++ dataset \cite{yeshwanth2023scannet++}, which has Faro Premium laser scans with 40M points per scan, high-resolution 33MP DSLR images and HD iPhone RGBD videos for 1006 indoor scenes. The ScanNet++ dataset provides an enhanced benchmark for 3D semantic and instance segmentation tasks through its high resolution and complete geometry capture, as well as for newer novel view synthesis methods that rely on dense and high quality RGB image captures registered in 3D space. 

Several ScanNet++ scenes are captured at the same location as the TUM2TWIN dataset and hence can be coregistered with the indoor and outdoor point clouds to enable holistic indoor-outdoor semantic understanding of urban scenes in the form of \textit{indoor digital twins} to complement our outdoor digital twins. 

\noindent \textbf{TUM2TWIN provides:} Indoor and outdoor~\gls{TLS}  (Sec.~\ref{sec:tls}) and~\gls{MLS} point clouds  (Sec.~\ref{sec:mls}).

\subsection{ \textbf{Education and Teaching}}

While the project originates from research demands, the comprehensive scope of digital representations is useful for valuable use cases.
The datasets are utilized as practical exercise materials to enhance learning within the study program \cite{Dubois2021}.
In particular, the different representations demonstrate the focus and interconnections between the various disciplines.

\subsection{ \textbf{Industry and Public Authorities}}

The open datasets enable established and emerging companies to prototype, validate, and present novel products and services.
Moreover, public authorities can utilize the benchmark datasets to assess the quality of surveying service suppliers.


\section{Limitations and Outlook}
\label{limit}
To date, there has been no such comprehensive dataset representing a multitude of modalities as TUM2TWIN.
We firmly believe this will be a milestone for the community.
Yet, we acknowledge that there are still some limitations in our dataset, such as a lack of online acquisition modality in the form of constant observation cameras or Internet of Things (IoT) devices.
Also, although co-registered, as of now, the interiors are only partially captured by our laser scanners.
This step will also ensure essential data for creating georeferenced~\gls{BIM} models.
We plan to extend our data in the future in these valuable assets.
Furthermore, we acknowledge that frequently benchmarks are associated with the pre-defined routines for input and output results, as we provided in ZAHA \cite{wysocki2024zahaintroducinglevelfacade}. 
Yet, as we observe, our comprehensive dataset is already in use by multiple researchers for various purposes, we encourage the community to brute-force test and submit the feedback via our continuously updated website \footnote{https://tum2t.win/}.

\section{Conclusion}
In this paper, we propose a holistic Urban Digital Twin (UDT) benchmark dataset comprising multiple data representations of the same area of roughly 100,000 $m^2$, enabling various applications.
Since its conception in 2021, starting with solely a few point clouds and limited semantic models, the project evolved to its current state, boasting 32 different data subsets -- to date, the largest UDT benchmark dataset.
Based on our experience, we conclude that georeferencing is the key to i) enabling novel research directions and ii) enhancing the validation of existing ones.
One of the striking examples is the first-of-its-kind combination of~\gls{LoD}2, textured~\gls{LoD}2, and~\gls{LoD}3 models georeferenced with point clouds stemming from~\gls{TLS},~\gls{UAS}, and~\gls{MLS}, which enables ground-truth validation of~\gls{LoD}3 model reconstruction for the first time from various sensors. 
The georeferencing aspect is also pivotal for further integration of other datasets in the future, owing to its unique identifiers that allow for association between datasets across timestamps.
An example of enhancing current research is the NeRF and Gaussian Splatting, which, despite gaining significant attention, lacked outdoor real-data validation datasets of high-accuracy~\gls{TLS} point cloud or low-poly semantic 3D building models.
We are convinced that the TUM2TWIN dataset will foster further development around~\gls{UDT} and will benefit multiple research groups worldwide.





{
    \small
    \bibliographystyle{elsarticle-num-names}
    \bibliography{main}

\begin{thebibliography}{85}
\expandafter\ifx\csname natexlab\endcsname\relax\def\natexlab#1{#1}\fi
\providecommand{\url}[1]{\texttt{#1}}
\providecommand{\href}[2]{#2}
\providecommand{\path}[1]{#1}
\providecommand{\DOIprefix}{doi:}
\providecommand{\ArXivprefix}{arXiv:}
\providecommand{\URLprefix}{URL: }
\providecommand{\Pubmedprefix}{pmid:}
\providecommand{\doi}[1]{\href{http://dx.doi.org/#1}{\path{#1}}}
\providecommand{\Pubmed}[1]{\href{pmid:#1}{\path{#1}}}
\providecommand{\bibinfo}[2]{#2}
\ifx\xfnm\relax \def\xfnm[#1]{\unskip,\space#1}\fi
\bibitem[{Allen(2021)}]{allen2021digital}
\bibinfo{author}{B.~D. Allen},
\newblock \bibinfo{title}{Digital twins and living models at {NASA}},
\newblock in: \bibinfo{booktitle}{Digital Twin Summit}, \bibinfo{year}{2021}.
\bibitem[{Deng et~al.(2021)Deng, Zhang, and Shen}]{DENG2021125}
\bibinfo{author}{T.~Deng}, \bibinfo{author}{K.~Zhang}, \bibinfo{author}{Z.-J.~M. Shen},
\newblock \bibinfo{title}{A systematic review of a digital twin city: A new pattern of urban governance toward smart cities},
\newblock \bibinfo{journal}{Journal of Management Science and Engineering} \bibinfo{volume}{6} (\bibinfo{year}{2021}) \bibinfo{pages}{125--134}. \DOIprefix\doi{https://doi.org/10.1016/j.jmse.2021.03.003}.
\bibitem[{Kolbe and Donaubauer(2021)}]{Kolbe2021}
\bibinfo{author}{T.~H. Kolbe}, \bibinfo{author}{A.~Donaubauer},
\newblock \bibinfo{title}{Semantic {3D} city modeling and {BIM}},
\newblock in: \bibinfo{editor}{W.~Shi}, \bibinfo{editor}{M.~F. Goodchild}, \bibinfo{editor}{M.~Batty}, \bibinfo{editor}{M.-P. Kwan}, \bibinfo{editor}{A.~Zhang} (Eds.), \bibinfo{booktitle}{Urban Informatics}, \bibinfo{publisher}{Springer Singapore}, \bibinfo{address}{Singapore}, \bibinfo{year}{2021}, pp. \bibinfo{pages}{609--636}.
\bibitem[{Haala and Kada(2010)}]{HAALA2010570}
\bibinfo{author}{N.~Haala}, \bibinfo{author}{M.~Kada},
\newblock \bibinfo{title}{An update on automatic {3D} building reconstruction},
\newblock \bibinfo{journal}{ISPRS Journal of Photogrammetry and Remote Sensing} \bibinfo{volume}{65} (\bibinfo{year}{2010}) \bibinfo{pages}{570 -- 580}.
\bibitem[{Grieves(2014)}]{grieves2014digital}
\bibinfo{author}{M.~Grieves},
\newblock \bibinfo{title}{Digital twin: manufacturing excellence through virtual factory replication},
\newblock \bibinfo{journal}{White paper} \bibinfo{volume}{1} (\bibinfo{year}{2014}) \bibinfo{pages}{1--7}.
\bibitem[{Wysocki et~al.(2023)Wysocki, Xia, Wysocki, Grilli, Hoegner, Cremers, and Stilla}]{wysocki2023scan2lod3}
\bibinfo{author}{O.~Wysocki}, \bibinfo{author}{Y.~Xia}, \bibinfo{author}{M.~Wysocki}, \bibinfo{author}{E.~Grilli}, \bibinfo{author}{L.~Hoegner}, \bibinfo{author}{D.~Cremers}, \bibinfo{author}{U.~Stilla},
\newblock \bibinfo{title}{{Scan2LoD3}: Reconstructing semantic {3D} building models at {LoD3} using ray casting and {Bayesian} networks},
\newblock \bibinfo{journal}{IEEE/CVF Conference on Computer Vision and Pattern Recognition Workshops (CVPRW)}  (\bibinfo{year}{2023}) \bibinfo{pages}{6547--6557}.
\bibitem[{Scharstein and Szeliski(2002)}]{scharstein2002taxonomy}
\bibinfo{author}{D.~Scharstein}, \bibinfo{author}{R.~Szeliski},
\newblock \bibinfo{title}{A taxonomy and evaluation of dense two-frame stereo correspondence algorithms},
\newblock \bibinfo{journal}{International journal of computer vision} \bibinfo{volume}{47} (\bibinfo{year}{2002}) \bibinfo{pages}{7--42}.
\bibitem[{Kirillov et~al.(2023)Kirillov, Mintun, Ravi, Mao, Rolland, Gustafson, Xiao, Whitehead, Berg, Lo et~al.}]{kirillov2023segment}
\bibinfo{author}{A.~Kirillov}, \bibinfo{author}{E.~Mintun}, \bibinfo{author}{N.~Ravi}, \bibinfo{author}{H.~Mao}, \bibinfo{author}{C.~Rolland}, \bibinfo{author}{L.~Gustafson}, \bibinfo{author}{T.~Xiao}, \bibinfo{author}{S.~Whitehead}, \bibinfo{author}{A.~C. Berg}, \bibinfo{author}{W.-Y. Lo}, et~al.,
\newblock \bibinfo{title}{Segment anything},
\newblock in: \bibinfo{booktitle}{Proceedings of the IEEE/CVF international conference on computer vision}, \bibinfo{year}{2023}, pp. \bibinfo{pages}{4015--4026}.
\bibitem[{Hackel et~al.(2017)Hackel, Savinov, Ladicky, Wegner, Schindler, and Pollefeys}]{hackel2017semantic3d}
\bibinfo{author}{T.~Hackel}, \bibinfo{author}{N.~Savinov}, \bibinfo{author}{L.~Ladicky}, \bibinfo{author}{J.~D. Wegner}, \bibinfo{author}{K.~Schindler}, \bibinfo{author}{M.~Pollefeys},
\newblock \bibinfo{title}{Semantic3d.net: A new large-scale point cloud classification benchmark},
\newblock \bibinfo{journal}{arXiv preprint arXiv:1704.03847}  (\bibinfo{year}{2017}).
\bibitem[{De~Deuge et~al.(2013)De~Deuge, Quadros, Hung, and Douillard}]{SydneyDatasetde2013unsupervised}
\bibinfo{author}{M.~De~Deuge}, \bibinfo{author}{A.~Quadros}, \bibinfo{author}{C.~Hung}, \bibinfo{author}{B.~Douillard},
\newblock \bibinfo{title}{Unsupervised feature learning for classification of outdoor 3d scans},
\newblock \bibinfo{journal}{Australasian Conference on Robotics and Automation} \bibinfo{volume}{2} (\bibinfo{year}{2013}).
\bibitem[{Matrone et~al.(2020)Matrone, Lingua, Pierdicca, Malinverni, Paolanti, Grilli, Remondino, Murtiyoso, and Landes}]{archDatasetPaper}
\bibinfo{author}{F.~Matrone}, \bibinfo{author}{A.~Lingua}, \bibinfo{author}{R.~Pierdicca}, \bibinfo{author}{E.~S. Malinverni}, \bibinfo{author}{M.~Paolanti}, \bibinfo{author}{E.~Grilli}, \bibinfo{author}{F.~Remondino}, \bibinfo{author}{A.~Murtiyoso}, \bibinfo{author}{T.~Landes},
\newblock \bibinfo{title}{A benchmark for large-scale heritage point cloud semantic segmentation},
\newblock \bibinfo{journal}{The International Archives of the Photogrammetry, Remote Sensing and Spatial Information Sciences} \bibinfo{volume}{XLIII-B2-2020} (\bibinfo{year}{2020}) \bibinfo{pages}{1419--1426}.
\bibitem[{Deschaud et~al.(2021)Deschaud, Duque, Richa, Velasco-Forero, Marcotegui, and Goulette}]{deschaud2021pariscarla3d}
\bibinfo{author}{J.-E. Deschaud}, \bibinfo{author}{D.~Duque}, \bibinfo{author}{J.~P. Richa}, \bibinfo{author}{S.~Velasco-Forero}, \bibinfo{author}{B.~Marcotegui}, \bibinfo{author}{F.~Goulette},
\newblock \bibinfo{title}{{Paris-CARLA-3D}: A real and synthetic outdoor point cloud dataset for challenging tasks in {3D} mapping},
\newblock \bibinfo{journal}{Remote Sensing} \bibinfo{volume}{13} (\bibinfo{year}{2021}) \bibinfo{pages}{4713}.
\bibitem[{Tan et~al.(2020)Tan, Qin, Ma, Li, Du, Cai, Yang, and Li}]{tan2020toronto}
\bibinfo{author}{W.~Tan}, \bibinfo{author}{N.~Qin}, \bibinfo{author}{L.~Ma}, \bibinfo{author}{Y.~Li}, \bibinfo{author}{J.~Du}, \bibinfo{author}{G.~Cai}, \bibinfo{author}{K.~Yang}, \bibinfo{author}{J.~Li},
\newblock \bibinfo{title}{{Toronto-3D}: A large-scale mobile {LiDAR} dataset for semantic segmentation of urban roadways},
\newblock in: \bibinfo{booktitle}{IEEE/CVF Conference on Computer Vision and Pattern Recognition Workshops (CVPRW)}, \bibinfo{year}{2020}, pp. \bibinfo{pages}{202--203}.
\bibitem[{Serna et~al.(2014)Serna, Marcotegui, Goulette, and Deschaud}]{serna2014parisMadame}
\bibinfo{author}{A.~Serna}, \bibinfo{author}{B.~Marcotegui}, \bibinfo{author}{F.~Goulette}, \bibinfo{author}{J.-E. Deschaud},
\newblock \bibinfo{title}{{Paris-rue-Madame} database: As {3D} mobile laser scanner dataset for benchmarking urban detection, segmentation and classification methods},
\newblock in: \bibinfo{booktitle}{Proceedings of the International Conference on Pattern Recognition Applications and Methods. ACM, Angers, France, 6–8 March}, \bibinfo{year}{2014}, pp. \bibinfo{pages}{819--824}.
\bibitem[{Stilla and Xu(2023)}]{yushengChangeDetectionReview}
\bibinfo{author}{U.~Stilla}, \bibinfo{author}{Y.~Xu},
\newblock \bibinfo{title}{Change detection of urban objects using 3d point clouds: A review},
\newblock \bibinfo{journal}{ISPRS Journal of Photogrammetry and Remote Sensing} \bibinfo{volume}{197} (\bibinfo{year}{2023}) \bibinfo{pages}{228--255}. \DOIprefix\doi{https://doi.org/10.1016/j.isprsjprs.2023.01.010}.
\bibitem[{Armeni et~al.(2016)Armeni, Sener, Zamir, Jiang, Brilakis, Fischer, and Savarese}]{armeni20163d}
\bibinfo{author}{I.~Armeni}, \bibinfo{author}{O.~Sener}, \bibinfo{author}{A.~R. Zamir}, \bibinfo{author}{H.~Jiang}, \bibinfo{author}{I.~Brilakis}, \bibinfo{author}{M.~Fischer}, \bibinfo{author}{S.~Savarese},
\newblock \bibinfo{title}{{3D} semantic parsing of large-scale indoor spaces},
\newblock in: \bibinfo{booktitle}{Proceedings of the IEEE conference on computer vision and pattern recognition}, \bibinfo{year}{2016}, pp. \bibinfo{pages}{1534--1543}.
\bibitem[{Nex et~al.(2024)Nex, Stathopoulou, Remondino, Yang, Madhuanand, Yogender, Alsadik, Weinmann, Jutzi, and Qin}]{nex2024usegeo}
\bibinfo{author}{F.~Nex}, \bibinfo{author}{E.~Stathopoulou}, \bibinfo{author}{F.~Remondino}, \bibinfo{author}{M.~Yang}, \bibinfo{author}{L.~Madhuanand}, \bibinfo{author}{Y.~Yogender}, \bibinfo{author}{B.~Alsadik}, \bibinfo{author}{M.~Weinmann}, \bibinfo{author}{B.~Jutzi}, \bibinfo{author}{R.~Qin},
\newblock \bibinfo{title}{Usegeo-a uav-based multi-sensor dataset for geospatial research},
\newblock \bibinfo{journal}{ISPRS Open Journal of Photogrammetry and Remote Sensing}  (\bibinfo{year}{2024}) \bibinfo{pages}{100070}.
\bibitem[{Dong et~al.(2020)Dong, Liang, Yang, Xu, Zang, Li, Wang, Dai, Fan, Hyypp{\"a} et~al.}]{dong2020registration}
\bibinfo{author}{Z.~Dong}, \bibinfo{author}{F.~Liang}, \bibinfo{author}{B.~Yang}, \bibinfo{author}{Y.~Xu}, \bibinfo{author}{Y.~Zang}, \bibinfo{author}{J.~Li}, \bibinfo{author}{Y.~Wang}, \bibinfo{author}{W.~Dai}, \bibinfo{author}{H.~Fan}, \bibinfo{author}{J.~Hyypp{\"a}}, et~al.,
\newblock \bibinfo{title}{Registration of large-scale terrestrial laser scanner point clouds: A review and benchmark},
\newblock \bibinfo{journal}{ISPRS Journal of Photogrammetry and Remote Sensing} \bibinfo{volume}{163} (\bibinfo{year}{2020}) \bibinfo{pages}{327--342}.
\bibitem[{Schwab et~al.(2021)Schwab, Haas~Goschenhofer, and Wysocki}]{ingolstadtLoD3}
\bibinfo{author}{B.~Schwab}, \bibinfo{author}{S.~Haas~Goschenhofer}, \bibinfo{author}{O.~Wysocki}, \bibinfo{title}{{LoD3} road space models, release v0.8.1}, \bibinfo{howpublished}{\url{https://github.com/savenow/lod3-road-space-models}}, \bibinfo{year}{2021}. \bibinfo{note}{{Accessed}: 2023-01-30}.
\bibitem[{Kölle et~al.(2021)Kölle, Laupheimer, Schmohl, Haala, Rottensteiner, Wegner, and Ledoux}]{hessigheim}
\bibinfo{author}{M.~Kölle}, \bibinfo{author}{D.~Laupheimer}, \bibinfo{author}{S.~Schmohl}, \bibinfo{author}{N.~Haala}, \bibinfo{author}{F.~Rottensteiner}, \bibinfo{author}{J.~D. Wegner}, \bibinfo{author}{H.~Ledoux},
\newblock \bibinfo{title}{The hessigheim 3d (h3d) benchmark on semantic segmentation of high-resolution 3d point clouds and textured meshes from uav lidar and multi-view-stereo},
\newblock \bibinfo{journal}{ISPRS Open Journal of Photogrammetry and Remote Sensing} \bibinfo{volume}{1} (\bibinfo{year}{2021}) \bibinfo{pages}{100001}. \DOIprefix\doi{https://doi.org/10.1016/j.ophoto.2021.100001}.
\bibitem[{Gao et~al.(2021)Gao, Nan, Boom, and Ledoux}]{gaoSUM}
\bibinfo{author}{W.~Gao}, \bibinfo{author}{L.~Nan}, \bibinfo{author}{B.~Boom}, \bibinfo{author}{H.~Ledoux},
\newblock \bibinfo{title}{Sum: A benchmark dataset of semantic urban meshes},
\newblock \bibinfo{journal}{ISPRS Journal of Photogrammetry and Remote Sensing} \bibinfo{volume}{179} (\bibinfo{year}{2021}) \bibinfo{pages}{108--120}. \DOIprefix\doi{https://doi.org/10.1016/j.isprsjprs.2021.07.008}.
\bibitem[{Liao et~al.(2021)Liao, Xie, and Geiger}]{liao2021kitti}
\bibinfo{author}{Y.~Liao}, \bibinfo{author}{J.~Xie}, \bibinfo{author}{A.~Geiger},
\newblock \bibinfo{title}{{KITTI-360}: A novel dataset and benchmarks for urban scene understanding in {2D and 3D}},
\newblock \bibinfo{journal}{arXiv preprint arXiv:2109.13410}  (\bibinfo{year}{2021}).
\bibitem[{Lin et~al.(2022)Lin, Liu, Hu, Yan, Xie, and Huang}]{lin2022capturingUrban3Dscene}
\bibinfo{author}{L.~Lin}, \bibinfo{author}{Y.~Liu}, \bibinfo{author}{Y.~Hu}, \bibinfo{author}{X.~Yan}, \bibinfo{author}{K.~Xie}, \bibinfo{author}{H.~Huang},
\newblock \bibinfo{title}{Capturing, reconstructing, and simulating: the urbanscene3d dataset},
\newblock in: \bibinfo{booktitle}{European Conference on Computer Vision}, \bibinfo{organization}{Springer}, \bibinfo{year}{2022}, pp. \bibinfo{pages}{93--109}.
\bibitem[{Wang et~al.(2023)Wang, Huang, and Yang}]{wang2023building3d}
\bibinfo{author}{R.~Wang}, \bibinfo{author}{S.~Huang}, \bibinfo{author}{H.~Yang},
\newblock \bibinfo{title}{{Building3D}: A urban-scale dataset and benchmarks for learning roof structures from point clouds},
\newblock in: \bibinfo{booktitle}{Proceedings of the IEEE/CVF International Conference on Computer Vision}, \bibinfo{year}{2023}, pp. \bibinfo{pages}{20076--20086}.
\bibitem[{González-Collazo et~al.(2024)González-Collazo, Balado, Garrido, Grandío, Rashdi, Tsiranidou, {del Río-Barral}, Rúa, Puente, and Lorenzo}]{SUDdata_SiliviaGonzalez}
\bibinfo{author}{S.~M. González-Collazo}, \bibinfo{author}{J.~Balado}, \bibinfo{author}{I.~Garrido}, \bibinfo{author}{J.~Grandío}, \bibinfo{author}{R.~Rashdi}, \bibinfo{author}{E.~Tsiranidou}, \bibinfo{author}{P.~{del Río-Barral}}, \bibinfo{author}{E.~Rúa}, \bibinfo{author}{I.~Puente}, \bibinfo{author}{H.~Lorenzo},
\newblock \bibinfo{title}{Santiago urban dataset sud: Combination of handheld and mobile laser scanning point clouds},
\newblock \bibinfo{journal}{Expert Systems with Applications} \bibinfo{volume}{238} (\bibinfo{year}{2024}) \bibinfo{pages}{121842}.
\bibitem[{Gao et~al.(2024)Gao, Peters, and Stoter}]{buildingpcc}
\bibinfo{author}{W.~Gao}, \bibinfo{author}{R.~Peters}, \bibinfo{author}{J.~Stoter},
\newblock \bibinfo{title}{Building-pcc: Building point cloud completion benchmarks},
\newblock \bibinfo{journal}{ISPRS Annals of the Photogrammetry, Remote Sensing and Spatial Information Sciences} \bibinfo{volume}{X-4/W5-2024} (\bibinfo{year}{2024}) \bibinfo{pages}{179--186}. \DOIprefix\doi{10.5194/isprs-annals-X-4-W5-2024-179-2024}.
\bibitem[{Yazdi et~al.(2024)Yazdi, Shu, R{\"o}tzer, Petzold, and Ludwig}]{yazdi2024multilayered}
\bibinfo{author}{H.~Yazdi}, \bibinfo{author}{Q.~Shu}, \bibinfo{author}{T.~R{\"o}tzer}, \bibinfo{author}{F.~Petzold}, \bibinfo{author}{F.~Ludwig},
\newblock \bibinfo{title}{A multilayered urban tree dataset of point clouds, quantitative structure and graph models},
\newblock \bibinfo{journal}{Scientific Data} \bibinfo{volume}{11} (\bibinfo{year}{2024}) \bibinfo{pages}{28}.
\bibitem[{Raumonen et~al.(2013)Raumonen, Kaasalainen, {\AA}kerblom, Kaasalainen, Kaartinen, Vastaranta, Holopainen, Disney, and Lewis}]{raumonen2013fast}
\bibinfo{author}{P.~Raumonen}, \bibinfo{author}{M.~Kaasalainen}, \bibinfo{author}{M.~{\AA}kerblom}, \bibinfo{author}{S.~Kaasalainen}, \bibinfo{author}{H.~Kaartinen}, \bibinfo{author}{M.~Vastaranta}, \bibinfo{author}{M.~Holopainen}, \bibinfo{author}{M.~Disney}, \bibinfo{author}{P.~Lewis},
\newblock \bibinfo{title}{Fast automatic precision tree models from terrestrial laser scanner data},
\newblock \bibinfo{journal}{Remote Sensing} \bibinfo{volume}{5} (\bibinfo{year}{2013}) \bibinfo{pages}{491--520}.
\bibitem[{Yazdi et~al.(2023)Yazdi, Shu, R{\"o}tzer, Petzold, and Ludwig}]{yazdi2023treeml}
\bibinfo{author}{H.~Yazdi}, \bibinfo{author}{Q.~Shu}, \bibinfo{author}{T.~R{\"o}tzer}, \bibinfo{author}{F.~Petzold}, \bibinfo{author}{F.~Ludwig},
\newblock \bibinfo{title}{Treeml-data; a multidisciplinary and multilayer urban tree dataset},
\newblock \bibinfo{journal}{Springer Nature}  (\bibinfo{year}{2023}).
\bibitem[{Zhu et~al.(2020)Zhu, Gehrung, Huang, Borgmann, Sun, Hoegner, Hebel, Xu, and Stilla}]{zhu_tum-mls-2016_2020}
\bibinfo{author}{J.~Zhu}, \bibinfo{author}{J.~Gehrung}, \bibinfo{author}{R.~Huang}, \bibinfo{author}{B.~Borgmann}, \bibinfo{author}{Z.~Sun}, \bibinfo{author}{L.~Hoegner}, \bibinfo{author}{M.~Hebel}, \bibinfo{author}{Y.~Xu}, \bibinfo{author}{U.~Stilla},
\newblock \bibinfo{title}{{TUM}-{MLS}-2016: An annotated mobile {LiDAR} dataset of the {TUM} {City Campus} for semantic point cloud interpretation in urban areas},
\newblock \bibinfo{journal}{Remote Sensing} \bibinfo{volume}{12} (\bibinfo{year}{2020}) \bibinfo{pages}{1875}.
\bibitem[{Wysocki et~al.(2024)Wysocki, Tan, Froech, Xia, Wysocki, Hoegner, Cremers, and Holst}]{wysocki2024zahaintroducinglevelfacade}
\bibinfo{author}{O.~Wysocki}, \bibinfo{author}{Y.~Tan}, \bibinfo{author}{T.~Froech}, \bibinfo{author}{Y.~Xia}, \bibinfo{author}{M.~Wysocki}, \bibinfo{author}{L.~Hoegner}, \bibinfo{author}{D.~Cremers}, \bibinfo{author}{C.~Holst}, \bibinfo{title}{Zaha: Introducing the level of facade generalization and the large-scale point cloud facade semantic segmentation benchmark dataset}, \bibinfo{year}{2024}. \URLprefix \url{https://arxiv.org/abs/2411.04865}. \href{http://arxiv.org/abs/2411.04865}{{\tt arXiv:2411.04865}}.
\bibitem[{Wysocki et~al.(2022)Wysocki, Hoegner, and Stilla}]{tumfacadePaper}
\bibinfo{author}{O.~Wysocki}, \bibinfo{author}{L.~Hoegner}, \bibinfo{author}{U.~Stilla},
\newblock \bibinfo{title}{{TUM-FAÇADE}: Reviewing and enriching point cloud benchmarks for {façade} segmentation},
\newblock \bibinfo{journal}{The International Archives of the Photogrammetry, Remote Sensing and Spatial Information Sciences} \bibinfo{volume}{XLVI-2/W1-2022} (\bibinfo{year}{2022}) \bibinfo{pages}{529--536}.
\bibitem[{Anders et~al.(2024)Anders, Wang, Chang, Letard, Schulte, and Winiwarter}]{anders_uas_tum_downtown}
\bibinfo{author}{K.~Anders}, \bibinfo{author}{J.~Wang}, \bibinfo{author}{M.~Chang}, \bibinfo{author}{M.~Letard}, \bibinfo{author}{F.~Schulte}, \bibinfo{author}{L.~Winiwarter}, \bibinfo{title}{Terrestrial and {UAV} laser scanning point clouds of {TUM} {Campus} {Ottobrunn}}, \bibinfo{year}{2024}. \URLprefix \url{https://doi.org/10.5281/zenodo.14443336}. \DOIprefix\doi{10.5281/zenodo.14443336}.
\bibitem[{{Pix4D SA}(2024)}]{pix4dmatic}
\bibinfo{author}{{Pix4D SA}}, \bibinfo{title}{{Pix4Dmatic Software, Version 1.71.0}}, \bibinfo{howpublished}{\url{https://www.pix4d.com/product/pix4dmatic} (24 April 2025)}, \bibinfo{year}{2024}.
\bibitem[{Sch\"{o}nberger et~al.(2016)Sch\"{o}nberger, Zheng, Pollefeys, and Frahm}]{schoenberger2016mvs}
\bibinfo{author}{J.~L. Sch\"{o}nberger}, \bibinfo{author}{E.~Zheng}, \bibinfo{author}{M.~Pollefeys}, \bibinfo{author}{J.-M. Frahm},
\newblock \bibinfo{title}{{Pixelwise View Selection for Unstructured Multi-View Stereo}},
\newblock in: \bibinfo{booktitle}{European Conference on Computer Vision (ECCV)}, \bibinfo{year}{2016}.
\bibitem[{{LDBV}(2025)}]{LDBVals}
\bibinfo{author}{{LDBV}}, \bibinfo{title}{{Das Landesamt für Digitalisierung, Breitband und Vermessung (LDBV)}}, \bibinfo{howpublished}{\url{https://www.ldbv.bayern.de/produkte/landschaftsinformationen/laser.html}}, \bibinfo{year}{2025}. \bibinfo{note}{Accessed: 2025-03-22}.
\bibitem[{Winiwarter et~al.(2022)Winiwarter, Pena, Weiser, Anders, S{\'a}nchez, Searle, and H{\"o}fle}]{winiwarter2022virtual}
\bibinfo{author}{L.~Winiwarter}, \bibinfo{author}{A.~M.~E. Pena}, \bibinfo{author}{H.~Weiser}, \bibinfo{author}{K.~Anders}, \bibinfo{author}{J.~M. S{\'a}nchez}, \bibinfo{author}{M.~Searle}, \bibinfo{author}{B.~H{\"o}fle},
\newblock \bibinfo{title}{Virtual laser scanning with helios++: A novel take on ray tracing-based simulation of topographic full-waveform 3d laser scanning},
\newblock \bibinfo{journal}{Remote Sensing of Environment} \bibinfo{volume}{269} (\bibinfo{year}{2022}) \bibinfo{pages}{112772}.
\bibitem[{{Google}(2023)}]{googleStreetView}
\bibinfo{author}{{Google}}, \bibinfo{title}{{Google Street View}}, \bibinfo{howpublished}{\url{https://www.google.com/maps}}, \bibinfo{year}{2023}. \bibinfo{note}{Accessed: 2025-03-22}.
\bibitem[{Tang et~al.(2025)Tang, Li, Liang, Wysocki, Biljecki, Holst, and Jutzi}]{tang2025texture2lod3}
\bibinfo{author}{W.~Tang}, \bibinfo{author}{W.~Li}, \bibinfo{author}{X.~Liang}, \bibinfo{author}{O.~Wysocki}, \bibinfo{author}{F.~Biljecki}, \bibinfo{author}{C.~Holst}, \bibinfo{author}{B.~Jutzi},
\newblock \bibinfo{title}{{Texture2LoD3: Enabling LoD3 Building Reconstruction With Panoramic Images}},
\newblock \bibinfo{journal}{https://arxiv.org/abs/2504.05249}  (\bibinfo{year}{2025}). \href{http://arxiv.org/abs/2504.05249}{{\tt arXiv:2504.05249}}.
\bibitem[{Zhu et~al.(2023)Zhu, Xu, Hoegner, and Stilla}]{zhu2023}
\bibinfo{author}{J.~Zhu}, \bibinfo{author}{Y.~Xu}, \bibinfo{author}{L.~Hoegner}, \bibinfo{author}{U.~Stilla},
\newblock \bibinfo{title}{Generation of thermal point clouds from uncalibrated thermal infrared image sequences and mobile laser scans},
\newblock \bibinfo{journal}{IEEE Transactions on Instrumentation and Measurement} \bibinfo{volume}{72} (\bibinfo{year}{2023}) \bibinfo{pages}{1--16}. \DOIprefix\doi{10.1109/TIM.2023.3284942}.
\bibitem[{{LDBV}(2025)}]{LDBVorthophoto}
\bibinfo{author}{{LDBV}}, \bibinfo{title}{{Das Landesamt für Digitalisierung, Breitband und Vermessung (LDBV)}}, \bibinfo{howpublished}{\url{https://www.ldbv.bayern.de/produkte/luftbilder/orthophotos/bayernbefliegung.html}}, \bibinfo{year}{2025}. \bibinfo{note}{Accessed: 2025-03-22}.
\bibitem[{Torres et~al.(2017)Torres, Navas-Traver, Bibby, Lokas, Snoeij, Rommen, Osborne, Ceba-Vega, Potin, and Geudtner}]{Torres2017}
\bibinfo{author}{R.~Torres}, \bibinfo{author}{I.~Navas-Traver}, \bibinfo{author}{D.~Bibby}, \bibinfo{author}{S.~Lokas}, \bibinfo{author}{P.~Snoeij}, \bibinfo{author}{B.~Rommen}, \bibinfo{author}{S.~Osborne}, \bibinfo{author}{F.~Ceba-Vega}, \bibinfo{author}{P.~Potin}, \bibinfo{author}{D.~Geudtner},
\newblock \bibinfo{title}{Sentinel-1 {SAR} system and mission},
\newblock in: \bibinfo{booktitle}{2017 IEEE Radar Conference (RadarConf)}, \bibinfo{year}{2017}, pp. \bibinfo{pages}{1582--1585}. \DOIprefix\doi{10.1109/RADAR.2017.7944460}.
\bibitem[{Spoto et~al.(2012)Spoto, Sy, Laberinti, Martimort, Fernandez, Colin, Hoersch, and Meygret}]{Spoto2010}
\bibinfo{author}{F.~Spoto}, \bibinfo{author}{O.~Sy}, \bibinfo{author}{P.~Laberinti}, \bibinfo{author}{P.~Martimort}, \bibinfo{author}{V.~Fernandez}, \bibinfo{author}{O.~Colin}, \bibinfo{author}{B.~Hoersch}, \bibinfo{author}{A.~Meygret},
\newblock \bibinfo{title}{Overview of {Sentinel-2}},
\newblock in: \bibinfo{booktitle}{2012 IEEE International Geoscience and Remote Sensing Symposium}, \bibinfo{year}{2012}, pp. \bibinfo{pages}{1707--1710}. \DOIprefix\doi{10.1109/IGARSS.2012.6351195}.
\bibitem[{Bouzinac et~al.(2018)Bouzinac, Lafrance, Pessiot, Touli, Jung, Massera, Neveu-VanMalle, Espesset, Francesconi, Clerc, Jackson, Alhammoud, Viallefont, Cadau, Iannone, and Gascon}]{Bouzinac2018}
\bibinfo{author}{C.~Bouzinac}, \bibinfo{author}{B.~Lafrance}, \bibinfo{author}{L.~Pessiot}, \bibinfo{author}{D.~Touli}, \bibinfo{author}{M.~Jung}, \bibinfo{author}{S.~Massera}, \bibinfo{author}{M.~Neveu-VanMalle}, \bibinfo{author}{A.~Espesset}, \bibinfo{author}{B.~Francesconi}, \bibinfo{author}{S.~Clerc}, \bibinfo{author}{J.~Jackson}, \bibinfo{author}{B.~Alhammoud}, \bibinfo{author}{F.~Viallefont}, \bibinfo{author}{E.~Cadau}, \bibinfo{author}{R.~Iannone}, \bibinfo{author}{F.~Gascon},
\newblock \bibinfo{title}{Sentinel-2 level-1 calibration and validation status from the mission performance centre},
\newblock in: \bibinfo{booktitle}{IGARSS 2018 - 2018 IEEE International Geoscience and Remote Sensing Symposium}, \bibinfo{year}{2018}, pp. \bibinfo{pages}{4347--4350}. \DOIprefix\doi{10.1109/IGARSS.2018.8518033}.
\bibitem[{Hagolle et~al.(2017)Hagolle, Huc, Desjardins, Auer, and Richter}]{hagolle2017}
\bibinfo{author}{O.~Hagolle}, \bibinfo{author}{M.~Huc}, \bibinfo{author}{C.~Desjardins}, \bibinfo{author}{S.~Auer}, \bibinfo{author}{R.~Richter}, \bibinfo{title}{{MAJA} algorithm theoretical basis document}, \bibinfo{year}{2017}. \URLprefix \url{https://doi.org/10.5281/zenodo.1209633}. \DOIprefix\doi{10.5281/zenodo.1209633}.
\bibitem[{Greza et~al.(2023)Greza, Hoegner, Hirt, Roschlaub, and Stilla}]{greza2023satellite}
\bibinfo{author}{M.~Greza}, \bibinfo{author}{L.~Hoegner}, \bibinfo{author}{P.-R. Hirt}, \bibinfo{author}{R.~Roschlaub}, \bibinfo{author}{U.~Stilla},
\newblock \bibinfo{title}{Satellite network bavaria - mission and data processing},
\newblock \bibinfo{journal}{Publikationen der Deutschen Gesellschaft f{\"u}r Photogrammetrie, Fernerkundung und Geoinformation (DGPF)} \bibinfo{volume}{43} (\bibinfo{year}{2023}) \bibinfo{pages}{174--182}.
\bibitem[{Lenz and Greza(2023)}]{lenz2023simulation}
\bibinfo{author}{N.~Lenz}, \bibinfo{author}{M.~Greza},
\newblock \bibinfo{title}{Simulation of earth observation data utilizing a virtual satellite camera},
\newblock \bibinfo{journal}{Publikationen der Deutschen Gesellschaft f{\"u}r Photogrammetrie, Fernerkundung und Geoinformation (DGPF)} \bibinfo{volume}{43} (\bibinfo{year}{2023}) \bibinfo{pages}{50--60}.
\bibitem[{Wysocki et~al.(2024)Wysocki, Schwab, Beil, Holst, and Kolbe}]{wysocki2024reviewing}
\bibinfo{author}{O.~Wysocki}, \bibinfo{author}{B.~Schwab}, \bibinfo{author}{C.~Beil}, \bibinfo{author}{C.~Holst}, \bibinfo{author}{T.~H. Kolbe},
\newblock \bibinfo{title}{Reviewing open data semantic {3D} city models to develop novel {3D} reconstruction methods},
\newblock \bibinfo{journal}{The International Archives of the Photogrammetry, Remote Sensing and Spatial Information Sciences} \bibinfo{volume}{48} (\bibinfo{year}{2024}) \bibinfo{pages}{493--500}.
\bibitem[{Kolbe et~al.(2021)Kolbe, Kutzner, Smyth, Nagel, Roensdorf, and Heazel}]{kolbeOGCCityGeography2021}
\bibinfo{author}{T.~H. Kolbe}, \bibinfo{author}{T.~Kutzner}, \bibinfo{author}{C.~S. Smyth}, \bibinfo{author}{C.~Nagel}, \bibinfo{author}{C.~Roensdorf}, \bibinfo{author}{C.~Heazel}, \bibinfo{title}{{OGC City Geography Markup Language} ({{CityGML}}) {{Part}} 1: {{Conceptual Model Standard}} v3.0}, \bibinfo{year}{2021}.
\bibitem[{Kutzner et~al.(2023)Kutzner, Smyth, Nagel, Coors, {Vinasco-Alvarez}, Ishimaru, Yao, Heazel, and Kolbe}]{kutznerOGCCityGeography2023}
\bibinfo{author}{T.~Kutzner}, \bibinfo{author}{C.~Smyth}, \bibinfo{author}{C.~Nagel}, \bibinfo{author}{V.~Coors}, \bibinfo{author}{D.~{Vinasco-Alvarez}}, \bibinfo{author}{N.~Ishimaru}, \bibinfo{author}{Z.~Yao}, \bibinfo{author}{C.~Heazel}, \bibinfo{author}{T.~H. Kolbe}, \bibinfo{title}{{OGC City Geography Markup Language ({{CityGML}}) Version 3.0 Part 2: {{GML}} Encoding Standard}}, \bibinfo{publisher}{Open Geospatial Consortium}, \bibinfo{year}{2023}.
\bibitem[{Biljecki et~al.(2015)Biljecki, Stoter, Ledoux, Zlatanova, and {\c C}{\"o}ltekin}]{biljeckiApplications3DCity2015}
\bibinfo{author}{F.~Biljecki}, \bibinfo{author}{J.~Stoter}, \bibinfo{author}{H.~Ledoux}, \bibinfo{author}{S.~Zlatanova}, \bibinfo{author}{A.~{\c C}{\"o}ltekin},
\newblock \bibinfo{title}{Applications of {3D} city models: {State} of the art review},
\newblock \bibinfo{journal}{ISPRS International Journal of Geo-Information} \bibinfo{volume}{4} (\bibinfo{year}{2015}) \bibinfo{pages}{2842--2889}.
\bibitem[{Yao et~al.(2018)Yao, Nagel, Kunde, Hudra, Willkomm, Donaubauer, Adolphi, and Kolbe}]{yao3DCityDB3DGeodatabase2018}
\bibinfo{author}{Z.~Yao}, \bibinfo{author}{C.~Nagel}, \bibinfo{author}{F.~Kunde}, \bibinfo{author}{G.~Hudra}, \bibinfo{author}{P.~Willkomm}, \bibinfo{author}{A.~Donaubauer}, \bibinfo{author}{T.~Adolphi}, \bibinfo{author}{T.~H. Kolbe},
\newblock \bibinfo{title}{{{3DCityDB}} - a {{3D}} geodatabase solution for the management, analysis, and visualization of semantic {{3D}} city models based on {{CityGML}}},
\newblock \bibinfo{journal}{Open Geospatial Data, Software and Standards} \bibinfo{volume}{3} (\bibinfo{year}{2018}). \DOIprefix\doi{10.1186/s40965-018-0046-7}.
\bibitem[{Ledoux et~al.(2021)Ledoux, Biljecki, Dukai, Kumar, Peters, Stoter, and Commandeur}]{3dfier}
\bibinfo{author}{H.~Ledoux}, \bibinfo{author}{F.~Biljecki}, \bibinfo{author}{B.~Dukai}, \bibinfo{author}{K.~Kumar}, \bibinfo{author}{R.~Peters}, \bibinfo{author}{J.~Stoter}, \bibinfo{author}{T.~Commandeur},
\newblock \bibinfo{title}{3dfier: Automatic reconstruction of 3d city models},
\newblock \bibinfo{journal}{Journal of Open Source Software} \bibinfo{volume}{6} (\bibinfo{year}{2021}) \bibinfo{pages}{2866}. \DOIprefix\doi{10.21105/joss.02866}.
\bibitem[{Huang(2024)}]{chenhao_msc_2023}
\bibinfo{author}{C.~Huang}, \bibinfo{title}{3D Cities: Evaluating 3D building reconstruction tools using point clouds}, Master's thesis, Department Aerospace and Geodesy, Technical University of Munich, \bibinfo{year}{2024}.
\bibitem[{Roschlaub and Batscheider(2016)}]{RoschlaubBatscheider}
\bibinfo{author}{R.~Roschlaub}, \bibinfo{author}{J.~Batscheider},
\newblock \bibinfo{title}{An {INSPIRE}-conform {3D} building model of {Bavaria} using cadastre information, {LiDAR} and image matching},
\newblock \bibinfo{journal}{The International Archives of the Photogrammetry, Remote Sensing and Spatial Information Sciences} \bibinfo{volume}{XLI-B4} (\bibinfo{year}{2016}) \bibinfo{pages}{747--754}.
\bibitem[{{3D Mapping Solutions}(2023)}]{mofa}
\bibinfo{author}{{3D Mapping Solutions}}, \bibinfo{title}{{MoSES} mobile mapping platform - technical details}, \bibinfo{howpublished}{\url{https://www.3d-mapping.de/ueber-uns/unternehmensbereiche/data-acquisition/unser-vermessungssystem/}}, \bibinfo{year}{2023}. \bibinfo{note}{{Accessed}: 2023-01-30}.
\bibitem[{Beil and Kolbe(2024)}]{beil2024}
\bibinfo{author}{C.~Beil}, \bibinfo{author}{T.~H. Kolbe},
\newblock \bibinfo{title}{Applications for semantic 3d streetspace models and their requirements—a review and look at the road ahead},
\newblock \bibinfo{journal}{ISPRS International Journal of Geo-Information} \bibinfo{volume}{13} (\bibinfo{year}{2024}). \DOIprefix\doi{10.3390/ijgi13100363}.
\bibitem[{Schwab et~al.(2020)Schwab, Beil, and Kolbe}]{schwab2020}
\bibinfo{author}{B.~Schwab}, \bibinfo{author}{C.~Beil}, \bibinfo{author}{T.~H. Kolbe},
\newblock \bibinfo{title}{Spatio-semantic road space modeling for vehicle–pedestrian simulation to test automated driving systems},
\newblock \bibinfo{journal}{Sustainability} \bibinfo{volume}{12} (\bibinfo{year}{2020}). \DOIprefix\doi{10.3390/su12093799}.
\bibitem[{Münzinger et~al.(2022)Münzinger, Prechtel, and Behnisch}]{muenzinger2022}
\bibinfo{author}{M.~Münzinger}, \bibinfo{author}{N.~Prechtel}, \bibinfo{author}{M.~Behnisch},
\newblock \bibinfo{title}{Mapping the urban forest in detail: From lidar point clouds to 3d tree models},
\newblock \bibinfo{journal}{Urban Forestry \& Urban Greening} \bibinfo{volume}{74} (\bibinfo{year}{2022}) \bibinfo{pages}{127637}. \DOIprefix\doi{https://doi.org/10.1016/j.ufug.2022.127637}.
\bibitem[{Zagst(2023)}]{zagst_msc_2023}
\bibinfo{author}{S.~Zagst}, \bibinfo{title}{Automatisierte Generierung eines Baumkatasters aus Punktwolken in unterschiedlichen urbanen Umgebungen}, Master's thesis, Department Aerospace and Geodesy, Technical University of Munich, \bibinfo{year}{2023}.
\bibitem[{Bieringer et~al.(2024)Bieringer, Wysocki, Tuttas, Hoegner, and Holst}]{bieringer2024analyzing}
\bibinfo{author}{A.~Bieringer}, \bibinfo{author}{O.~Wysocki}, \bibinfo{author}{S.~Tuttas}, \bibinfo{author}{L.~Hoegner}, \bibinfo{author}{C.~Holst},
\newblock \bibinfo{title}{Analyzing the impact of semantic lod3 building models on image-based vehicle localization},
\newblock \bibinfo{journal}{ISPRS Annals of the Photogrammetry, Remote Sensing and Spatial Information Sciences} \bibinfo{volume}{10} (\bibinfo{year}{2024}) \bibinfo{pages}{55--62}.
\bibitem[{Chen et~al.(2024)Chen, Shi, Nan, Xiong, and Zhu}]{PolyGNNZhaiyu}
\bibinfo{author}{Z.~Chen}, \bibinfo{author}{Y.~Shi}, \bibinfo{author}{L.~Nan}, \bibinfo{author}{Z.~Xiong}, \bibinfo{author}{X.~X. Zhu},
\newblock \bibinfo{title}{Polygnn: Polyhedron-based graph neural network for 3d building reconstruction from point clouds},
\newblock \bibinfo{journal}{ISPRS Journal of Photogrammetry and Remote Sensing} \bibinfo{volume}{218} (\bibinfo{year}{2024}) \bibinfo{pages}{693--706}. \DOIprefix\doi{https://doi.org/10.1016/j.isprsjprs.2024.09.031}.
\bibitem[{Wysocki et~al.(2023)Wysocki, Hoegner, and Stilla}]{wysockiMLS2LoD3}
\bibinfo{author}{O.~Wysocki}, \bibinfo{author}{L.~Hoegner}, \bibinfo{author}{U.~Stilla},
\newblock \bibinfo{title}{{MLS2LoD3}: Refining low lods building models with mls point clouds to reconstruct semantic lod3 building models},
\newblock \bibinfo{journal}{International 3D GeoInfo Conference 2023, Recent Advances in 3D Geoinformation Science}  (\bibinfo{year}{2023}) \bibinfo{pages}{367--380}.
\bibitem[{Wang et~al.(2024)Wang, Jiao, Fan, and Zhou}]{wang2024framework}
\bibinfo{author}{Y.~Wang}, \bibinfo{author}{W.~Jiao}, \bibinfo{author}{H.~Fan}, \bibinfo{author}{G.~Zhou},
\newblock \bibinfo{title}{A framework for fully automated reconstruction of semantic building model at urban-scale using textured lod2 data},
\newblock \bibinfo{journal}{ISPRS Journal of Photogrammetry and Remote Sensing} \bibinfo{volume}{216} (\bibinfo{year}{2024}) \bibinfo{pages}{90--108}.
\bibitem[{Weinmann et~al.(2013)Weinmann, Jutzi, and Mallet}]{jutziFeatures}
\bibinfo{author}{M.~Weinmann}, \bibinfo{author}{B.~Jutzi}, \bibinfo{author}{C.~Mallet},
\newblock \bibinfo{title}{Feature relevance assessment for the semantic interpretation of {3D} point cloud data},
\newblock \bibinfo{journal}{ISPRS Annals of the Photogrammetry, Remote Sensing and Spatial Information Sciences} \bibinfo{volume}{II-5/W2} (\bibinfo{year}{2013}) \bibinfo{pages}{313--318}.
\bibitem[{Tan et~al.(2023)Tan, Wysocki, Hoegner, and Stilla}]{yuetanDeepLearning}
\bibinfo{author}{Y.~Tan}, \bibinfo{author}{O.~Wysocki}, \bibinfo{author}{L.~Hoegner}, \bibinfo{author}{U.~Stilla},
\newblock \bibinfo{title}{Classifying point clouds at the facade-level using geometric features and deep learning networks},
\newblock \bibinfo{journal}{Accepted to proceedings of 3D GeoInfo 2023, Lecture Notes in Geoinformation and Cartography}  (\bibinfo{year}{2023}).
\bibitem[{Wysocki et~al.(2022)Wysocki, Grilli, Hoegner, and Stilla}]{wysockiVisibility}
\bibinfo{author}{O.~Wysocki}, \bibinfo{author}{E.~Grilli}, \bibinfo{author}{L.~Hoegner}, \bibinfo{author}{U.~Stilla},
\newblock \bibinfo{title}{Combining visibility analysis and deep learning for refinement of semantic {3D} building models by conflict classification},
\newblock \bibinfo{journal}{ISPRS Annals of the Photogrammetry, Remote Sensing and Spatial Information Sciences} \bibinfo{volume}{X-4/W2-2022} (\bibinfo{year}{2022}) \bibinfo{pages}{289--296}.
\bibitem[{Tuttas et~al.(2015)Tuttas, Stilla, Braun, and Borrmann}]{tuttas2015validation}
\bibinfo{author}{S.~Tuttas}, \bibinfo{author}{U.~Stilla}, \bibinfo{author}{A.~Braun}, \bibinfo{author}{A.~Borrmann},
\newblock \bibinfo{title}{Validation of {BIM} components by photogrammetric point clouds for construction site monitoring},
\newblock \bibinfo{journal}{ISPRS Annals of the Photogrammetry, Remote Sensing and Spatial Information Sciences} \bibinfo{volume}{II-3/W4} (\bibinfo{year}{2015}) \bibinfo{pages}{231–237}.
\bibitem[{Fröch et~al.(2025)Fröch, Wysocki, Xia, Xie, Schwab, Cremers, and Kolbe}]{facadiffyFroech}
\bibinfo{author}{T.~Fröch}, \bibinfo{author}{O.~Wysocki}, \bibinfo{author}{Y.~Xia}, \bibinfo{author}{J.~Xie}, \bibinfo{author}{B.~Schwab}, \bibinfo{author}{D.~Cremers}, \bibinfo{author}{T.~H. Kolbe},
\newblock \bibinfo{title}{Facadiffy: Inpainting unseen facade parts using diffusion models},
\newblock \bibinfo{journal}{Accepted for ISPRS Annals of the Photogrammetry, Remote Sensing and Spatial Information Sciences (ISPRS Geospatial Week)}  (\bibinfo{year}{2025}).
\bibitem[{Tuttas and Stilla(2013)}]{tuttas_reconstruction_2013}
\bibinfo{author}{S.~Tuttas}, \bibinfo{author}{U.~Stilla},
\newblock \bibinfo{title}{Reconstruction of fa\c{c}ades in point clouds from multi aspect oblique {ALS}},
\newblock \bibinfo{journal}{ISPRS Annals of the Photogrammetry, Remote Sensing and Spatial Information Sciences} \bibinfo{volume}{II-3/W3} (\bibinfo{year}{2013}) \bibinfo{pages}{91--96}.
\bibitem[{Zhu et~al.(2021)Zhu, Xu, Ye, Hoegner, and Stilla}]{zhu2021fusion}
\bibinfo{author}{J.~Zhu}, \bibinfo{author}{Y.~Xu}, \bibinfo{author}{Z.~Ye}, \bibinfo{author}{L.~Hoegner}, \bibinfo{author}{U.~Stilla},
\newblock \bibinfo{title}{Fusion of urban 3d point clouds with thermal attributes using {MLS} data and {TIR} image sequences},
\newblock \bibinfo{journal}{Infrared Physics {\&} Technology} \bibinfo{volume}{113} (\bibinfo{year}{2021}) \bibinfo{pages}{103622}. \DOIprefix\doi{10.1016/j.infrared.2020.103622}.
\bibitem[{Biswanath et~al.(2023)Biswanath, Hoegner, and Stilla}]{biswanath2023thermal}
\bibinfo{author}{M.~K. Biswanath}, \bibinfo{author}{L.~Hoegner}, \bibinfo{author}{U.~Stilla},
\newblock \bibinfo{title}{Thermal mapping from point clouds to 3d building model facades},
\newblock \bibinfo{journal}{Remote Sensing} \bibinfo{volume}{15} (\bibinfo{year}{2023}) \bibinfo{pages}{4830}.
\bibitem[{Borrmann et~al.(2023)Borrmann, Biswanath, Braun, Zhaiyu, Cremers, Heeramaglore, Hoegner, Mehranfar, Kolbe, Petzold, Rueda, Solonets, and Zhu}]{borrmann2023ai4twinning}
\bibinfo{author}{A.~Borrmann}, \bibinfo{author}{M.~K. Biswanath}, \bibinfo{author}{A.~Braun}, \bibinfo{author}{C.~Zhaiyu}, \bibinfo{author}{D.~Cremers}, \bibinfo{author}{M.~Heeramaglore}, \bibinfo{author}{L.~Hoegner}, \bibinfo{author}{M.~Mehranfar}, \bibinfo{author}{T.~H. Kolbe}, \bibinfo{author}{F.~Petzold}, \bibinfo{author}{A.~Rueda}, \bibinfo{author}{S.~Solonets}, \bibinfo{author}{X.~Zhu},
\newblock \bibinfo{title}{Artificial intelligence for the automated creation of multi-scale digital twins of the built world – ai4twinning},
\newblock in: \bibinfo{booktitle}{Proc. of the 18th 3D GeoInfo Conference}, \bibinfo{year}{2023}, pp. \bibinfo{pages}{233--247}.
\bibitem[{Mildenhall et~al.(2021)Mildenhall, Srinivasan, Tancik, Barron, Ramamoorthi, and Ng}]{mildenhall2021nerf}
\bibinfo{author}{B.~Mildenhall}, \bibinfo{author}{P.~P. Srinivasan}, \bibinfo{author}{M.~Tancik}, \bibinfo{author}{J.~T. Barron}, \bibinfo{author}{R.~Ramamoorthi}, \bibinfo{author}{R.~Ng},
\newblock \bibinfo{title}{Nerf: Representing scenes as neural radiance fields for view synthesis},
\newblock \bibinfo{journal}{Communications of the ACM} \bibinfo{volume}{65} (\bibinfo{year}{2021}) \bibinfo{pages}{99--106}.
\bibitem[{Petrovska and Jutzi(2024)}]{petrovska2024vision}
\bibinfo{author}{I.~Petrovska}, \bibinfo{author}{B.~Jutzi},
\newblock \bibinfo{title}{Vision through obstacles—{3D} geometric reconstruction and evaluation of {Neural Radiance Fields (NeRFs)}},
\newblock \bibinfo{journal}{Remote Sensing} \bibinfo{volume}{16} (\bibinfo{year}{2024}) \bibinfo{pages}{1188}.
\bibitem[{Tancik et~al.(2023)Tancik, Weber, Ng, Li, Yi, Wang, Kristoffersen, Austin, Salahi, Ahuja et~al.}]{tancik2023nerfstudio}
\bibinfo{author}{M.~Tancik}, \bibinfo{author}{E.~Weber}, \bibinfo{author}{E.~Ng}, \bibinfo{author}{R.~Li}, \bibinfo{author}{B.~Yi}, \bibinfo{author}{T.~Wang}, \bibinfo{author}{A.~Kristoffersen}, \bibinfo{author}{J.~Austin}, \bibinfo{author}{K.~Salahi}, \bibinfo{author}{A.~Ahuja}, et~al.,
\newblock \bibinfo{title}{Nerfstudio: A modular framework for neural radiance field development},
\newblock in: \bibinfo{booktitle}{ACM SIGGRAPH 2023 Conference Proceedings}, \bibinfo{year}{2023}, pp. \bibinfo{pages}{1--12}.
\bibitem[{Barron et~al.(2022)Barron, Mildenhall, Verbin, Srinivasan, and Hedman}]{barron2022mip}
\bibinfo{author}{J.~T. Barron}, \bibinfo{author}{B.~Mildenhall}, \bibinfo{author}{D.~Verbin}, \bibinfo{author}{P.~P. Srinivasan}, \bibinfo{author}{P.~Hedman},
\newblock \bibinfo{title}{Mip-nerf 360: Unbounded anti-aliased neural radiance fields},
\newblock in: \bibinfo{booktitle}{Proceedings of the IEEE/CVF conference on computer vision and pattern recognition}, \bibinfo{year}{2022}, pp. \bibinfo{pages}{5470--5479}.
\bibitem[{M{\"u}ller et~al.(2022)M{\"u}ller, Evans, Schied, and Keller}]{muller2022instant}
\bibinfo{author}{T.~M{\"u}ller}, \bibinfo{author}{A.~Evans}, \bibinfo{author}{C.~Schied}, \bibinfo{author}{A.~Keller},
\newblock \bibinfo{title}{Instant neural graphics primitives with a multiresolution hash encoding},
\newblock \bibinfo{journal}{ACM transactions on graphics (TOG)} \bibinfo{volume}{41} (\bibinfo{year}{2022}) \bibinfo{pages}{1--15}.
\bibitem[{Kerbl et~al.(2023)Kerbl, Kopanas, Leimk{\"u}hler, and Drettakis}]{kerbl20233d}
\bibinfo{author}{B.~Kerbl}, \bibinfo{author}{G.~Kopanas}, \bibinfo{author}{T.~Leimk{\"u}hler}, \bibinfo{author}{G.~Drettakis},
\newblock \bibinfo{title}{3d gaussian splatting for real-time radiance field rendering.},
\newblock \bibinfo{journal}{ACM Trans. Graph.} \bibinfo{volume}{42} (\bibinfo{year}{2023}) \bibinfo{pages}{139--1}.
\bibitem[{Zhang et~al.(2024)Zhang, Wysocki, Urban, and Jutzi}]{zhang2024cdgs}
\bibinfo{author}{Q.~Zhang}, \bibinfo{author}{O.~Wysocki}, \bibinfo{author}{S.~Urban}, \bibinfo{author}{B.~Jutzi},
\newblock \bibinfo{title}{Cdgs: Confidence-aware depth regularization for 3d gaussian splatting},
\newblock \bibinfo{journal}{The International Archives of the Photogrammetry, Remote Sensing and Spatial Information Sciences} \bibinfo{volume}{48} (\bibinfo{year}{2024}) \bibinfo{pages}{189--196}.
\bibitem[{Willenborg et~al.(2018)Willenborg, P{\"u}ltz, and Kolbe}]{willenborg2018integration}
\bibinfo{author}{B.~Willenborg}, \bibinfo{author}{M.~P{\"u}ltz}, \bibinfo{author}{T.~H. Kolbe},
\newblock \bibinfo{title}{Integration of semantic 3d city models and 3d mesh models for accuracy improvements of solar potential analyses},
\newblock \bibinfo{journal}{The International Archives of the Photogrammetry, Remote Sensing and Spatial Information Sciences} \bibinfo{volume}{42} (\bibinfo{year}{2018}) \bibinfo{pages}{223--230}.
\bibitem[{Rueda~Segura et~al.(2025)Rueda~Segura, Sun, Bratoev, and Petzold}]{caadria2025}
\bibinfo{author}{A.~Rueda~Segura}, \bibinfo{author}{Y.~Sun}, \bibinfo{author}{I.~Bratoev}, \bibinfo{author}{F.~Petzold},
\newblock \bibinfo{title}{Generating façade segmentation datasets using diffusion models},
\newblock in: \bibinfo{booktitle}{Computer-Aided Architectural Design Research in Asia (CAADRIA)}, \bibinfo{year}{2025}. \bibinfo{note}{Accepted for publication, forthcoming}.
\bibitem[{Heeramaglore and Kolbe(2022)}]{heeramaglore2022}
\bibinfo{author}{M.~Heeramaglore}, \bibinfo{author}{T.~H. Kolbe},
\newblock \bibinfo{title}{Semantically enriched voxels as a common representation for comparison and evaluation of 3d building models},
\newblock in: \bibinfo{booktitle}{Proceedings of the 17th International 3D GeoInfo Conference 2022}, ISPRS Annals of the Photogrammetry, Remote Sensing and Spatial Information Sciences, \bibinfo{organization}{UNSW Sydney}, \bibinfo{year}{2022}. \URLprefix \url{https://www.isprs-ann-photogramm-remote-sens-spatial-inf-sci.net/X-4-W2-2022/89/2022/}. \DOIprefix\doi{10.5194/isprs-annals-X-4-W2-2022-89-2022}.
\bibitem[{Yeshwanth et~al.(2023)Yeshwanth, Liu, Nie{\ss}ner, and Dai}]{yeshwanth2023scannet++}
\bibinfo{author}{C.~Yeshwanth}, \bibinfo{author}{Y.-C. Liu}, \bibinfo{author}{M.~Nie{\ss}ner}, \bibinfo{author}{A.~Dai},
\newblock \bibinfo{title}{{ScanNet}++: A high-fidelity dataset of 3d indoor scenes},
\newblock in: \bibinfo{booktitle}{Proceedings of the IEEE/CVF International Conference on Computer Vision}, \bibinfo{year}{2023}, pp. \bibinfo{pages}{12--22}.
\bibitem[{Dubois et~al.(2021)Dubois, Jutzi, Olijslagers, Pathe, Schmullius, Stelmaszczuk-G\'orska, Vandenbroucke, and Weinmann}]{Dubois2021}
\bibinfo{author}{C.~Dubois}, \bibinfo{author}{B.~Jutzi}, \bibinfo{author}{M.~Olijslagers}, \bibinfo{author}{C.~Pathe}, \bibinfo{author}{C.~Schmullius}, \bibinfo{author}{M.~A. Stelmaszczuk-G\'orska}, \bibinfo{author}{D.~Vandenbroucke}, \bibinfo{author}{M.~Weinmann},
\newblock \bibinfo{title}{Knowledge and skills related to active optical sensors in the body of knowledge for earth observation and geoinformation ({EO4GEO BOK})},
\newblock \bibinfo{journal}{ISPRS Annals of the Photogrammetry, Remote Sensing and Spatial Information Sciences} \bibinfo{volume}{V-5-2021} (\bibinfo{year}{2021}) \bibinfo{pages}{9--16}. \DOIprefix\doi{10.5194/isprs-annals-V-5-2021-9-2021}.

\end{thebibliography}
}




\end{document}